\documentclass[pdflatex,sn-nature]{sn-jnl}

\usepackage{graphicx}
\usepackage{multirow}
\usepackage{amsmath,amssymb,amsfonts}
\usepackage{booktabs}
\usepackage{longtable}
\usepackage{nicefrac}

\makeatletter
\renewenvironment{table}[1][]{\begin{tableorg}[#1]}{\end{tableorg}}
\makeatother

\begin{document}

\title[GlaucomaMarker]{GlaucomaMarker - Biomarker-Grounded Glaucoma Vision-Language Model}

\author[1]{\fnm{Ahammed Sakir} \sur{Nabil}}
\author[1]{\fnm{Sadia} \sur{Siraz}}
\author[2]{\fnm{Nikita} \sur{Chincholkar}}
\author[2]{\fnm{Atalie C.} \sur{Thompson}}
\author[3]{\fnm{Jason} \sur{Xiao}}
\author[3]{\fnm{Theodore} \sur{Leng}}
\author[4]{\fnm{Jennifer I.} \sur{Lim}}
\author*[1]{\fnm{Minhaj Nur} \sur{Alam}}\email{malam8@charlotte.edu}

\affil*[1]{\orgdiv{Department of Electrical and Computer Engineering}, \orgname{University of North Carolina at Charlotte}, \orgaddress{\city{Charlotte}, \state{North Carolina}, \country{USA}}}

\affil[2]{\orgdiv{Department of Ophthalmology}, \orgname{Wake Forest University School of Medicine}, \orgaddress{\city{Winston-Salem}, \state{North Carolina}, \country{USA}}}

\affil[3]{\orgdiv{Department of Ophthalmology, Byers Eye Institute}, \orgname{Stanford University School of Medicine}, \orgaddress{\city{Palo Alto}, \state{California}, \country{USA}}}

\affil[4]{\orgdiv{Department of Ophthalmology and Visual Science}, \orgname{University of Illinois Chicago}, \orgaddress{\city{Chicago}, \state{Illinois}, \country{USA}}}

\abstract{Glaucoma is a leading cause of irreversible blindness. In the current clinical setting, glaucoma diagnosis is generally conducted by assessing optic disc cupping and asymmetry. Vision-language models (VLMs) have been utilized in recent years for glaucoma, mostly trained on fixed templates or AI-generated captions, as no public text describes the biomarkers. Our proposed framework, GlaucomaMarker, hypothesizes that retinal structural biomarkers can ground glaucoma-based VLMs. It converts public segmentation masks into sentences describing eleven biomarkers, to teach retinal foundation models to read them. We applied this framework to RETFound (GlaucomaMarker-R) and FLAIR (GlaucomaMarker-F), training on 9,778 images that yielded 45,190 image-text pairs. On a real-world clinical cohort, GlaucomaMarker-F detects glaucoma with an AUROC of 0.905, against 0.848 for FLAIR, and both read the biomarkers where baselines stay near random guessing. Since the biomarker text is computed from existing masks, it comes at no extra cost.}

\maketitle


Glaucoma is a leading cause of irreversible blindness worldwide, and it often stays silent until vision has already been lost~\cite{tham2014}. Early screening is the main way to prevent this. The diagnosis rests on measuring optic nerve head features such as the cup-to-disc ratio, the thickness of the rim around the cup, and the rim area. However, screening relies on a specialist assessing the optic nerve head in retinal imaging such as a fundus photograph, and specialists are scarce where screening is needed most.

Vision-language models, or VLMs, connect images with text and have shown promise for building systems that work across diagnostic tasks without a labeled dataset for every condition~\cite{radford2021clip,silva2025flair}. Such a model could describe the measurements required to diagnose glaucoma if it were trained to do so, but that needs paired image-text data containing that information. Medical VLMs usually learn from electronic health reports, as ConVIRT, GLoRIA, MedKLIP, KAD and MedCLIP do with radiology reports~\cite{zhang2020convirt,huang2021gloria,wu2023medklip,zhang2023kad,wang2022medclip}, but such text data are scarce for training domain-specific models on glaucoma. Ophthalmic VLMs have adapted in different ways to pair text with ophthalmic images. FLAIR pairs fundus images with descriptive prompts written for each disease category~\cite{silva2025flair}, KeepFIT adds image-text pairs from fundus atlas books~\cite{wu2024mmretinal}, EyeCLIP combines images from 11 modalities with partial clinical text~\cite{shi2025eyeclip}, and UniMed-CLIP uses a language model to turn labels into captions~\cite{khattak2024unimedclip}. For all of these models, the text describes a category in general. The one public glaucoma dataset that pairs images with clinical notes holds 10,000 scanning laser ophthalmoscopy images with free-text summaries of the diagnosis~\cite{luo2024fairclip}. To our knowledge, no public text dataset or electronic health report describes the individual optic nerve head measurements of each image, and hence none of these models can read them.

We hypothesize that this text can be computed from segmentation masks. Several public glaucoma datasets include expert segmentation masks of the optic disc and cup~\cite{orlando2020refuge,zhang2010origa,kumar2023chaksu,kovalyk2022papila,fang2022refuge2}, and these masks contain the measurements a clinician reads. Although these masks were originally intended for segmentation tasks, in this study we propose GlaucomaMarker, a framework that repurposes and processes them to train VLMs. It measures disease-specific biomarkers from the mask pixels, splits each measurement into clinically meaningful categories, and writes each category as a template sentence, which gives text that is specific to each image. Every sentence can be traced back to the pixels it was measured from, and our measurement of the vertical cup-to-disc ratio agreed with independent expert grading on 650 images, with a mean error of 0.0012~\cite{zhang2010origa}. Unlike generated captions, this text does not depend on a language model reading the image, and unlike reports, it does not need new expert-written texts, only masks that already exist.

\begin{figure}[htbp]
\centering
\includegraphics[width=\textwidth]{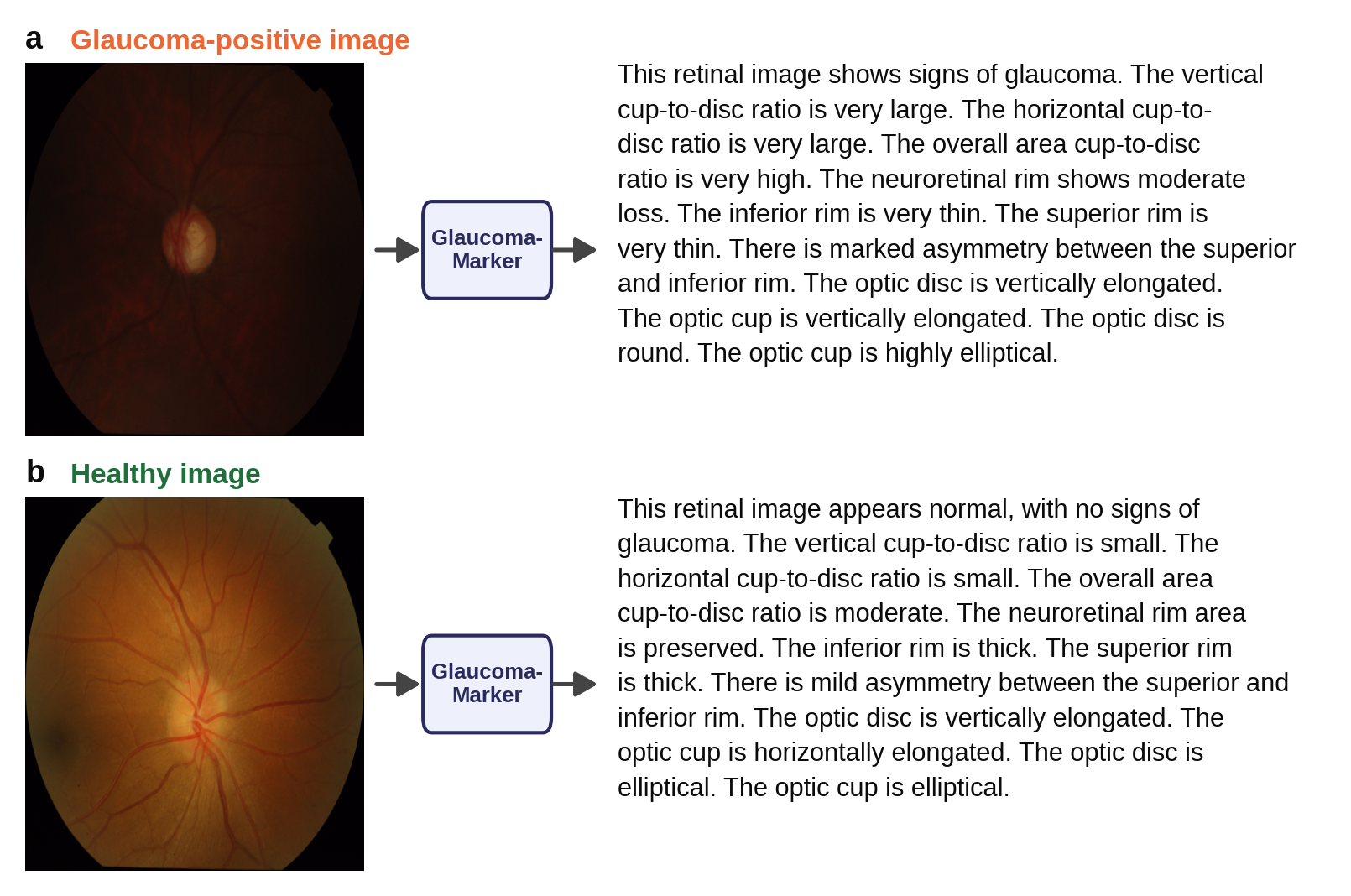}
\caption{\textbf{Zero-shot output of GlaucomaMarker on two fundus images.} \textbf{a} A glaucoma-positive image from Drishti-GS and the sentences chosen by GlaucomaMarker-R, one disease sentence and one sentence for each of the eleven biomarkers. \textbf{b} The same for a healthy image. The sentences are the highest-scoring entries of the 48-sentence catalog, from the RETFound-initialized instance with seed 42.}
\label{fig:usecase_example}
\end{figure}

GlaucomaMarker has three stages. First, it computes eleven biomarkers from each mask, such as the vertical and horizontal cup-to-disc ratio and the rim thickness, and maps each one to a sentence from a catalog of 48, which pairs every image with text that describes it. Second, it trains a dual-encoder vision-language model on these image-text pairs with a contrastive loss and an epoch sampler that balance the healthy and glaucoma classes and the biomarker categories, because public glaucoma data are skewed toward healthy eyes and common findings. Third, it reads out the diagnosis and the biomarkers zero-shot, by comparing the embedding of an image with the embeddings of the sentences, and Figure~\ref{fig:usecase_example} shows this output on two fundus images. We applied the framework to two retinal models, RETFound, which gave GlaucomaMarker-R, and FLAIR, which gave GlaucomaMarker-F. We trained on a combined dataset of 9,778 images, 3,238 of them with masks, that yielded 45,190 image-text pairs, and we evaluated on two external cohorts, the public Drishti-GS benchmark and a real-world clinical cohort from Wake Forest University School of Medicine.

On the clinical cohort, GlaucomaMarker-F reaches an AUROC of 0.905 for glaucoma, against 0.848 for FLAIR, and the highest Matthews correlation coefficient (MCC) of all models compared, and both instances read the biomarkers where the five vision-language baselines stay near random guessing. Controlled three-seed ablations show that the biomarker text produces the biomarker reading of GlaucomaMarker-R and improves the cup-to-disc readings of GlaucomaMarker-F, that the effect of the starting model differs between the two instances, and that the epoch sampler removes the preference for common buckets without improving the external results. Our contributions are fourfold. First, we introduce a framework that computes biomarker sentences from public segmentation masks to train retinal VLMs. Second, we show that it teaches both RETFound and FLAIR, two well-known retinal foundation models, to read glaucoma biomarkers zero-shot. Third, we counter the skew of public glaucoma data with an epoch sampler that balances the biomarker categories, and we show that it removes the model's preference for common buckets. Fourth, we evaluate it against seven baselines on two external cohorts, where GlaucomaMarker-F leads in diagnosis on the clinical cohort and both instances lead in biomarker reading, and our ablations show what drives these gains.

\section*{Results}

In this section, we first describe the GlaucomaMarker framework, and then the training data, the evaluation cohorts and the baselines. We then test whether the framework improves diagnosis, whether it reads biomarkers that the baselines cannot, whether it improves the embedding space of the retinal models it starts from, and whether it performs similarly across demographic subgroups. Finally, ablations show which design choices produce these results.

\subsection*{GlaucomaMarker computes biomarker-rich text from masks and teaches retinal models to read it}
\label{sec:results-framework}

\begin{figure}[!htbp]
\centering
\includegraphics[width=\textwidth]{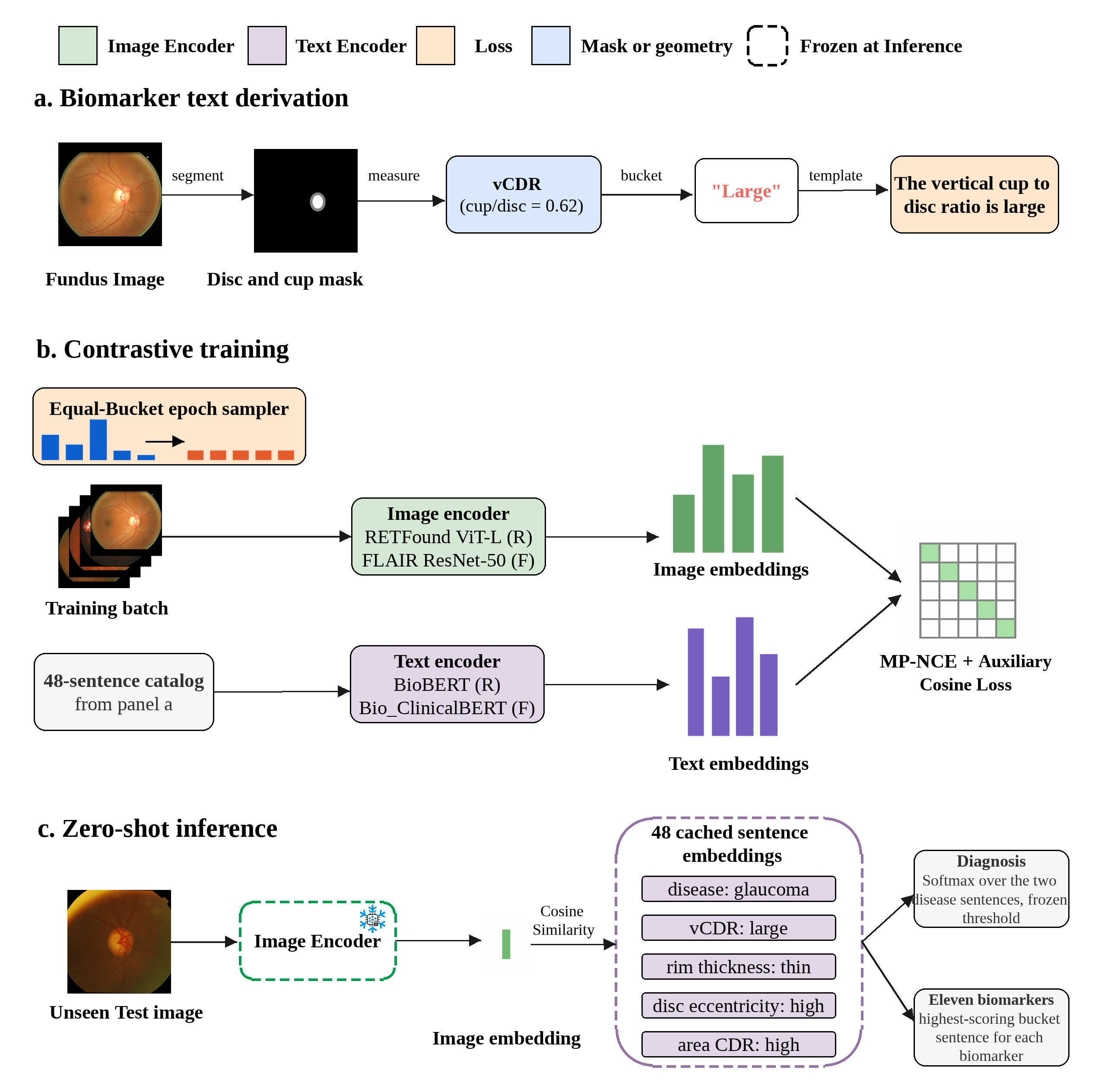}
\caption{\textbf{The GlaucomaMarker framework.} \textbf{a} Eleven biomarkers are measured from the disc and cup mask of each image, assigned to buckets with predefined cutoffs as per clinical descriptions, and written as template sentences from a catalog of 48. \textbf{b} A dual-encoder model is trained on batches that the equal-bucket epoch sampler draws, with the MP-NCE multi-positive contrastive loss plus an auxiliary cosine loss. The image encoder is RETFound ViT-L in GlaucomaMarker-R (R) and FLAIR ResNet-50 in GlaucomaMarker-F (F), and the text encoder is BioBERT in R and Bio\_ClinicalBERT in F. \textbf{c} At test time, the frozen model compares the embedding of an unseen image with 48 cached sentence embeddings. Diagnosis is a softmax over the two disease sentences with a frozen threshold, and each biomarker is the highest-scoring bucket sentence. Green marks the image pathway, purple the text pathway, orange the losses, blue the mask, and dashed outlines the parts that are frozen.}
\label{fig:architecture_schematic}
\end{figure}

GlaucomaMarker has three stages, shown in Figure~\ref{fig:architecture_schematic}. In the first stage, it measures eleven biomarkers from the optic disc and cup mask of an image, such as the vertical cup-to-disc ratio and the rim thickness, and assigns each measurement to one of three to five buckets with predefined cutoffs as per clinical descriptions. Each bucket maps to a template sentence from a fixed catalog of 48, so an image with a mask is paired with up to eleven sentences that describe it, plus a disease sentence when its label is known. In the second stage, a dual-encoder vision-language model is trained to match each image with its sentences, using a multi-positive contrastive loss, MP-NCE, that treats every sentence of the same biomarker and bucket as a match, together with an auxiliary cosine loss that draws each image toward the sentences it carries. In the third stage, the trained model reads an image that has no mask and no label, by comparing the embedding of the image with the embeddings of the 48 sentences.

The model never writes free text. Every output is one of the 48 sentences, so the diagnosis is a choice between two disease sentences and each biomarker is a choice among its three to five bucket sentences. Figure~\ref{fig:usecase_example} shows this output for two fundus images. Because the output is a fixed sentence, we can score it against a measured value, and no clinician is needed to evaluate it further.

We applied the framework to two retinal models, and Table~\ref{tab:instances} lists their parts. GlaucomaMarker-R starts from RETFound, with a ViT-L image encoder and a BioBERT text encoder, and GlaucomaMarker-F starts from FLAIR, with a ResNet-50 image encoder and a Bio\_ClinicalBERT text encoder. Both were trained for 20 epochs on the same image-sentence pairs, with three seeds each, and the checkpoint of each seed was chosen on an internal validation set. The rest of the pipeline is identical, so differences between the two instances come from the starting model.

Public glaucoma data are skewed toward healthy eyes and common findings, so each training epoch draws an equal number of sentences from every well-supported bucket. Methods describes the sampler and the loss.

\subsection*{GlaucomaMarker is trained on public data and evaluated on two cohorts that played no part in training}
\label{sec:results-setup}

We trained on six public datasets, namely AIROGS \citep{devente2024airogs}, Chaksu \citep{kumar2023chaksu}, ORIGA \citep{zhang2010origa}, PAPILA \citep{kovalyk2022papila}, REFUGE \citep{orlando2020refuge} and REFUGE2 \citep{fang2022refuge2}, which gave 9,778 images and 45,190 image-sentence pairs. Table~\ref{tab:dataset_composition_sources} lists the composition by source and Methods describes the data and the validation split.

\begin{figure}[htbp]
\centering
\includegraphics[width=\textwidth]{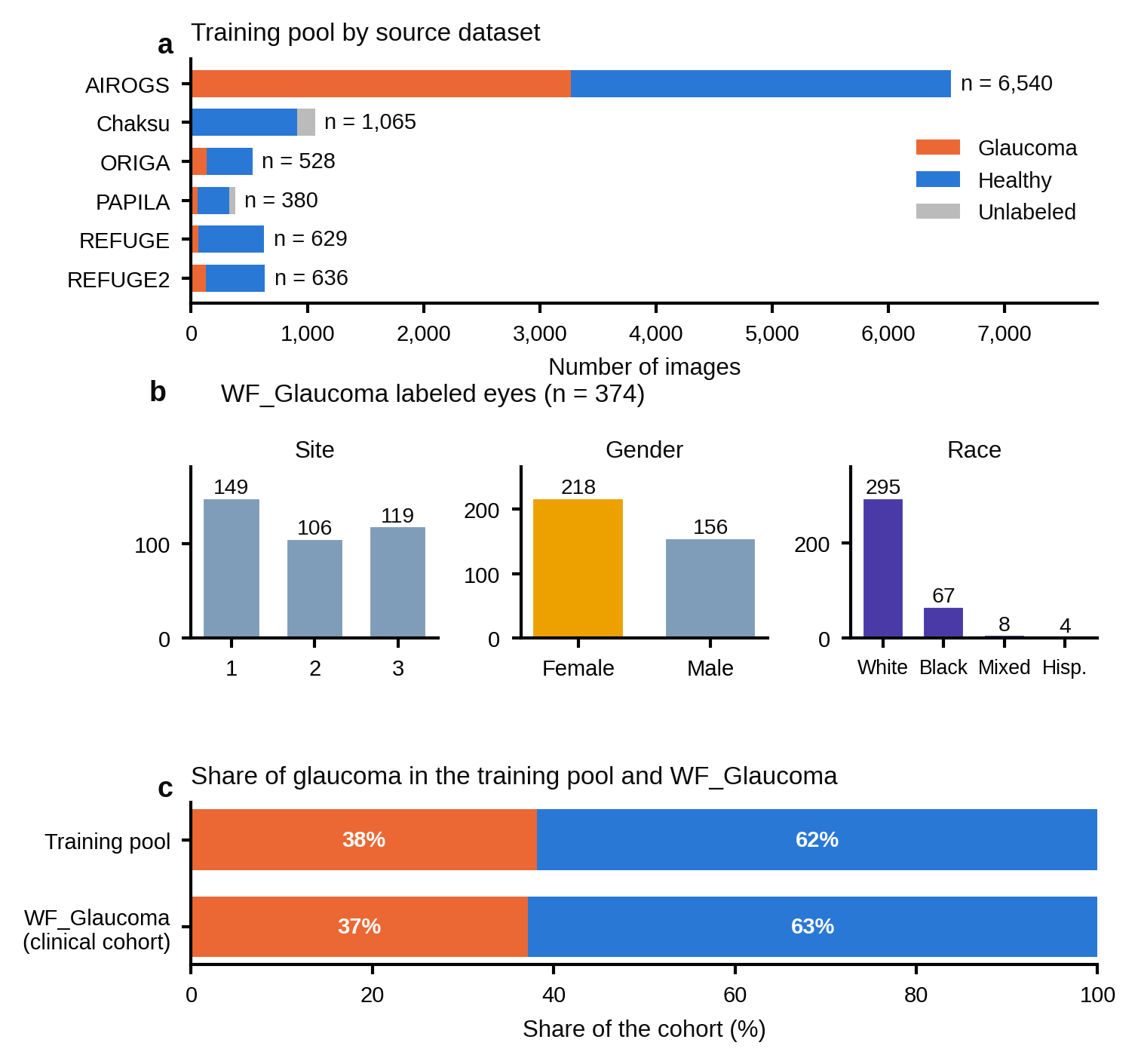}
\caption{\textbf{Data used for training and evaluation.} \textbf{a} The training pool by source dataset, with glaucoma (orange), healthy (blue) and unlabeled (gray) images. \textbf{b} Site, gender and race of the 374 labeled eyes of the WF\_Glaucoma clinical cohort from Wake Forest University School of Medicine. \textbf{c} The share of glaucoma and healthy images in the training pool and in WF\_Glaucoma, which were 38 and 37 percent glaucoma.}
\label{fig:dataset_composition_demographics}
\end{figure}

We evaluated on two cohorts that played no part in training, threshold selection or checkpoint selection. The first, Drishti-GS \citep{sivaswamy2014drishti}, is a public benchmark of 101 images, 70 glaucoma and 31 healthy, with expert disc and cup segmentation. The second is a clinical cohort from Wake Forest University School of Medicine, which we call WF\_Glaucoma, with 374 labeled eyes, 139 glaucoma and 235 healthy, as shown in Table~\ref{tab:dataset_composition_demographics} and Figure~\ref{fig:dataset_composition_demographics}. Methods describes both cohorts.

\begin{table}[htbp]
\centering
\small
\caption{Demographic and site composition of the WF\_Glaucoma clinical evaluation cohort (Wake Forest University School of Medicine). The table gives the number of images used for each analysis, and site, gender, race, and age describe the 374 labeled eyes.}
\label{tab:dataset_composition_demographics}
\resizebox{\ifdim\width>\linewidth\linewidth\else\width\fi}{!}{%
\begin{tabular}{lc}
\toprule
Field & Value \\
\midrule
Photographs taken (patients) & 448 (225) \\
Without a diagnosis label & 74 \\
Labeled eyes, used for diagnosis, the embedding and fairness (patients) & 374 (198) \\
\quad Glaucoma & 139 \\
\quad Healthy & 235 \\
Images with OCT measurements, used for biomarker reading and retrieval (patients) & 444 (223) \\
\quad Of these, with a diagnosis label, used for the within-class correlation (patients) & 370 (196) \\
Without OCT measurements (all have a diagnosis label) & 4 \\
Imaging device & Topcon Maestro \\
\addlinespace
Site 1 / 2 / 3, labeled eyes & 149 / 106 / 119 \\
Gender Female / Male, labeled eyes & 218 / 156 \\
Race White / Black / Mixed / Hispanic, labeled eyes & 295 / 67 / 8 / 4 \\
Age range (mean), years, labeled eyes & 35.8--94.0 (75.9) \\
\botrule
\end{tabular}}
\end{table}

We compared both instances with seven publicly released baselines, five vision-language models and two supervised glaucoma classifiers. Methods describes them. FLAIR, UniMed-CLIP, KeepFIT and GlaucomaMarker-F are reported on WF\_Glaucoma only, because their pretraining overlaps with Drishti-GS.

For the disease, the ground truth is the clinician diagnosis recorded for each cohort. For each biomarker, the ground truth is a measurement that we turned into a bucket with the same cutoffs as the training text. On Drishti-GS the measurements come from the expert disc and cup masks, processed by the same pipeline as the training text, so this checks that the model learned our definitions. On WF\_Glaucoma they come from the OCT scan measurements made by clinicians, which cover five of the eleven biomarkers. As an independent check of the measurements themselves, our vertical cup-to-disc ratio agreed with the ORIGA expert grading on all 650 images, with a mean error of 0.0012.

We report AUROC and AUPRC as threshold-free ranking metrics, and accuracy, balanced accuracy, sensitivity, specificity, F1, MCC and Cohen's kappa at a single operating threshold chosen on the internal validation split. Every GlaucomaMarker result is the mean over three seeds, while baselines are single runs. Methods describes the metrics, the bootstrap intervals and the tests, and the comparisons that were reported with an interval and no test are listed in Supplementary Table~\ref{tab:statistical_comparisons}.

\subsection*{GlaucomaMarker-F diagnoses glaucoma more accurately than FLAIR on a clinical cohort}
\label{sec:results-zeroshot}

\begin{figure}[htbp]
\centering
\includegraphics[width=\textwidth]{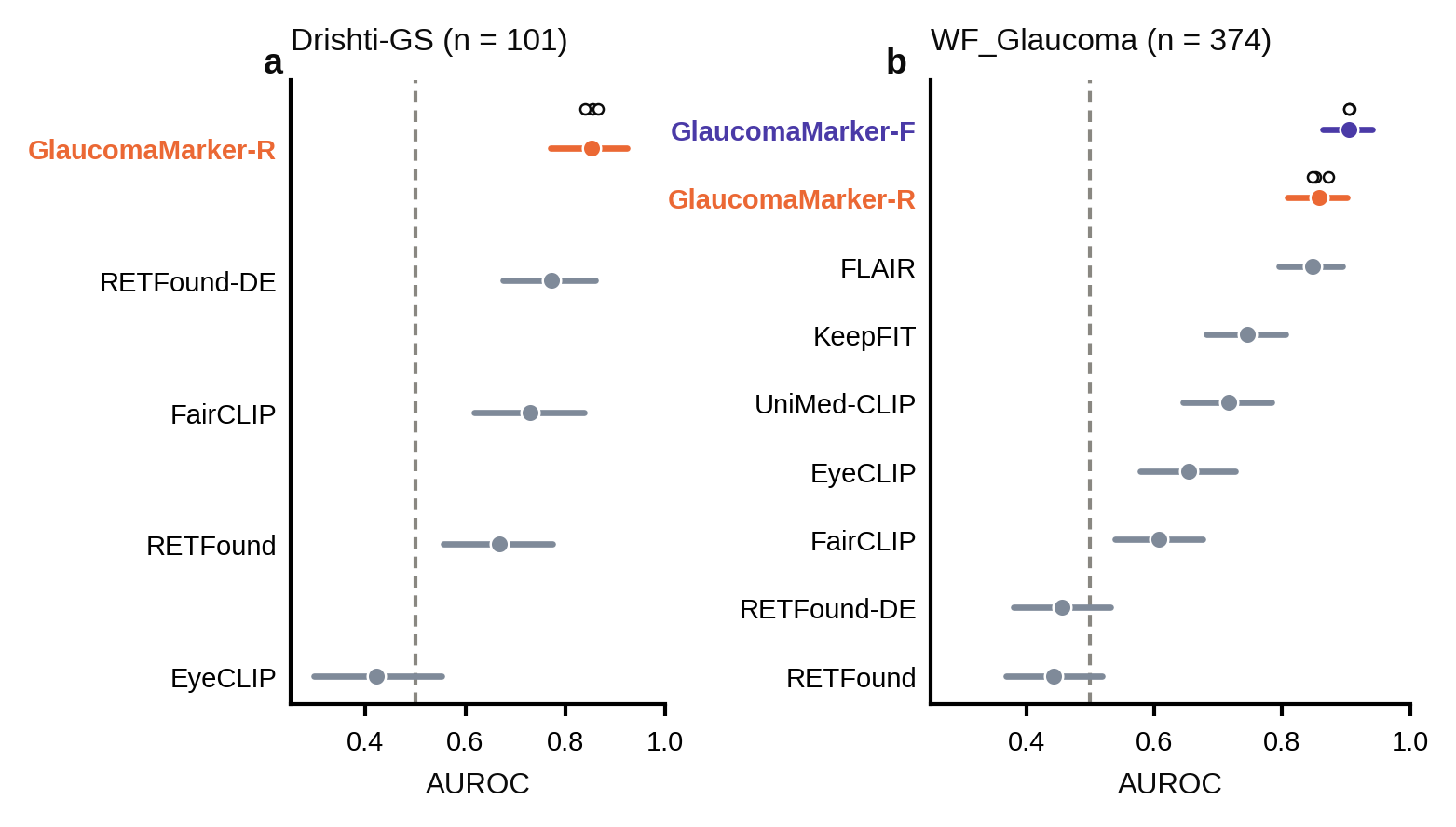}
\caption{\textbf{Zero-shot AUROC of every model.} \textbf{a} Drishti-GS. \textbf{b} WF\_Glaucoma. Filled circles are the AUROC and lines are 95\% bootstrap confidence intervals. Orange is GlaucomaMarker-R and purple is GlaucomaMarker-F, with the open circles above the filled circle showing the three seeds. Gray marks the baselines, which are single runs, and the dashed line marks an AUROC of 0.5. Panel \textbf{a} omits models whose pretraining data overlap with Drishti-GS.}
\label{fig:zeroshot_classification_comparison}
\end{figure}

\begin{figure}[htbp]
\centering
\includegraphics[width=\textwidth]{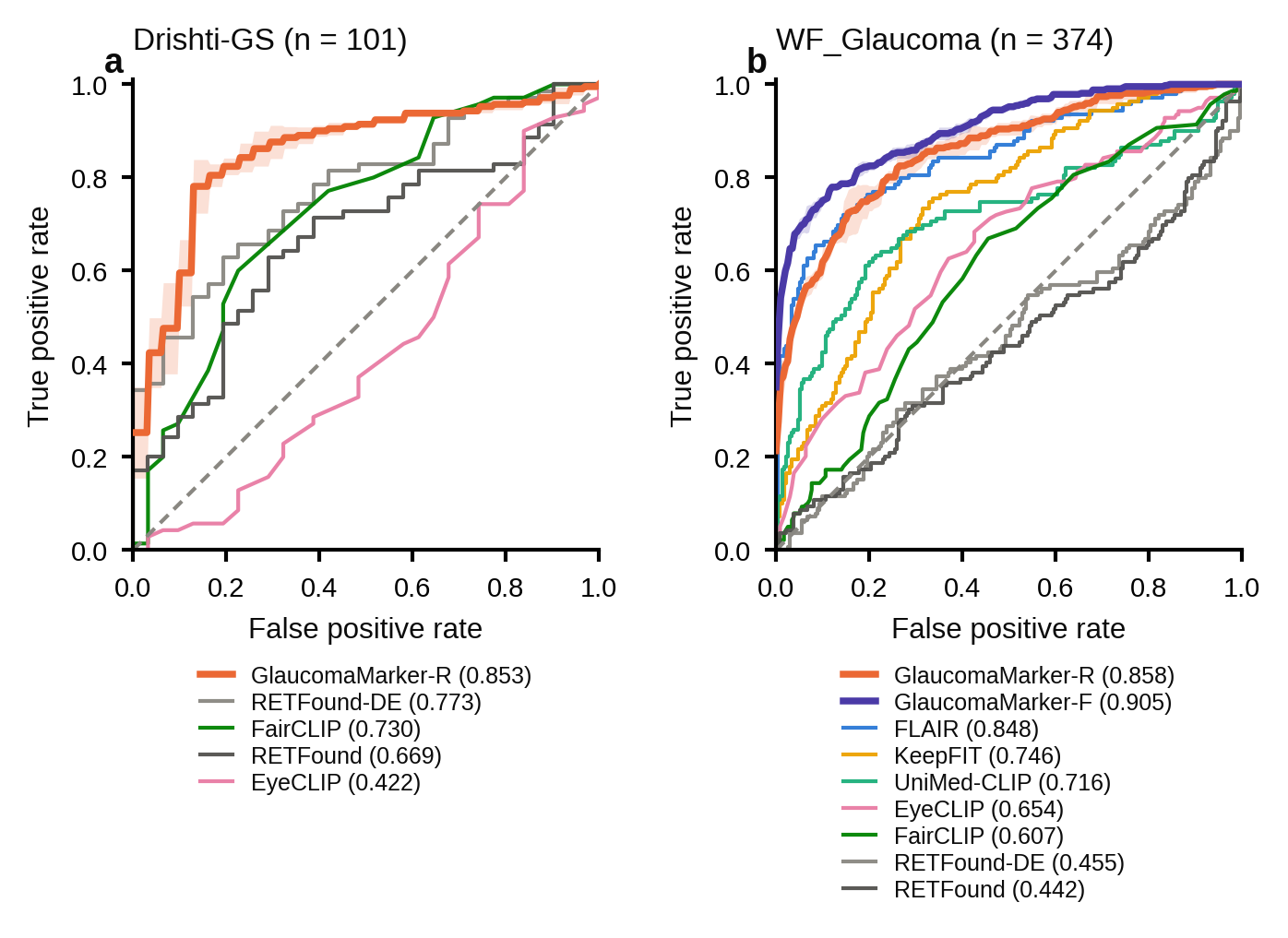}
\caption{\textbf{Zero-shot ROC curves.} \textbf{a} Drishti-GS. \textbf{b} WF\_Glaucoma. The thick lines are GlaucomaMarker-R (orange) and GlaucomaMarker-F (purple), as the mean over three seeds with a shaded band of one standard deviation of the true positive rate. Thin lines are baselines, which are single runs. The AUROC of each model is in the legend and the dashed diagonal is chance. Panel \textbf{a} omits models whose pretraining data overlap with Drishti-GS.}
\label{fig:roc_curves}
\end{figure}

Table~\ref{tab:main_zeroshot_results} and Figure~\ref{fig:zeroshot_classification_comparison} report zero-shot diagnosis on both cohorts, and Figure~\ref{fig:roc_curves} shows the ROC curves. On WF\_Glaucoma, GlaucomaMarker-F reached an AUROC of 0.905 and an MCC of 0.662, against 0.848 and 0.473 for FLAIR, the strongest baseline. The AUROC was higher by 0.057, with a 95 percent confidence interval of 0.023 to 0.093 and $P = 0.0014$, and the MCC was higher by 0.189, with an interval of 0.097 to 0.280 and $P < 0.0002$. After Holm adjustment these P values were 0.0042 and 0.0008. Table~\ref{tab:statistical_tests} lists the tested comparisons with their exact P values. No other baseline went above an AUROC of 0.746.

GlaucomaMarker-R reached an AUROC of 0.858 and an MCC of 0.563 on WF\_Glaucoma. Its AUROC was not distinguishable from that of FLAIR, with a difference of 0.010, an interval of -0.038 to 0.060 and $P = 0.69$. Its MCC was higher by 0.090, but the interval of -0.007 to 0.186 included zero, with $P = 0.072$ and a Holm-adjusted $P = 0.14$. We therefore read GlaucomaMarker-R as comparable to FLAIR on this cohort. Outside the primary comparisons, GlaucomaMarker-F was higher than GlaucomaMarker-R by 0.047 in AUROC, with an interval of 0.008 to 0.088, and by 0.099 in MCC, with an interval of 0.012 to 0.186.

On Drishti-GS, GlaucomaMarker-R reached an AUROC of 0.853 and an MCC of 0.536, against AUROCs of 0.422 to 0.773 for the four baselines that remain after the exclusion described above. Against RETFound-DE, the strongest of these on AUROC, its MCC was higher by 0.354, with an interval of 0.203 to 0.503. The AUROC difference of 0.080 had an interval of -0.005 to 0.171, so the lead in ranking is not established on this small cohort of 101 images.

\begin{figure}[htbp]
\centering
\includegraphics[width=\textwidth]{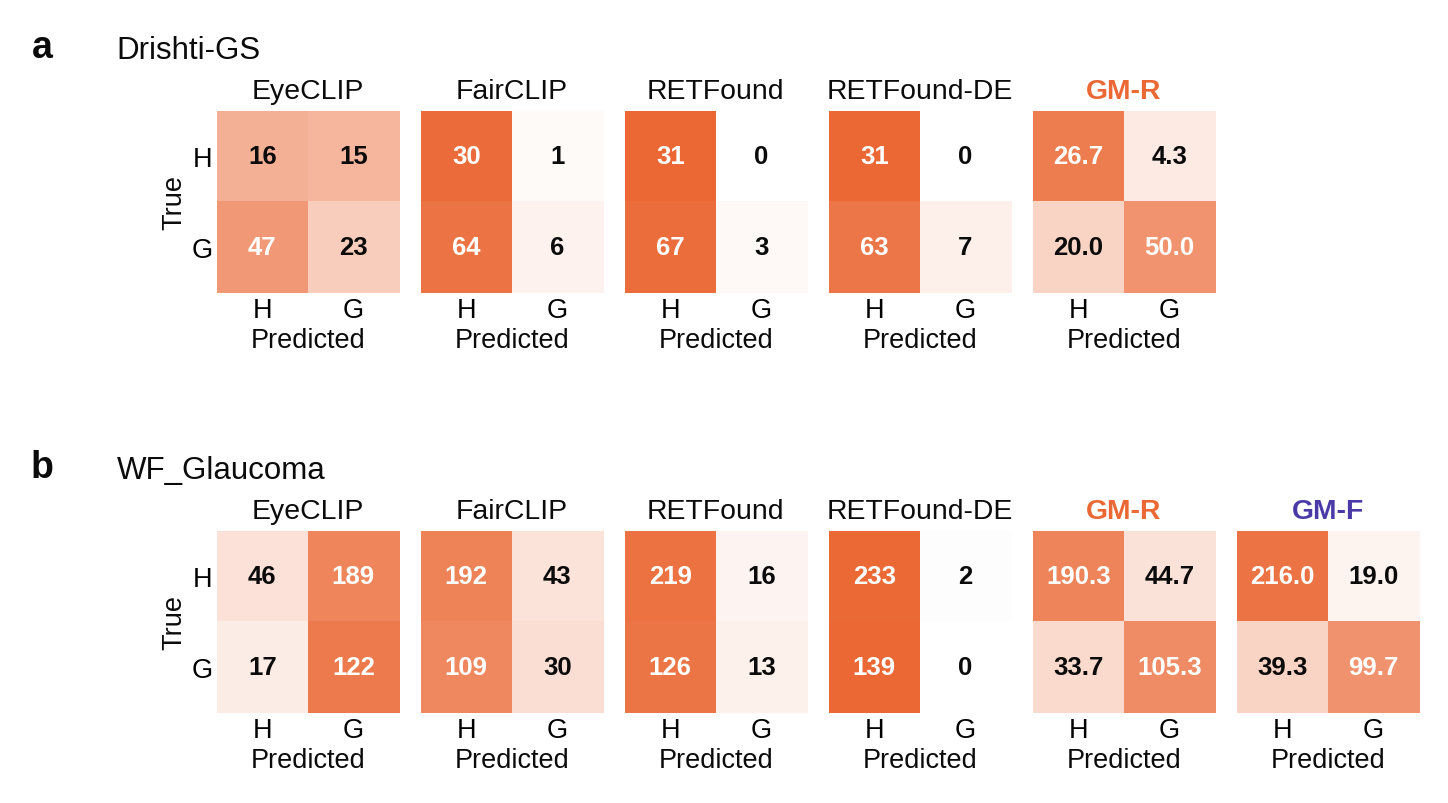}
\caption{\textbf{Confusion matrices at the frozen decision threshold.} \textbf{a} Drishti-GS. \textbf{b} WF\_Glaucoma. Rows are true classes and columns are predicted classes, with H for healthy and G for glaucoma. Cell color shows the fraction of the true class and the number is the image count. GM-R and GM-F are GlaucomaMarker-R and GlaucomaMarker-F, with counts averaged over three seeds, so they are not integers. Models whose pretraining data overlap with Drishti-GS are omitted from panel \textbf{a}. Supplementary Table~\ref{tab:confusion_matrices} gives all models with their thresholds.}
\label{fig:confusion_matrices}
\end{figure}

A high specificity alone did not signal a better classifier. On Drishti-GS, RETFound and RETFound-DE had a specificity of 1.000 but a sensitivity of only 0.043 and 0.100. On WF\_Glaucoma, UniMed-CLIP and RETFound-DE had a specificity of 0.996 and 0.992, with a sensitivity of 0.043 and 0.000. These models predicted almost no glaucoma cases, and Figure~\ref{fig:confusion_matrices} and Supplementary Table~\ref{tab:confusion_matrices} show this directly. We therefore report specificity next to sensitivity throughout.

GlaucomaMarker-F and GlaucomaMarker-R differ in where they place their operating point. On WF\_Glaucoma, GlaucomaMarker-F had the higher specificity, 0.919 against 0.810, and GlaucomaMarker-R had the higher sensitivity, 0.758 against 0.717. Table~\ref{tab:glaucomarker_full_diagnostic_metrics} gives the full set of metrics for both instances.

\begin{table}[htbp]
\centering
\small
\caption{Zero-shot classification performance of GlaucomaMarker and the external baselines on Drishti-GS and WF\_Glaucoma. AUROC and MCC are given with 95\% bootstrap confidence intervals, and the other metrics are the mean $\pm$ standard deviation over three seeds for GlaucomaMarker and a single value for the baselines. Models whose pretraining data overlap with Drishti-GS, including GlaucomaMarker-F, are excluded from that panel. Best value per column and panel in bold.}
\label{tab:main_zeroshot_results}
\resizebox{\ifdim\width>\linewidth\linewidth\else\width\fi}{!}{%
\begin{tabular}{lcccccc}
\toprule
Model & AUROC (95\% CI) & Bal.\ Acc. & Sens. & Spec. & Macro F1 & MCC (95\% CI) \\
\midrule
\multicolumn{7}{l}{\textit{Drishti-GS (n=101)}} \\
EyeCLIP & 0.422 (0.297 to 0.553) & 0.422 & 0.329 & 0.516 & 0.383 & $-$0.148 ($-$0.346 to 0.049) \\
FairCLIP & 0.730 (0.618 to 0.837) & 0.527 & 0.086 & 0.968 & 0.318 & 0.097 ($-$0.087 to 0.218) \\
RETFound & 0.669 (0.556 to 0.774) & 0.521 & 0.043 & \textbf{1.000} & 0.281 & 0.116 (0.000 to 0.182) \\
RETFound-DE & 0.773 (0.676 to 0.861) & 0.550 & 0.100 & \textbf{1.000} & 0.339 & 0.182 (0.103 to 0.255) \\
\textbf{GlaucomaMarker-R} & \textbf{0.853 (0.771 to 0.923)} & \textbf{0.787{\scriptsize$\pm$0.051}} & \textbf{0.714{\scriptsize$\pm$0.100}} & 0.860{\scriptsize$\pm$0.019} & \textbf{0.746{\scriptsize$\pm$0.066}} & \textbf{0.536 (0.383 to 0.674)} \\
\addlinespace
\multicolumn{7}{l}{\textit{WF\_Glaucoma (n=374)}} \\
FLAIR & 0.848 (0.795 to 0.894) & 0.744 & 0.842 & 0.647 & 0.717 & 0.473 (0.377 to 0.564) \\
UniMed-CLIP & 0.716 (0.645 to 0.783) & 0.519 & 0.043 & \textbf{0.996} & 0.430 & 0.139 (0.036 to 0.215) \\
EyeCLIP & 0.654 (0.578 to 0.726) & 0.537 & \textbf{0.878} & 0.196 & 0.425 & 0.095 ($-$0.026 to 0.207) \\
FairCLIP & 0.607 (0.538 to 0.676) & 0.516 & 0.216 & 0.817 & 0.500 & 0.040 ($-$0.080 to 0.160) \\
RETFound & 0.442 (0.368 to 0.518) & 0.513 & 0.093 & 0.932 & 0.455 & 0.046 ($-$0.076 to 0.170) \\
RETFound-DE & 0.455 (0.380 to 0.531) & 0.496 & 0.000 & 0.992 & 0.384 & $-$0.056 ($-$0.093 to 0.000) \\
KeepFIT (MM-Retinal) & 0.746 (0.682 to 0.806) & 0.592 & 0.252 & 0.932 & 0.577 & 0.259 (0.143 to 0.369) \\
\textbf{GlaucomaMarker-R} & 0.858 (0.808 to 0.901) & 0.784{\scriptsize$\pm$0.020} & 0.758{\scriptsize$\pm$0.047} & 0.810{\scriptsize$\pm$0.071} & 0.779{\scriptsize$\pm$0.028} & 0.563 (0.469 to 0.651) \\
\textbf{GlaucomaMarker-F} & \textbf{0.905 (0.864 to 0.940)} & \textbf{0.818{\scriptsize$\pm$0.003}} & 0.717{\scriptsize$\pm$0.029} & 0.919{\scriptsize$\pm$0.034} & \textbf{0.827{\scriptsize$\pm$0.008}} & \textbf{0.662 (0.571 to 0.748)} \\
\botrule
\end{tabular}}
\end{table}

\begin{table}[htbp]
\centering
\footnotesize
\caption{Full diagnostic metrics for the two GlaucomaMarker instances on each external cohort (mean $\pm$ standard deviation over three seeds), including positive and negative predictive value and Cohen's $\kappa$, at the validation-frozen threshold. F1 is the F1 score of the glaucoma class. The FLAIR-initialized model is omitted on Drishti-GS because FLAIR was pretrained on that dataset.}
\label{tab:glaucomarker_full_diagnostic_metrics}
\resizebox{\ifdim\width>\linewidth\linewidth\else\width\fi}{!}{%
\begin{tabular}{llcccccccccc}
\toprule
Model & Cohort & AUROC & AUPRC & Acc. & F1 & Sens. & Spec. & PPV & NPV & MCC & $\kappa$ \\
\midrule
\textbf{GlaucomaMarker-R} & Drishti-GS & 0.853{\scriptsize$\pm$0.013} & 0.927{\scriptsize$\pm$0.013} & 0.759{\scriptsize$\pm$0.070} & 0.802{\scriptsize$\pm$0.068} & 0.714{\scriptsize$\pm$0.100} & 0.860{\scriptsize$\pm$0.019} & 0.919{\scriptsize$\pm$0.013} & 0.580{\scriptsize$\pm$0.088} & 0.536{\scriptsize$\pm$0.101} & 0.508{\scriptsize$\pm$0.117} \\
\textbf{GlaucomaMarker-R} & WF\_Glaucoma & 0.858{\scriptsize$\pm$0.013} & 0.817{\scriptsize$\pm$0.019} & 0.791{\scriptsize$\pm$0.031} & 0.730{\scriptsize$\pm$0.023} & 0.758{\scriptsize$\pm$0.047} & 0.810{\scriptsize$\pm$0.071} & 0.708{\scriptsize$\pm$0.063} & 0.851{\scriptsize$\pm$0.016} & 0.563{\scriptsize$\pm$0.047} & 0.560{\scriptsize$\pm$0.052} \\
\textbf{GlaucomaMarker-F} & WF\_Glaucoma & 0.905{\scriptsize$\pm$0.001} & 0.885{\scriptsize$\pm$0.002} & 0.844{\scriptsize$\pm$0.011} & 0.774{\scriptsize$\pm$0.006} & 0.717{\scriptsize$\pm$0.029} & 0.919{\scriptsize$\pm$0.034} & 0.843{\scriptsize$\pm$0.052} & 0.846{\scriptsize$\pm$0.009} & 0.662{\scriptsize$\pm$0.024} & 0.656{\scriptsize$\pm$0.018} \\
\botrule
\end{tabular}}
\end{table}

\begin{table}[htbp]
\centering
\footnotesize
\caption{Tested comparisons and their Holm-adjusted $P$ values. The primary family has four tests and the biomarker family six, as set out in Methods under Statistical analysis. All tests are two-sided with $\alpha = 0.05$. A $P$ value below 0.0002 is the resolution limit of the resampling, and the interval is the 95\% bootstrap interval of the difference.}
\label{tab:statistical_tests}
\resizebox{\ifdim\width>\linewidth\linewidth\else\width\fi}{!}{%
\begin{tabular}{llllllc}
\toprule
Cohort & Comparison & Metric & Difference & 95\% CI & $P$ & Holm-adjusted $P$ \\
\midrule
\multicolumn{7}{l}{\textit{Primary family, diagnosis}} \\
WF\_Glaucoma & GlaucomaMarker-F minus FLAIR & AUROC & 0.057 & 0.023 to 0.093 & 0.0014 & 0.0042 \\
WF\_Glaucoma & GlaucomaMarker-F minus FLAIR & MCC & 0.189 & 0.097 to 0.280 & below 0.0002 & 0.0008 \\
WF\_Glaucoma & GlaucomaMarker-R minus FLAIR & AUROC & 0.010 & $-$0.038 to 0.060 & 0.69 & 0.69 \\
WF\_Glaucoma & GlaucomaMarker-R minus FLAIR & MCC & 0.090 & $-$0.007 to 0.186 & 0.072 & 0.14 \\
\addlinespace
\multicolumn{7}{l}{\textit{Biomarker family}} \\
WF\_Glaucoma & GlaucomaMarker-R against the random guessing level & Mean balanced accuracy & 0.098 & 0.074 to 0.120 & below 0.0002 & 0.0012 \\
WF\_Glaucoma & GlaucomaMarker-R minus KeepFIT & Mean balanced accuracy & 0.069 & 0.036 to 0.103 & below 0.0002 & 0.0012 \\
WF\_Glaucoma & GlaucomaMarker-F against the random guessing level & Mean balanced accuracy & 0.076 & 0.045 to 0.106 & below 0.0002 & 0.0012 \\
WF\_Glaucoma & GlaucomaMarker-F minus KeepFIT & Mean balanced accuracy & 0.047 & 0.006 to 0.088 & 0.025 & 0.025 \\
Drishti-GS & GlaucomaMarker-R against the random guessing level & Mean balanced accuracy & 0.128 & 0.069 to 0.152 & below 0.0002 & 0.0012 \\
Drishti-GS & GlaucomaMarker-R minus EyeCLIP & Mean balanced accuracy & 0.159 & 0.075 to 0.185 & below 0.0002 & 0.0012 \\
\botrule
\end{tabular}}
\end{table}

\subsection*{Both instances read glaucoma biomarkers where the baselines stay near random guessing}
\label{sec:results-concept}

The central premise of GlaucomaMarker is that a model should recover the measurements it was trained to describe, in addition to classifying the disease. We tested this by scoring the zero-shot similarity of an image against the bucket sentences of each biomarker, on images whose masks the model never saw, and comparing the predicted bucket with the ground truth. We report balanced accuracy, the mean over the buckets present in the cohort of the share of images in that bucket that were predicted correctly. A model that chooses a bucket without using the image is correct for one bucket in every K on average, where K is the number of buckets present, so we call 1 over K the random guessing level. Averaged over the biomarkers of a cohort, it is 0.257 on WF\_Glaucoma and 0.286 on Drishti-GS. Only the five vision-language baselines have a biomarker readout.

\begin{figure}[htbp]
\centering
\includegraphics[width=\textwidth]{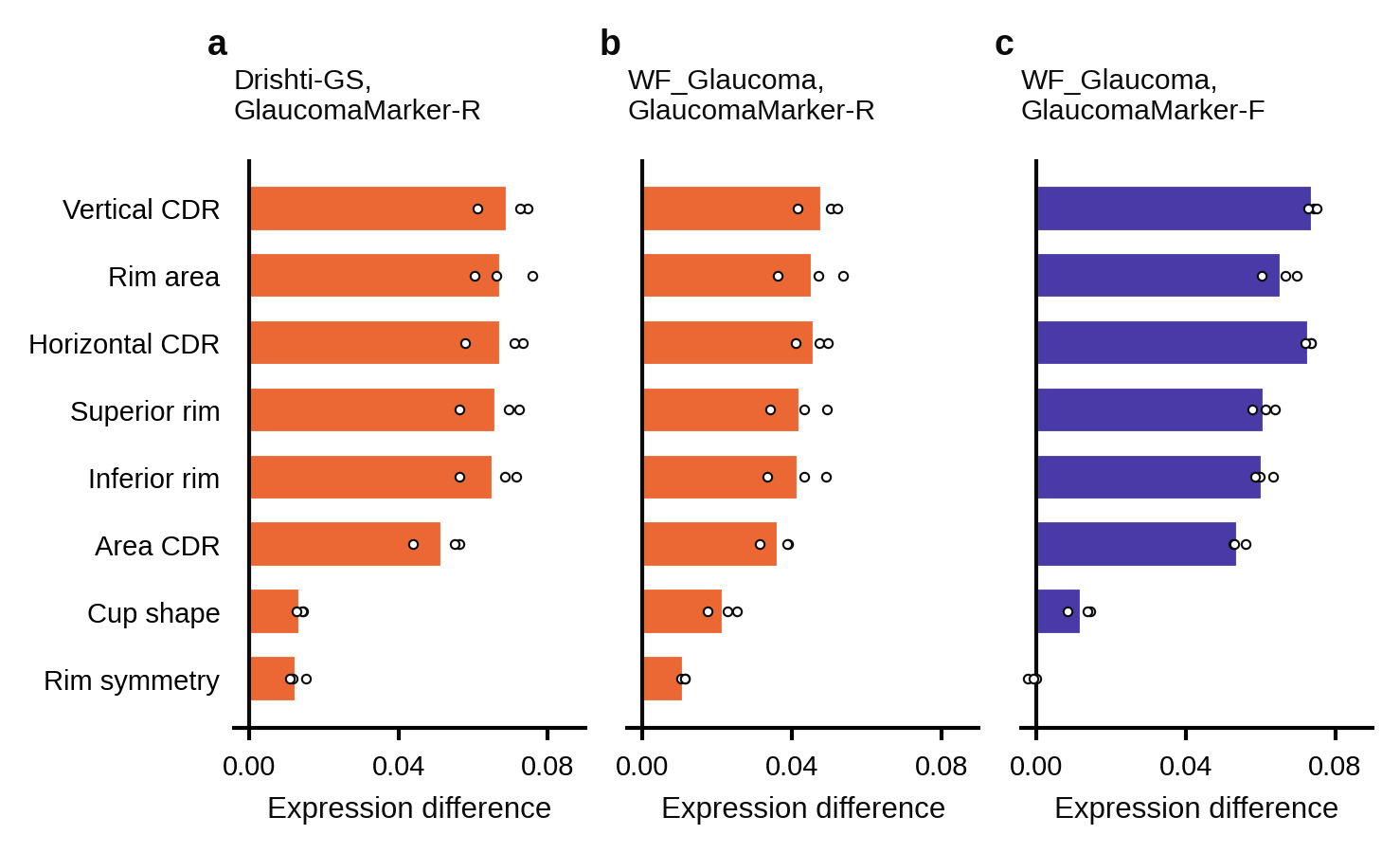}
\caption{\textbf{Disease-label-only expression of each biomarker.} Each bar is the difference in the mean zero-shot probability of the abnormal bucket between glaucomatous and healthy images, computed with disease labels only and no biomarker ground truth. \textbf{a} GlaucomaMarker-R on Drishti-GS. \textbf{b} GlaucomaMarker-R on WF\_Glaucoma. \textbf{c} GlaucomaMarker-F on WF\_Glaucoma, not shown on Drishti-GS because of pretraining overlap. Bars are the mean over three seeds and the open circles are the three seeds. CDR is the cup-to-disc ratio.}
\label{fig:biomarker_disease_correlation}
\end{figure}

On WF\_Glaucoma, the mean balanced accuracy over the five biomarkers with ground truth was 0.354 for GlaucomaMarker-R, with a 95 percent confidence interval of 0.331 to 0.377, and 0.332 for GlaucomaMarker-F, with an interval of 0.302 to 0.362. The best baseline, KeepFIT, reached 0.285, and the other four were between 0.251 and 0.266, close to the random guessing level. GlaucomaMarker-R was above KeepFIT by 0.069, with an interval of 0.036 to 0.103, and GlaucomaMarker-F by 0.047, with an interval of 0.006 to 0.088. On Drishti-GS, GlaucomaMarker-R reached 0.414 over all eleven biomarkers, with an interval of 0.356 to 0.438, against 0.255 for EyeCLIP and 0.250 for FairCLIP, the two baselines that remain after the exclusion. All six tests in this family, three against the random guessing level and three against the best baseline, had a Holm-adjusted $P$ of 0.025 or less, as Table~\ref{tab:statistical_tests} shows. Table~\ref{tab:concept_accuracy_ours_all_cohorts} gives the results for both instances, and Table~\ref{tab:concept_accuracy_baselines} shows the baselines. Using disease labels only, the zero-shot probability of the abnormal bucket of each biomarker was higher in glaucomatous than in healthy images for both instances, as Figure~\ref{fig:biomarker_disease_correlation} shows.

\begin{table}[htbp]
\centering
\footnotesize
\caption{Zero-shot biomarker balanced accuracy of the two GlaucomaMarker instances on the internal validation set and the two external cohorts, as the mean $\pm$ standard deviation over three seeds. The random guessing level is given as D / W for Drishti-GS and WF\_Glaucoma. The last row is the mean over the biomarkers, with 95\% bootstrap intervals for the external cohorts and the plain average for the internal set. Blank cells have no ground truth or are not evaluated for pretraining overlap.}
\label{tab:concept_accuracy_ours_all_cohorts}
\resizebox{\ifdim\width>\linewidth\linewidth\else\width\fi}{!}{%
\begin{tabular}{lcccccccc}
\toprule
& & & \multicolumn{3}{c}{GlaucomaMarker-R} & \multicolumn{2}{c}{GlaucomaMarker-F} \\
\cmidrule(lr){4-6}\cmidrule(lr){7-8}
Biomarker & \#classes & Random guess (D / W) & Drishti-GS & Internal & WF\_Glaucoma & Internal & WF\_Glaucoma \\
\midrule
Vertical CDR & 5 & 0.250 / 0.200 & 0.405{\scriptsize$\pm$0.048} & 0.680{\scriptsize$\pm$0.008} & 0.354{\scriptsize$\pm$0.024} & 0.673{\scriptsize$\pm$0.009} & 0.384{\scriptsize$\pm$0.005}  \\
Horizontal CDR & 5 & 0.250 / -- & 0.538{\scriptsize$\pm$0.035} & 0.619{\scriptsize$\pm$0.007} &  & 0.600{\scriptsize$\pm$0.007} &   \\
Area CDR & 4 & 0.333 / 0.250 & 0.627{\scriptsize$\pm$0.016} & 0.718{\scriptsize$\pm$0.010} & 0.360{\scriptsize$\pm$0.035} & 0.664{\scriptsize$\pm$0.025} & 0.385{\scriptsize$\pm$0.011}  \\
Rim area fraction & 4 & 0.250 / 0.250 & 0.546{\scriptsize$\pm$0.025} & 0.621{\scriptsize$\pm$0.012} & 0.427{\scriptsize$\pm$0.051} & 0.659{\scriptsize$\pm$0.061} & 0.466{\scriptsize$\pm$0.055}  \\
Rim thickness (inf.) & 5 & 0.200 / -- & 0.316{\scriptsize$\pm$0.023} & 0.543{\scriptsize$\pm$0.010} &  & 0.582{\scriptsize$\pm$0.011} &   \\
Rim thickness (sup.) & 5 & 0.200 / -- & 0.376{\scriptsize$\pm$0.011} & 0.576{\scriptsize$\pm$0.009} &  & 0.597{\scriptsize$\pm$0.006} &   \\
Rim vert. asymmetry & 4 & 0.250 / -- & 0.228{\scriptsize$\pm$0.017} & 0.351{\scriptsize$\pm$0.055} &  & 0.439{\scriptsize$\pm$0.033} &   \\
Disc V/H ratio & 3 & 0.500 / 0.333 & 0.302{\scriptsize$\pm$0.197} & 0.413{\scriptsize$\pm$0.040} & 0.338{\scriptsize$\pm$0.041} & 0.292{\scriptsize$\pm$0.034} & 0.196{\scriptsize$\pm$0.050}  \\
Cup V/H ratio & 3 & 0.333 / -- & 0.602{\scriptsize$\pm$0.015} & 0.494{\scriptsize$\pm$0.017} &  & 0.464{\scriptsize$\pm$0.011} &   \\
Disc eccentricity & 4 & 0.333 / 0.250 & 0.270{\scriptsize$\pm$0.006} & 0.323{\scriptsize$\pm$0.013} & 0.293{\scriptsize$\pm$0.036} & 0.401{\scriptsize$\pm$0.019} & 0.232{\scriptsize$\pm$0.022}  \\
Cup eccentricity & 4 & 0.250 / -- & 0.347{\scriptsize$\pm$0.007} & 0.410{\scriptsize$\pm$0.009} &  & 0.421{\scriptsize$\pm$0.006} &   \\
\midrule
\textit{Mean over biomarkers} &  & 0.286 / 0.257 & 0.414{\scriptsize\ (0.356--0.438)} & 0.523 & 0.354{\scriptsize\ (0.331--0.377)} & 0.527 & 0.332{\scriptsize\ (0.302--0.362)} \\
\botrule
\end{tabular}}
\end{table}

The two instances were strongest on different biomarkers. On WF\_Glaucoma, GlaucomaMarker-F was ahead on the vertical cup-to-disc ratio, 0.384 against 0.354, on the area cup-to-disc ratio, 0.385 against 0.360, and on the rim area fraction, 0.466 against 0.427. GlaucomaMarker-R was ahead on the vertical-to-horizontal ratio of the disc, 0.338 against 0.196, where it was level with KeepFIT at 0.337, and on disc eccentricity, 0.293 against 0.232. GlaucomaMarker-F scored below the random guessing level on both of these disc shape biomarkers, and the best result on disc eccentricity belonged to KeepFIT, 0.316. On Drishti-GS, GlaucomaMarker-R was above both remaining baselines on 8 of 11 biomarkers and below the random guessing level on three. These were disc eccentricity, 0.270 against 0.333, the disc vertical-to-horizontal ratio, 0.302 against 0.500, since only two of its three buckets occur in this cohort, and the rim vertical asymmetry, 0.228 against 0.250. The readout is a reliable signal for the cup and rim biomarkers, while the disc shape biomarkers remain too weak to read on their own.

Some biomarkers are tied to the diagnosis, so a model could score well on them by echoing its own diagnosis. We call a biomarker disease-linked when its true bucket is clearly associated with the true diagnosis in the training split, with a Cram\'er $V$ of at least 0.30. Six biomarkers meet this, namely the vertical, horizontal and area cup-to-disc ratios, the inferior and superior rim thickness, and the rim area fraction, and the remaining five all fall at 0.15 or below, as Supplementary Table~\ref{tab:linked_biomarker_definition} shows. WF\_Glaucoma has ground truth for three of the linked biomarkers. On these, the mean balanced accuracy was 0.411 for GlaucomaMarker-F, with an interval of 0.389 to 0.433, and 0.380 for GlaucomaMarker-R, with an interval of 0.356 to 0.403, against a random guessing level of 0.233. No baseline went above 0.258.

To check that the readout carries more than the diagnosis, we correlated it with the true bucket inside healthy images only and inside glaucoma images only, so that the diagnosis is held fixed. Table~\ref{tab:consistency_analysis} gives these correlations. For the vertical cup-to-disc ratio and the rim area fraction they were clearly positive inside both classes on both cohorts, for example 0.71 and 0.57 for the vertical ratio on Drishti-GS, and 0.45 and 0.44 for GlaucomaMarker-F on WF\_Glaucoma. The readout follows the measured structure beyond the diagnosis. The area cup-to-disc ratio was weaker inside glaucoma images, with 0.12 for GlaucomaMarker-R on WF\_Glaucoma and an interval from -0.025 to 0.245 that included zero. The rim vertical asymmetry shows the opposite pattern. Its readout tracked the diagnosis, with a correlation of 0.90, but correlated negatively with the true bucket inside each class, -0.24 and -0.13, with intervals that included zero, so we do not count it as a measured biomarker.

Beyond the biomarkers that fell below the random guessing level, described above, the biomarker reading has a further limit in the middle of each range. For each biomarker we separated the extreme buckets, the lowest and the highest, from the middle buckets, and averaged the recall of each group over the biomarkers and seeds. The reading was weakest in the middle buckets, with a recall of 0.205 for GlaucomaMarker-R and 0.103 for GlaucomaMarker-F on WF\_Glaucoma and 0.316 for GlaucomaMarker-R on Drishti-GS, against 0.505, 0.580 and 0.542 for the extreme buckets. Both instances lean toward the extreme buckets, which made up 62 to 88 percent of their predictions against 44 to 45 percent of the true values, and the lean was largest for GlaucomaMarker-F on WF\_Glaucoma, at 88 percent. The readout separates small measurements from large ones more reliably than it grades the values between.

\begin{table}[htbp]
\centering
\footnotesize
\caption{Agreement between the biomarker readout and the measured structure, as Spearman correlations with 95\% bootstrap intervals. The readout is the expected bucket index of the zero-shot bucket probabilities, signed so that larger means more glaucoma-like. \emph{Diagnosis} is the readout against the model's own glaucoma probability, and \emph{Truth, healthy} and \emph{Truth, glaucoma} are the readout against the true bucket within one diagnosis class. Biomarkers with no direction toward disease are not included, and blank cells have no ground truth or are not evaluated for pretraining overlap.}
\label{tab:consistency_analysis}
\resizebox{\ifdim\width>\linewidth\linewidth\else\width\fi}{!}{%
\begin{tabular}{lccccccccc}
\toprule
& \multicolumn{3}{c}{Drishti-GS, GlaucomaMarker-R} & \multicolumn{3}{c}{WF\_Glaucoma, GlaucomaMarker-R} & \multicolumn{3}{c}{WF\_Glaucoma, GlaucomaMarker-F} \\
\cmidrule(lr){2-4}\cmidrule(lr){5-7}\cmidrule(lr){8-10}
Biomarker & Diagnosis & Truth, healthy & Truth, glaucoma & Diagnosis & Truth, healthy & Truth, glaucoma & Diagnosis & Truth, healthy & Truth, glaucoma \\
\midrule
\addlinespace[2pt]
\multicolumn{10}{l}{\scriptsize\textsc{Disease-linked}} \\
\quad Vertical CDR & 0.79{\scriptsize\ (0.72--0.84)} & 0.71{\scriptsize\ (0.48--0.83)} & 0.57{\scriptsize\ (0.41--0.70)} & 0.82{\scriptsize\ (0.78--0.85)} & 0.41{\scriptsize\ (0.28--0.53)} & 0.32{\scriptsize\ (0.18--0.44)} & 0.87{\scriptsize\ (0.83--0.90)} & 0.45{\scriptsize\ (0.31--0.56)} & 0.44{\scriptsize\ (0.32--0.56)} \\
\quad Horizontal CDR & 0.77{\scriptsize\ (0.69--0.82)} & 0.71{\scriptsize\ (0.50--0.84)} & 0.44{\scriptsize\ (0.24--0.62)} &  &  &  &  &  &   \\
\quad Area CDR & 0.65{\scriptsize\ (0.53--0.73)} & 0.76{\scriptsize\ (0.55--0.86)} & 0.28{\scriptsize\ (0.10--0.44)} & 0.69{\scriptsize\ (0.63--0.74)} & 0.44{\scriptsize\ (0.33--0.54)} & 0.12{\scriptsize\ ($-$0.02--0.24)} & 0.78{\scriptsize\ (0.73--0.83)} & 0.53{\scriptsize\ (0.40--0.64)} & 0.29{\scriptsize\ (0.14--0.42)} \\
\quad Rim area fraction & 0.89{\scriptsize\ (0.85--0.91)} & 0.68{\scriptsize\ (0.45--0.83)} & 0.60{\scriptsize\ (0.45--0.71)} & 0.86{\scriptsize\ (0.83--0.89)} & 0.45{\scriptsize\ (0.32--0.56)} & 0.46{\scriptsize\ (0.32--0.58)} & 0.95{\scriptsize\ (0.93--0.96)} & 0.38{\scriptsize\ (0.23--0.51)} & 0.60{\scriptsize\ (0.47--0.69)} \\
\quad Rim thickness (inf.) & 0.81{\scriptsize\ (0.74--0.85)} & 0.46{\scriptsize\ (0.16--0.68)} & 0.34{\scriptsize\ (0.13--0.52)} &  &  &  &  &  &   \\
\quad Rim thickness (sup.) & 0.80{\scriptsize\ (0.73--0.85)} & 0.29{\scriptsize\ ($-$0.02--0.55)} & 0.37{\scriptsize\ (0.15--0.55)} &  &  &  &  &  &   \\
\addlinespace[2pt]
\multicolumn{10}{l}{\scriptsize\textsc{Other}} \\
\quad Rim vert. asymmetry & 0.90{\scriptsize\ (0.87--0.92)} & $-$0.24{\scriptsize\ ($-$0.50--0.05)} & $-$0.13{\scriptsize\ ($-$0.34--0.08)} &  &  &  &  &  &   \\
\botrule
\end{tabular}}
\end{table}

\begin{table}[htbp]
\centering
\footnotesize
\caption{Zero-shot biomarker balanced accuracy of GlaucomaMarker and the baselines on Drishti-GS and WF\_Glaucoma, with 101 and 444 images. GlaucomaMarker is the mean $\pm$ standard deviation over three seeds and the baselines are single runs. Blank cells have no ground truth or are excluded for pretraining overlap. The random guessing level is given as D / W for Drishti-GS and WF\_Glaucoma, and the last row is the mean over the biomarkers with 95\% bootstrap intervals. EyeCLIP predicts a single bucket for every image on WF\_Glaucoma, so its interval has zero width. Best model per biomarker and cohort in bold.}
\label{tab:concept_accuracy_baselines}
\resizebox{\ifdim\width>\linewidth\linewidth\else\width\fi}{!}{%
\begin{tabular}{lcccccccccccc}
\toprule
& & & \multicolumn{3}{c}{Drishti-GS (n=101)} & \multicolumn{7}{c}{WF\_Glaucoma (n=444)} \\
\cmidrule(lr){4-6}\cmidrule(lr){7-13}
Biomarker & \#classes & Random guess (D / W) & Ours-R & EyeCLIP & FairCLIP & Ours-R & Ours-F & FLAIR & KeepFIT & UniMed & EyeCLIP & FairCLIP \\
\midrule
Vertical CDR & 5 & 0.250 / 0.200 & \textbf{0.405{\scriptsize$\pm$0.048}} & 0.250 & 0.260 & 0.354{\scriptsize$\pm$0.024} & \textbf{0.384{\scriptsize$\pm$0.005}} & 0.201 & 0.250 & 0.196 & 0.200 & 0.186  \\
Horizontal CDR & 5 & 0.250 / -- & \textbf{0.538{\scriptsize$\pm$0.035}} & 0.240 & 0.250 &  &  &  &  &  &  &   \\
Area CDR & 4 & 0.333 / 0.250 & \textbf{0.627{\scriptsize$\pm$0.016}} & 0.333 & 0.000 & 0.360{\scriptsize$\pm$0.035} & \textbf{0.385{\scriptsize$\pm$0.011}} & 0.250 & 0.238 & 0.251 & 0.250 & 0.236  \\
Rim area fraction & 4 & 0.250 / 0.250 & \textbf{0.546{\scriptsize$\pm$0.025}} & 0.250 & 0.179 & 0.427{\scriptsize$\pm$0.051} & \textbf{0.466{\scriptsize$\pm$0.055}} & 0.258 & 0.285 & 0.240 & 0.250 & 0.255  \\
Rim thickness (inf.) & 5 & 0.200 / -- & \textbf{0.316{\scriptsize$\pm$0.023}} & 0.200 & 0.200 &  &  &  &  &  &  &   \\
Rim thickness (sup.) & 5 & 0.200 / -- & \textbf{0.376{\scriptsize$\pm$0.011}} & 0.200 & 0.200 &  &  &  &  &  &  &   \\
Rim vert. asymmetry & 4 & 0.250 / -- & 0.228{\scriptsize$\pm$0.017} & \textbf{0.250} & \textbf{0.250} &  &  &  &  &  &  &   \\
Disc V/H ratio & 3 & 0.500 / 0.333 & 0.302{\scriptsize$\pm$0.197} & \textbf{0.500} & \textbf{0.500} & \textbf{0.338{\scriptsize$\pm$0.041}} & 0.196{\scriptsize$\pm$0.050} & 0.334 & 0.337 & 0.319 & 0.333 & 0.333  \\
Cup V/H ratio & 3 & 0.333 / -- & \textbf{0.602{\scriptsize$\pm$0.015}} & 0.333 & 0.333 &  &  &  &  &  &  &   \\
Disc eccentricity & 4 & 0.333 / 0.250 & 0.270{\scriptsize$\pm$0.006} & 0.000 & \textbf{0.333} & 0.293{\scriptsize$\pm$0.036} & 0.232{\scriptsize$\pm$0.022} & 0.289 & \textbf{0.316} & 0.252 & 0.250 & 0.250  \\
Cup eccentricity & 4 & 0.250 / -- & \textbf{0.347{\scriptsize$\pm$0.007}} & 0.250 & 0.250 &  &  &  &  &  &  &   \\
\midrule
\textit{Mean over biomarkers} &  & 0.286 / 0.257 & 0.414{\scriptsize\ (0.356--0.438)} & 0.255{\scriptsize\ (0.232--0.293)} & 0.250{\scriptsize\ (0.214--0.279)} & 0.354{\scriptsize\ (0.331--0.377)} & 0.332{\scriptsize\ (0.302--0.362)} & 0.266{\scriptsize\ (0.235--0.302)} & 0.285{\scriptsize\ (0.258--0.314)} & 0.251{\scriptsize\ (0.229--0.272)} & 0.257{\scriptsize\ (0.257--0.257)} & 0.252{\scriptsize\ (0.242--0.263)} \\
\botrule
\end{tabular}}
\end{table}

\subsection*{Training on biomarker text improves the image embedding, most for the RETFound start}
\label{sec:results-embedding}

\begin{figure}[htbp]
\centering
\includegraphics[width=\textwidth]{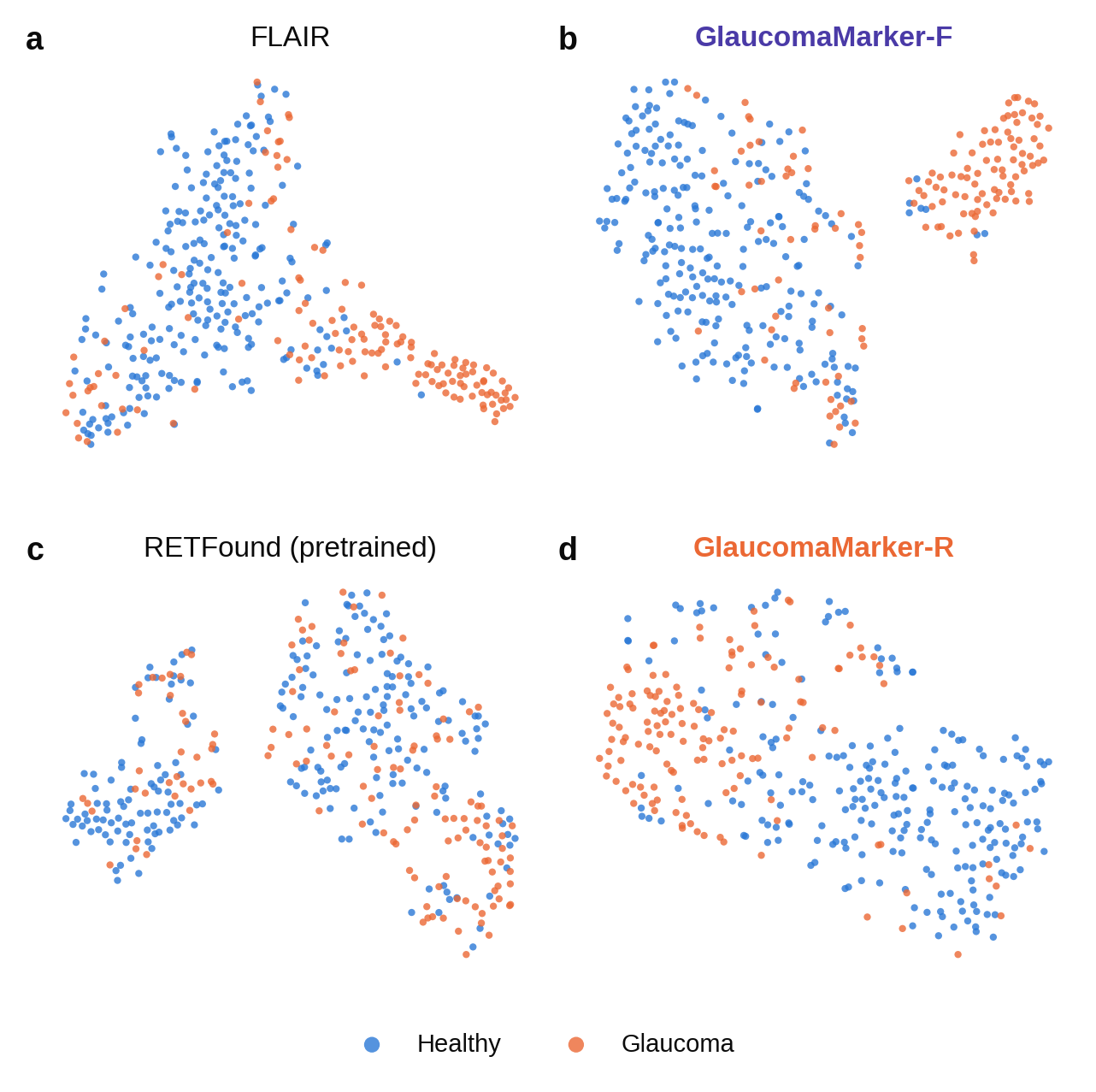}
\caption{\textbf{Image embeddings of WF\_Glaucoma before and after training on biomarker text.} t-SNE of frozen image embeddings, colored by disease label, blue for healthy and orange for glaucoma. \textbf{a} FLAIR. \textbf{b} GlaucomaMarker-F, which starts from FLAIR. \textbf{c} RETFound before training on our text. \textbf{d} GlaucomaMarker-R, which starts from RETFound. Panels \textbf{b} and \textbf{d} show seed 42. t-SNE is a qualitative view and its axes have no units.}
\label{fig:tsne_embedding_visualization}
\end{figure}

We next investigated whether training on biomarker text improves the image embeddings themselves, beyond the zero-shot readout. Figure~\ref{fig:tsne_embedding_visualization} shows t-SNE plots of the frozen image embeddings of WF\_Glaucoma, with each instance placed next to the model it starts from. Before training on our text, the self-supervised RETFound embeddings mix the two classes, and afterwards GlaucomaMarker-R gathers the glaucoma images in one region. FLAIR already separates the classes along one arm, and GlaucomaMarker-F tightens this into a compact cluster with a clear gap. As a number behind the picture, we measured the share of the ten nearest neighbors of an image that carry its label. For RETFound it rose from 0.603 to 0.763 in GlaucomaMarker-R, a gain of 0.160 with a 95 percent confidence interval of 0.127 to 0.194, and the starting value is below the 0.628 that a rule naming the more common class would reach. For FLAIR it rose from 0.787 to 0.806 in GlaucomaMarker-F, a gain of 0.019 with an interval of 0.002 to 0.037. The gain is large for RETFound and small for FLAIR. The GlaucomaMarker values are means over three seeds, and t-SNE is a qualitative view of one seed.

A linear probe gives the same picture. We fit a linear classifier on the frozen image features to see how much diagnostic information they hold, and Table~\ref{tab:linear_probe} gives the result. On WF\_Glaucoma, the probe AUROC of GlaucomaMarker-F was 0.898, with an interval of 0.855 to 0.935, against 0.820 for FLAIR, a difference of 0.079 with an interval of 0.033 to 0.128. GlaucomaMarker-R reached 0.850, which is above every baseline, but its difference from FLAIR of 0.031 had an interval of -0.027 to 0.089. On Drishti-GS, GlaucomaMarker-R reached 0.841, level with the RETFound glaucoma checkpoint at 0.838 and not distinguishable from FairCLIP at 0.854, with a difference of -0.014 and an interval of -0.100 to 0.074. It was below RETFound-DE at 0.915, by 0.075 with an interval of 0.025 to 0.129.

\begin{figure}[htbp]
\centering
\includegraphics[width=\textwidth]{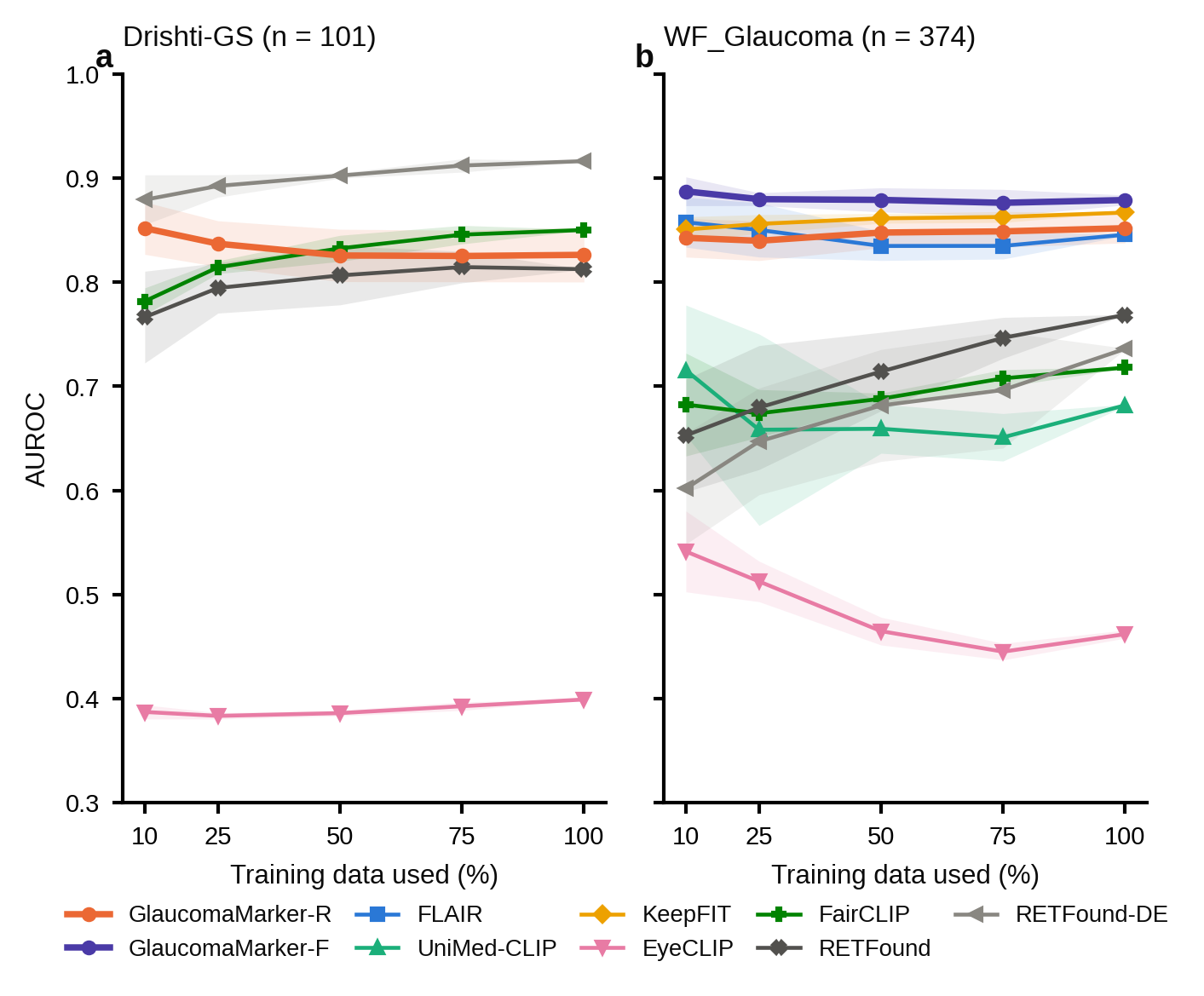}
\caption{\textbf{Linear-probe AUROC against the fraction of training data.} \textbf{a} Drishti-GS. \textbf{b} WF\_Glaucoma. Each line is a model, with GlaucomaMarker-R in orange and GlaucomaMarker-F in purple. Lines are means and shaded bands are one standard deviation over resamples. Panel \textbf{a} omits models whose pretraining data overlap with Drishti-GS.}
\label{fig:label_efficiency_curves}
\end{figure}

\begin{figure}[htbp]
\centering
\includegraphics[width=\textwidth]{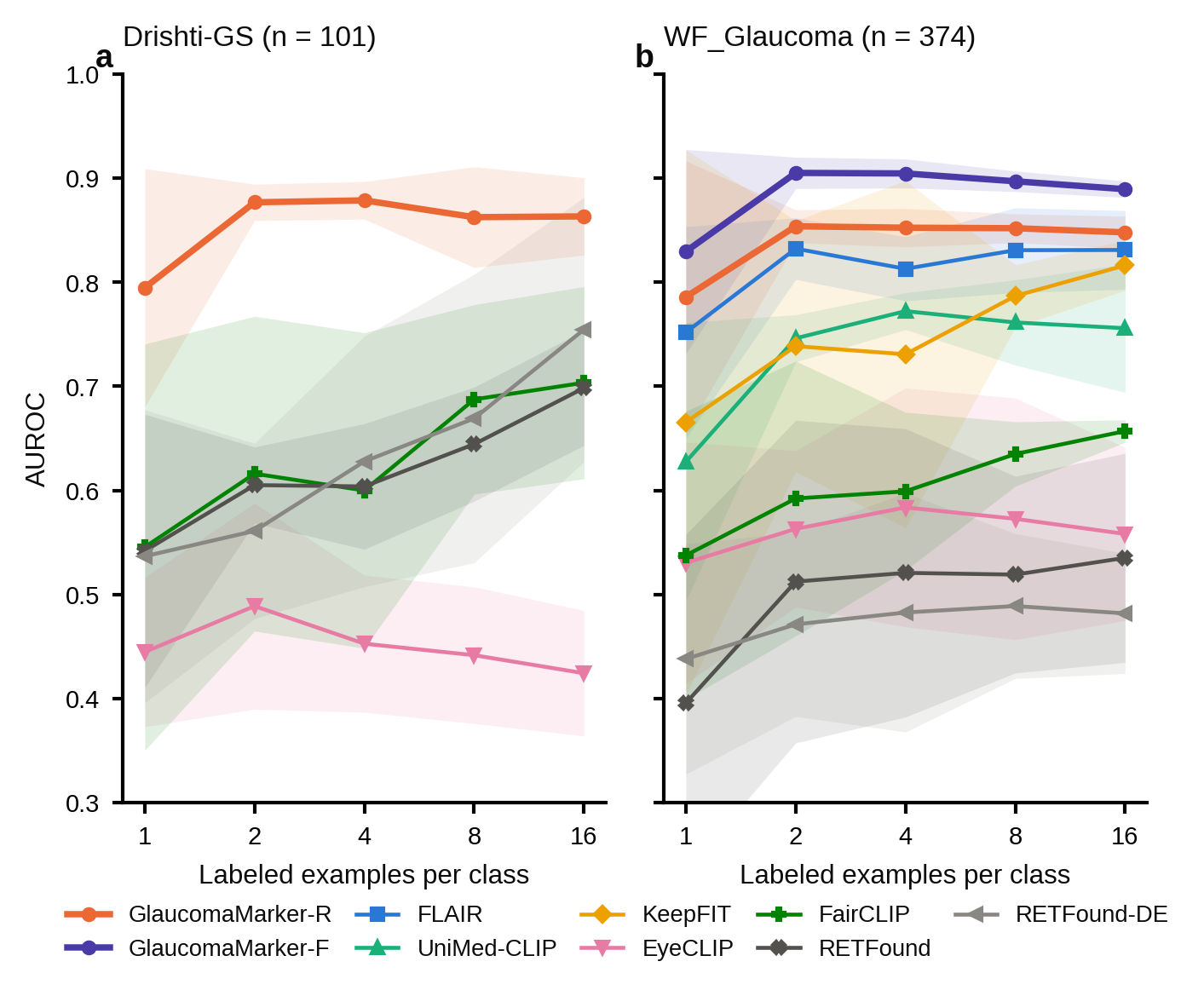}
\caption{\textbf{Few-shot linear-probe AUROC against the number of labeled examples per class.} \textbf{a} Drishti-GS. \textbf{b} WF\_Glaucoma. The number per class is 1, 2, 4, 8 or 16, on a logarithmic axis. Each line is a model, with GlaucomaMarker-R in orange and GlaucomaMarker-F in purple. Lines are means and shaded bands are one standard deviation over resamples. Panel \textbf{a} omits models whose pretraining data overlap with Drishti-GS.}
\label{fig:fewshot_classification_curves}
\end{figure}

The features also needed very few labels. With two labeled images per class, the probe already reached 0.905 on WF\_Glaucoma for GlaucomaMarker-F and 0.854 for GlaucomaMarker-R, against 0.832 for FLAIR. More labels barely changed this. From 10 to 100 percent of the training split, the probe AUROC of GlaucomaMarker-F on WF\_Glaucoma went from 0.887 to 0.879, while the RETFound glaucoma checkpoint rose from 0.653 to 0.769, as Figures~\ref{fig:label_efficiency_curves} and~\ref{fig:fewshot_classification_curves} show. These curves use the same probe settings for all models, which differ slightly from those of the table.

\begin{table}[htbp]
\centering
\small
\caption{Linear-probe classification performance on frozen image-encoder features. AUROC is given with a 95\% bootstrap confidence interval, and AUPRC is the mean $\pm$ standard deviation over three seeds for GlaucomaMarker and a single value for the baselines. Models whose pretraining data overlap with Drishti-GS, including GlaucomaMarker-F, are blank or omitted there. Best value per column in bold.}
\label{tab:linear_probe}
\resizebox{\ifdim\width>\linewidth\linewidth\else\width\fi}{!}{%
\begin{tabular}{lcccc}
\toprule
& \multicolumn{2}{c}{Drishti-GS (n=101)} & \multicolumn{2}{c}{WF\_Glaucoma (n=374)} \\
\cmidrule(lr){2-3}\cmidrule(lr){4-5}
Model & AUROC (95\% CI) & AUPRC & AUROC (95\% CI) & AUPRC \\
\midrule
\textbf{GlaucomaMarker-F} &  &  & \textbf{0.898{\scriptsize\ (0.855--0.935)}} & \textbf{0.882{\scriptsize$\pm$0.004}}  \\
\textbf{GlaucomaMarker-R} & 0.841{\scriptsize\ (0.756--0.911)} & 0.914{\scriptsize$\pm$0.032} & 0.850{\scriptsize\ (0.803--0.893)} & 0.802{\scriptsize$\pm$0.015}  \\
FLAIR &  &  & 0.820{\scriptsize\ (0.760--0.873)} & 0.814  \\
UniMed-CLIP &  &  & 0.790{\scriptsize\ (0.730--0.846)} & 0.752  \\
RETFound & 0.838{\scriptsize\ (0.745--0.916)} & 0.916 & 0.785{\scriptsize\ (0.723--0.841)} & 0.702  \\
KeepFIT (MM-Retinal) &  &  & 0.766{\scriptsize\ (0.706--0.822)} & 0.675  \\
RETFound-DE & \textbf{0.915{\scriptsize\ (0.848--0.969)}} & \textbf{0.958} & 0.719{\scriptsize\ (0.650--0.783)} & 0.649  \\
FairCLIP & 0.854{\scriptsize\ (0.763--0.929)} & 0.916 & 0.692{\scriptsize\ (0.627--0.755)} & 0.550  \\
EyeCLIP & 0.411{\scriptsize\ (0.286--0.539)} & 0.639 & 0.634{\scriptsize\ (0.558--0.708)} & 0.488  \\
\botrule
\end{tabular}}
\end{table}

\begin{figure}[htbp]
\centering
\includegraphics[width=\textwidth]{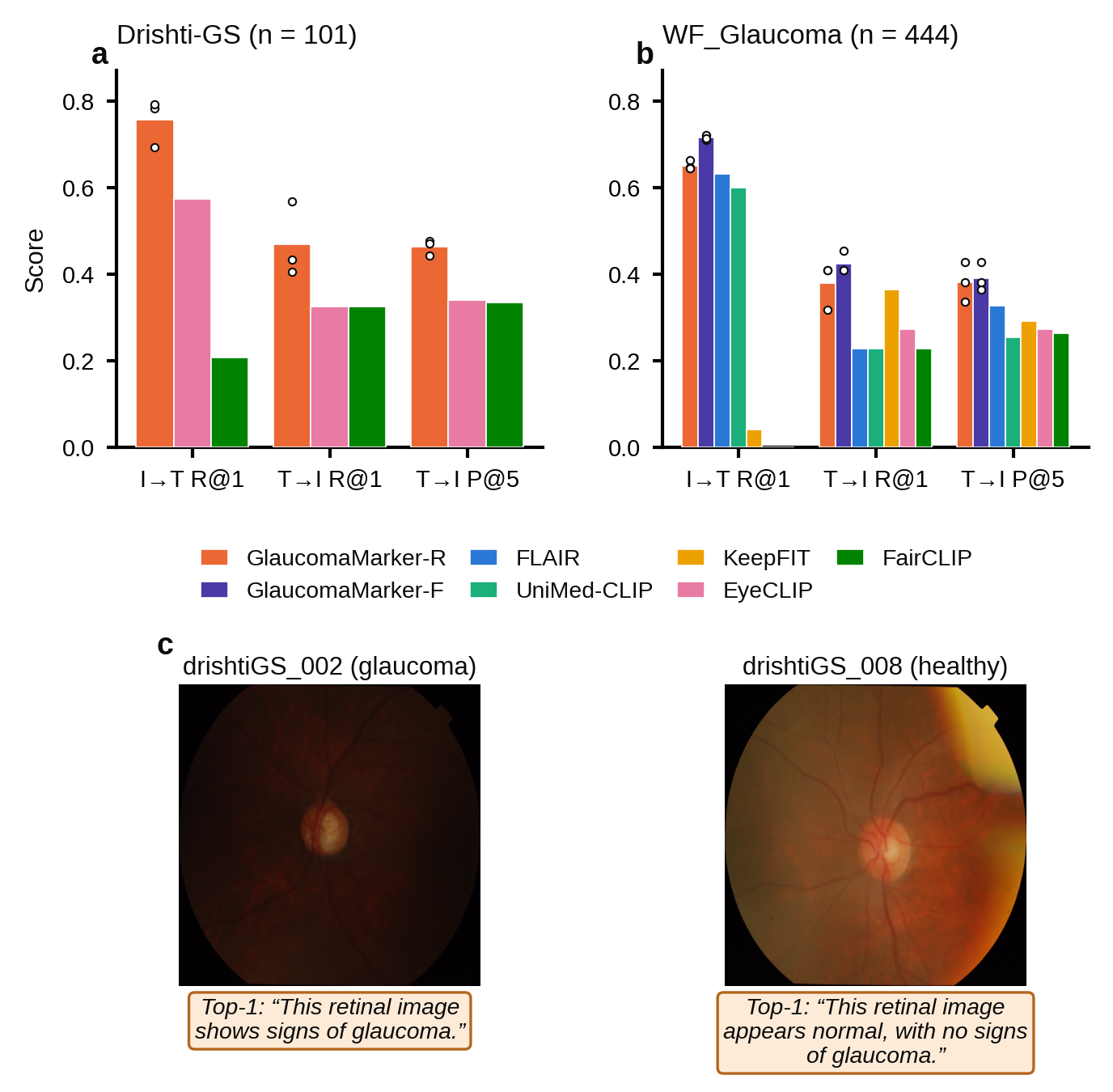}
\caption{\textbf{Cross-modal retrieval.} \textbf{a} Drishti-GS and \textbf{b} WF\_Glaucoma: image-to-text recall at rank 1 (I to T R@1), text-to-image recall at rank 1 (T to I R@1) and text-to-image precision at rank 5 (T to I P@5). Bars are the mean over three seeds for GlaucomaMarker, with the three seeds as open circles, and single runs for the baselines. Panel \textbf{a} omits the FLAIR-initialized instance and the baselines pretrained on Drishti-GS. \textbf{c} Example image-to-text retrievals from GlaucomaMarker-R, showing the top-1 of the 48 sentences for a glaucoma-positive and a healthy image.}
\label{fig:crossmodal_retrieval_performance}
\end{figure}

Retrieval is a stricter test than diagnosis, because the model must pick out the sentences of one image from those of all others. Table~\ref{tab:retrieval} reports it over all 48 sentences, and Figure~\ref{fig:crossmodal_retrieval_performance} shows example retrievals. Given an image, GlaucomaMarker-F found one of its true sentences first most often on WF\_Glaucoma, with a Recall@1 of 0.715, ahead of GlaucomaMarker-R at 0.651 and FLAIR at 0.631. KeepFIT, EyeCLIP and FairCLIP scored 0.041 or below. Given a sentence, GlaucomaMarker-F also ranked matching images best, with a Recall@1 of 0.424 and a P@5 of 0.391, and GlaucomaMarker-R was level with KeepFIT on Recall@1, within its seed variation. On Drishti-GS, GlaucomaMarker-R was best on all three metrics among the models shown. Retrieval supports the same reading, that the embeddings are specific to the image and carry more than its class.

\begin{table}[htbp]
\centering
\small
\caption{Cross-modal retrieval performance. GlaucomaMarker-R starts from RETFound (ViT-L/16, BioBERT) and GlaucomaMarker-F from FLAIR (ResNet-50, Bio\_ClinicalBERT). Both are reported as the mean $\pm$ standard deviation over three seeds; baselines are single runs. Drishti-GS: n=101 images; WF\_Glaucoma: n=444 images. FLAIR, UniMed-CLIP, and KeepFIT are blank or omitted on Drishti-GS because their pretraining data overlap with Drishti-GS, and the FLAIR-initialized model is blank there for the same reason. Best value per column in bold.}
\label{tab:retrieval}
\resizebox{\ifdim\width>\linewidth\linewidth\else\width\fi}{!}{%
\begin{tabular}{lcccccc}
\toprule
& \multicolumn{3}{c}{Drishti-GS (n=101)} & \multicolumn{3}{c}{WF\_Glaucoma (n=444)} \\
\cmidrule(lr){2-4}\cmidrule(lr){5-7}
Model & I$\to$T R@1 & T$\to$I R@1 & T$\to$I P@5 & I$\to$T R@1 & T$\to$I R@1 & T$\to$I P@5 \\
\midrule
\textbf{GlaucomaMarker-R} & \textbf{0.756{\scriptsize$\pm$0.055}} & \textbf{0.468{\scriptsize$\pm$0.087}} & \textbf{0.463{\scriptsize$\pm$0.017}} & 0.651{\scriptsize$\pm$0.010} & 0.379{\scriptsize$\pm$0.052} & 0.382{\scriptsize$\pm$0.045} \\
\textbf{GlaucomaMarker-F} &  &  &  & \textbf{0.715{\scriptsize$\pm$0.006}} & \textbf{0.424{\scriptsize$\pm$0.026}} & \textbf{0.391{\scriptsize$\pm$0.033}} \\
FLAIR &  &  &  & 0.631 & 0.227 & 0.327 \\
UniMed-CLIP &  &  &  & 0.599 & 0.227 & 0.255 \\
KeepFIT (MM-Retinal) &  &  &  & 0.041 & 0.364 & 0.291 \\
EyeCLIP & 0.574 & 0.324 & 0.341 & 0.000 & 0.273 & 0.273 \\
FairCLIP & 0.208 & 0.324 & 0.335 & 0.000 & 0.227 & 0.264 \\
\botrule
\end{tabular}}
\end{table}

\subsection*{Performance is similar across demographic subgroups}
\label{sec:results-fairness}

\begin{figure}[htbp]
\centering
\includegraphics[width=\textwidth]{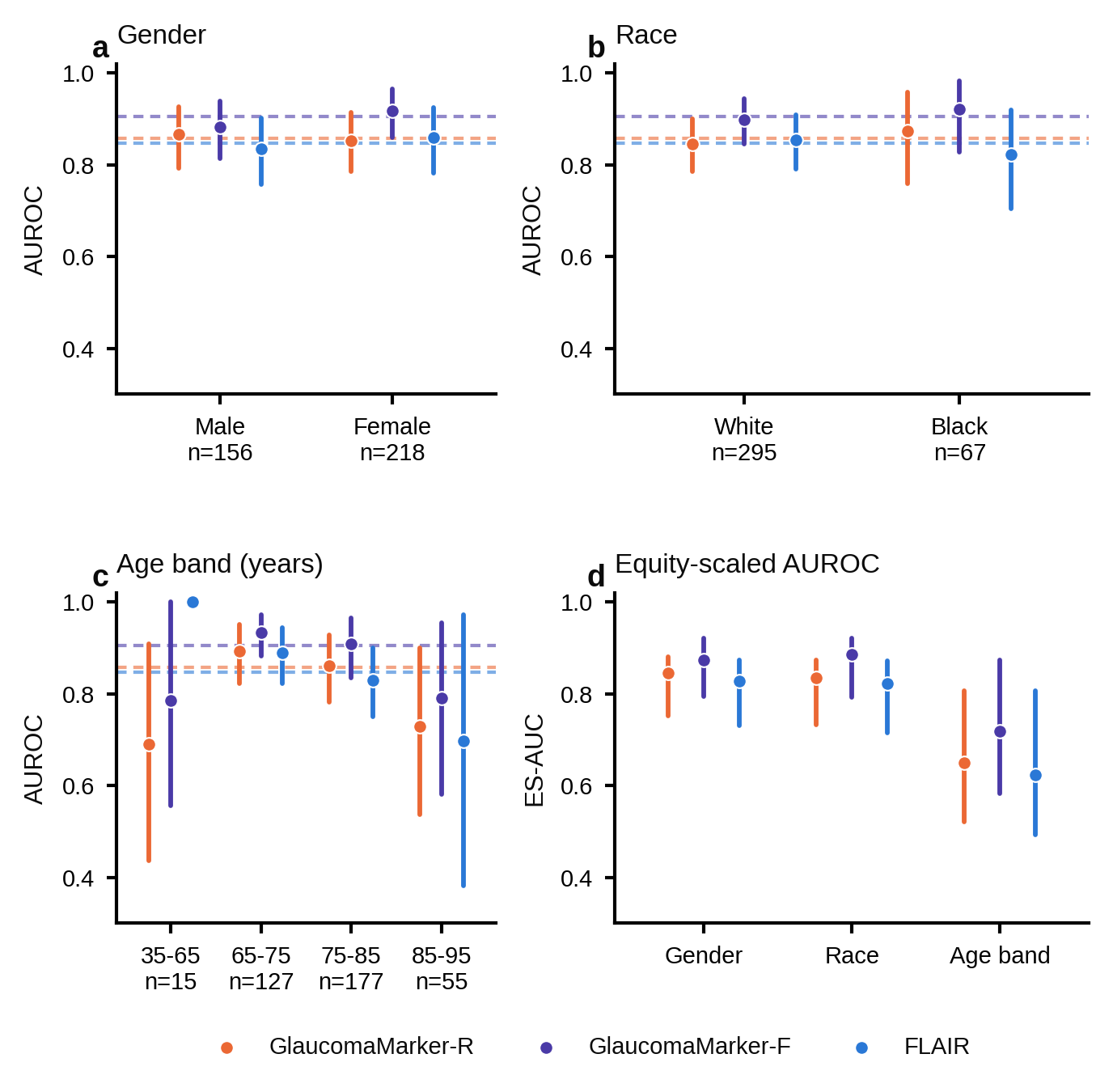}
\caption{\textbf{Performance across demographic subgroups on WF\_Glaucoma.} \textbf{a} AUROC by gender. \textbf{b} AUROC by race. \textbf{c} AUROC by age band in years. \textbf{d} Equity-scaled AUROC (ES-AUC) by gender, race and age band. Points are the AUROC or ES-AUC and lines are 95\% bootstrap confidence intervals. Orange is GlaucomaMarker-R, purple is GlaucomaMarker-F and blue is the released FLAIR model. Dashed lines in panels \textbf{a} to \textbf{c} mark the overall AUROC of each model, and n is the number of eyes. The 35 to 65 band has only 15 eyes, as noted under Table~\ref{tab:fairness_subgroup}.}
\label{fig:fairness_subgroup_analysis}
\end{figure}

A screening model should work equally well for different groups of patients. WF\_Glaucoma records gender, race and age, so we measured the AUROC within each subgroup and summarized the spread with the equity-scaled AUROC of Luo et al.~\cite{luo2024harvardgf}, which is an AUROC that is lowered when the subgroups differ. Table~\ref{tab:fairness_subgroup} and Figure~\ref{fig:fairness_subgroup_analysis} compare both instances with FLAIR, the strongest baseline. Within each model, the intervals of the subgroups overlapped by gender, by race and by age. By gender, GlaucomaMarker-F had an AUROC of 0.882 for men and 0.918 for women, and GlaucomaMarker-R had 0.866 and 0.853. By race, GlaucomaMarker-F had 0.899 for White and 0.920 for Black patients, and GlaucomaMarker-R had 0.846 and 0.873.

The equity-scaled AUROC by gender and by race was 0.874 and 0.886 for GlaucomaMarker-F, 0.846 and 0.835 for GlaucomaMarker-R, and 0.827 and 0.822 for FLAIR. The equity-scaled AUROC of both instances was higher than that of FLAIR in every comparison, by 0.013 to 0.095 across gender, race and age. These differences have intervals that include zero, for example 0.047 for GlaucomaMarker-F by gender, with an interval of -0.004 to 0.119, and 0.013 for GlaucomaMarker-R by race, with an interval of -0.085 to 0.103. They point in the same direction as the overall AUROC, so there is no sign that the higher accuracy of the two instances came with a wider gap between groups.

Age showed the largest disparity for all models, with an equity-scaled AUROC of 0.717 for GlaucomaMarker-F, 0.650 for GlaucomaMarker-R and 0.622 for FLAIR. The weakest band was 85 to 95 years, with an AUROC of 0.791 for GlaucomaMarker-F, 0.729 for GlaucomaMarker-R and 0.697 for FLAIR, and the intervals for this band of 55 eyes were wide, from 0.382 to 0.972 for FLAIR. The youngest band has only 15 eyes, and its values are too unstable to interpret, as FLAIR scored 1.000 on it while GlaucomaMarker-R scored 0.690, and only 5,682 of 10,000 resamples contained both classes. We left out the Hispanic and mixed-race subgroups, which have only 4 and 8 images.

\begin{table}[htbp]
\centering
\small
\caption{Subgroup performance and equity-scaled AUROC (ES-AUC) by gender, race, and age band on WF\_Glaucoma, for the two GlaucomaMarker instances and the released FLAIR model. AUROC and ES-AUC are given with 95\% bootstrap intervals from 10,000 patient-level resamples, with the three seeds of GlaucomaMarker averaged inside each resample. AUPRC is the mean $\pm$ standard deviation over three seeds for GlaucomaMarker and a single run for FLAIR. The 35-65 band has 15 eyes, and only 5,682 of 10,000 resamples contained both classes. Subgroups with fewer than ten images or a single class are not scored.}
\label{tab:fairness_subgroup}
\resizebox{\ifdim\width>\linewidth\linewidth\else\width\fi}{!}{%
\begin{tabular}{lccccccc}
\toprule
& & \multicolumn{2}{c}{GlaucomaMarker-R} & \multicolumn{2}{c}{GlaucomaMarker-F} & \multicolumn{2}{c}{FLAIR (released)} \\
\cmidrule(lr){3-4}\cmidrule(lr){5-6}\cmidrule(lr){7-8}
Subgroup & n & AUROC (95\% CI) & AUPRC & AUROC (95\% CI) & AUPRC & AUROC (95\% CI) & AUPRC \\
\midrule
Overall & 374 & 0.858{\scriptsize\ (0.808--0.901)} & 0.817{\scriptsize$\pm$0.019} & 0.905{\scriptsize\ (0.864--0.940)} & 0.885{\scriptsize$\pm$0.002} & 0.848{\scriptsize\ (0.795--0.894)} & 0.813  \\
\quad Male & 156 & 0.866{\scriptsize\ (0.792--0.925)} & 0.883{\scriptsize$\pm$0.005} & 0.882{\scriptsize\ (0.814--0.938)} & 0.893{\scriptsize$\pm$0.004} & 0.835{\scriptsize\ (0.757--0.902)} & 0.829  \\
\quad Female & 218 & 0.853{\scriptsize\ (0.785--0.913)} & 0.757{\scriptsize$\pm$0.040} & 0.918{\scriptsize\ (0.859--0.965)} & 0.884{\scriptsize$\pm$0.002} & 0.860{\scriptsize\ (0.782--0.925)} & 0.814  \\
\quad ES-AUC (gender) & -- & \multicolumn{2}{c}{0.846{\scriptsize\ (0.753--0.880)}} & \multicolumn{2}{c}{0.874{\scriptsize\ (0.794--0.921)}} & \multicolumn{2}{c}{0.827{\scriptsize\ (0.731--0.874)}} \\
\quad White & 295 & 0.846{\scriptsize\ (0.785--0.900)} & 0.777{\scriptsize$\pm$0.026} & 0.899{\scriptsize\ (0.845--0.944)} & 0.865{\scriptsize$\pm$0.005} & 0.853{\scriptsize\ (0.790--0.908)} & 0.784  \\
\quad Black & 67 & 0.873{\scriptsize\ (0.759--0.958)} & 0.913{\scriptsize$\pm$0.016} & 0.920{\scriptsize\ (0.828--0.983)} & 0.946{\scriptsize$\pm$0.003} & 0.822{\scriptsize\ (0.704--0.919)} & 0.902  \\
\quad Hispanic & 4 & -- & -- & -- & -- & -- & --  \\
\quad Mixed & 8 & -- & -- & -- & -- & -- & --  \\
\quad ES-AUC (race) & -- & \multicolumn{2}{c}{0.835{\scriptsize\ (0.732--0.873)}} & \multicolumn{2}{c}{0.886{\scriptsize\ (0.792--0.922)}} & \multicolumn{2}{c}{0.822{\scriptsize\ (0.714--0.872)}} \\
\quad 35-65 & 15 & 0.690{\scriptsize\ (0.436--0.909)} & 0.231{\scriptsize$\pm$0.112} & 0.786{\scriptsize\ (0.556--1.000)} & 0.261{\scriptsize$\pm$0.067} & 1.000{\scriptsize\ (1.000--1.000)} & 1.000  \\
\quad 65-75 & 127 & 0.893{\scriptsize\ (0.822--0.950)} & 0.936{\scriptsize$\pm$0.006} & 0.933{\scriptsize\ (0.882--0.972)} & 0.963{\scriptsize$\pm$0.001} & 0.889{\scriptsize\ (0.822--0.943)} & 0.939  \\
\quad 75-85 & 177 & 0.862{\scriptsize\ (0.782--0.927)} & 0.757{\scriptsize$\pm$0.028} & 0.908{\scriptsize\ (0.834--0.964)} & 0.834{\scriptsize$\pm$0.004} & 0.830{\scriptsize\ (0.749--0.899)} & 0.675  \\
\quad 85-95 & 55 & 0.729{\scriptsize\ (0.536--0.899)} & 0.482{\scriptsize$\pm$0.042} & 0.791{\scriptsize\ (0.580--0.955)} & 0.662{\scriptsize$\pm$0.029} & 0.697{\scriptsize\ (0.382--0.972)} & 0.657  \\
\quad ES-AUC (age\_band) & -- & \multicolumn{2}{c}{0.650{\scriptsize\ (0.522--0.807)}} & \multicolumn{2}{c}{0.717{\scriptsize\ (0.583--0.874)}} & \multicolumn{2}{c}{0.622{\scriptsize\ (0.492--0.807)}} \\
\botrule
\end{tabular}}
\end{table}

\subsection*{Ablation studies}
\label{sec:results-ablation}

\begin{figure}[htbp]
\centering
\includegraphics[width=\textwidth]{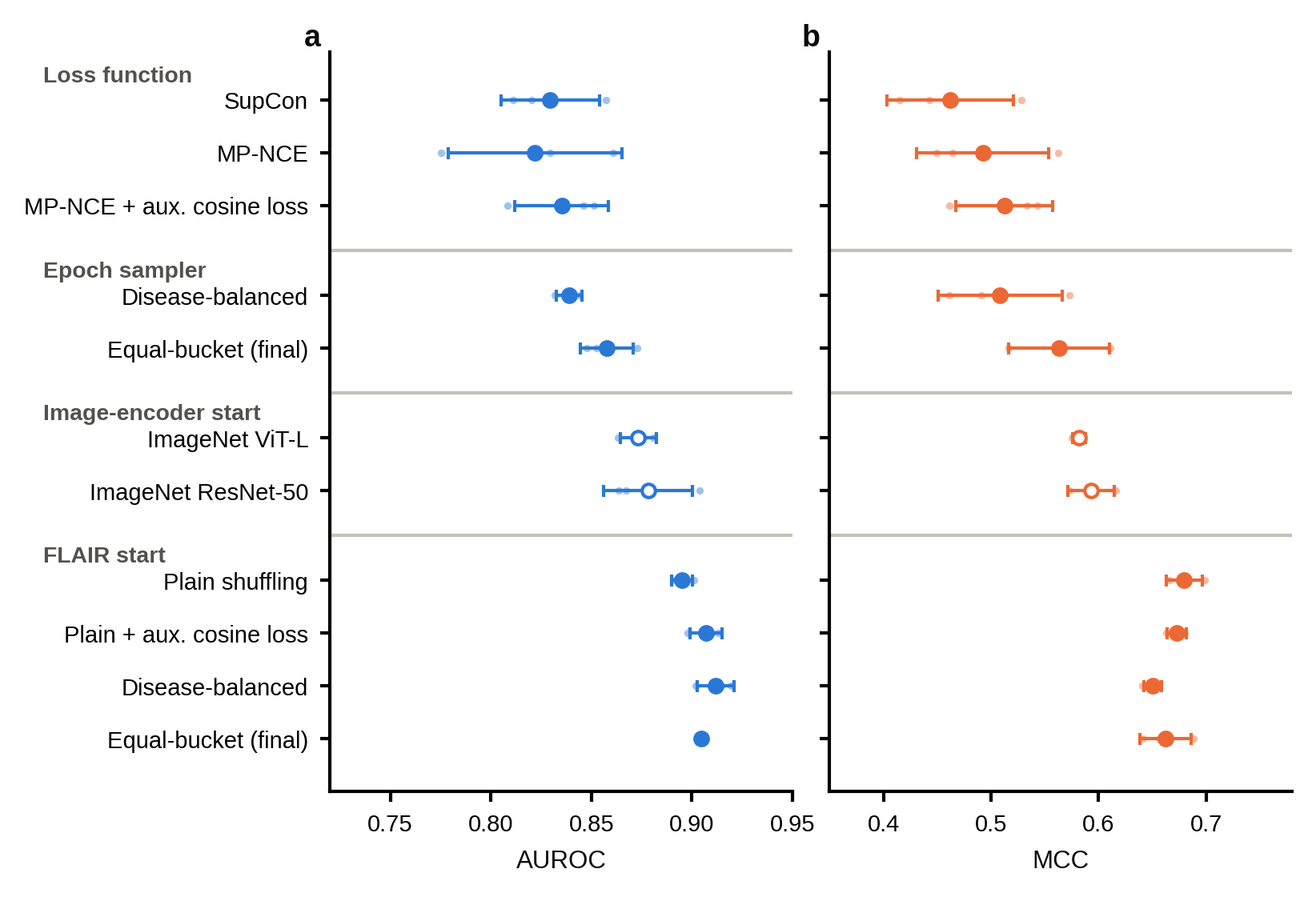}
\caption{\textbf{Ablation of the training pipeline on WF\_Glaucoma.} \textbf{a} AUROC and \textbf{b} MCC for each configuration, grouped as loss function, epoch sampler with the RETFound start, image-encoder start, and FLAIR start. Filled circles are the mean over three seeds with error bars of one standard deviation, and the small points are the individual seeds. Hollow markers replace the RETFound weights of the image encoder with ImageNet weights, with the rest of the pipeline fixed. Aux. cosine loss is the auxiliary cosine loss. The comparison of five rebalancing techniques is in Supplementary Table~\ref{tab:ablation_rebalancing_technique}.}
\label{fig:ablation_studies}
\end{figure}

We ablated the training text, the starting model, the training data, the loss and the epoch sampler, as set out in Methods under Ablation design, and Tables~\ref{tab:ablation_loss_ladder}, \ref{tab:ablation_operating_point} and~\ref{tab:ablation_representation} and Figure~\ref{fig:ablation_studies}a give the results. Every row is the mean over three seeds, and all rows start from RETFound unless the row says otherwise. We call a difference between two configurations clear when the seed ranges of the two do not overlap, and Supplementary Tables~\ref{tab:ablation_seed_ranges} and~\ref{tab:ablation_gap} give the seed values.

\emph{Removing the biomarker sentences removes the biomarker reading of the RETFound start.} If the biomarker sentences are what teach the model to read biomarkers, removing them should remove the reading. We tested this with two variants of the training text for the RETFound start, the disease sentence alone, and the biomarker sentences shuffled across images, which keeps their number and form but removes their link to the image. Both were compared with the full text under the disease-balanced sampler. Diagnosis did not change. The AUROC was 0.872 with the full text, 0.878 with the disease sentence alone and 0.879 with shuffled sentences on Drishti-GS, and 0.839, 0.851 and 0.839 on WF\_Glaucoma. The biomarker reading, however, fell on Drishti-GS from 0.385 with the full text to 0.244 with the disease sentence alone, which is below the random guessing level of 0.286, and to 0.300 with shuffled sentences. On WF\_Glaucoma it fell from 0.317 to 0.295 and to 0.271. For the biomarker reading, the seed ranges of the full text did not overlap with those of either variant on either cohort, while for diagnosis they did. The gain depends on the sentences being correct for their image, as well as present.

For the FLAIR start, we could compare the disease sentence alone with the final model on WF\_Glaucoma only. Removing the biomarker sentences lowered the AUROC from 0.905 to 0.893, with non-overlapping seed ranges, and left the mean biomarker reading unchanged, 0.344 against 0.332, with overlapping ranges. The effect of the biomarker text differed between the biomarkers. It raised the readings of the vertical and area cup-to-disc ratios, from 0.317 to 0.384 and from 0.307 to 0.385, but lowered the readings of the rim area fraction, from 0.511 to 0.466, and of the disc vertical-to-horizontal ratio, from 0.363 to 0.196. The biomarker sentences therefore matter most for the cup-to-disc ratios, and the clearest evidence is for the RETFound start.

\emph{The starting model matters for the biomarker reading with RETFound and for diagnosis with FLAIR.} The gains of GlaucomaMarker could come from the pretraining of the retinal models it starts from, with no contribution from the biomarker text. We therefore replaced the starting weights with ImageNet weights, for the same ViT-L architecture and for a ResNet-50, and kept the rest of the pipeline. The ImageNet ViT-L diagnosed about as well as GlaucomaMarker-R, with an AUROC of 0.833 against 0.853 on Drishti-GS and 0.873 against 0.858 on WF\_Glaucoma, and the seed ranges overlapped. It read the biomarkers less accurately, with 0.308 against 0.414 on Drishti-GS and 0.287 against 0.354 on WF\_Glaucoma, and these seed ranges did not overlap. For the ResNet-50, the FLAIR start gave a higher MCC on WF\_Glaucoma, 0.662 against 0.593, with non-overlapping seed ranges, and a higher AUROC, 0.905 against 0.878, although one ImageNet seed matched it. The linear probe was also higher, 0.898 against 0.860, with slightly overlapping ranges. The biomarker reading did not differ, 0.332 against 0.328.

\emph{The equal-bucket sampler removes the preference for common buckets that skewed training data would otherwise teach.} Public glaucoma data are skewed toward healthy eyes and toward the middle bucket of each biomarker, as Supplementary Figure~\ref{fig:bucket_skew} shows, which can lead a model to prefer the common buckets. We addressed this with two measures, the sampler, which acts on the training data, and the auxiliary cosine loss, which acts on the embedding. We measured the preference as the difference between the accuracy and the balanced accuracy of the biomarker reading, averaged over the biomarkers, which is zero for a model that reads all buckets equally well, positive when it favors the common ones, and slightly negative when it reads rare buckets a little better. On the internal validation split, the sampler lowered the difference for the RETFound start from 0.134 with the disease-balanced sampler to 0.007, with non-overlapping seed ranges, and for the FLAIR start from 0.057 with plain shuffling to -0.031. The reduction held on Drishti-GS for the RETFound start, where the difference fell from 0.080 without the loss term and 0.061 with it to -0.042, with non-overlapping ranges. The sampler also gave the highest internal balanced accuracy, 0.522 against 0.482 for RETFound and 0.527 against 0.497 for FLAIR, at a cost of 0.014 in internal AUROC for FLAIR.

The auxiliary cosine loss moved the difference toward zero on the clinical cohort, from 0.030 to -0.018 for RETFound, but the seed ranges overlapped, and on the internal split it changed little, from 0.108 to 0.096. On WF\_Glaucoma the difference was already near zero with the loss term, -0.018 for RETFound and -0.012 for FLAIR, and it stayed there with the sampler, -0.003 and -0.031. Removing the preference did not change diagnosis or the mean biomarker reading beyond the seed ranges. The three losses diagnosed alike, with a WF\_Glaucoma AUROC of 0.830, 0.822 and 0.835 for SupCon, MP-NCE and MP-NCE with the loss term, and for the RETFound start the final pipeline was within the seed ranges of plain shuffling on every external metric, for example an AUROC of 0.858 against 0.835 on WF\_Glaucoma. The equal-bucket sampler made the biomarker reading as accurate on rare buckets as on common ones, without changing diagnosis on the external cohorts.

\begin{table}[htbp]
\centering
\footnotesize
\setlength{\tabcolsep}{3pt}
\caption{Ablation of the GlaucomaMarker training pipeline on zero-shot disease classification, as the mean $\pm$ standard deviation over three seeds. The groups compare the losses, the epoch samplers, the training text, the starting weights and the FLAIR start, as set out in Methods under Ablation design. The text group is read against the disease-balanced sampler row, and the FLAIR group is shown on WF\_Glaucoma only. Bold rows are the final models and bold values are the best entry in each column. Threshold-based metrics use the validation-frozen threshold. Seed values and ranges are in Supplementary Table~\ref{tab:ablation_seed_ranges}.}
\label{tab:ablation_loss_ladder}
\resizebox{\ifdim\width>\linewidth\linewidth\else\width\fi}{!}{%
\begin{tabular}{lcccccccccc}
\toprule
& \multicolumn{5}{c}{Drishti-GS (n=101)} & \multicolumn{5}{c}{WF\_Glaucoma (n=374)} \\
\cmidrule(lr){2-6}\cmidrule(lr){7-11}
Configuration & AUROC & AUPRC & Bal.\ Acc. & Sens. & Spec. & AUROC & AUPRC & Bal.\ Acc. & Sens. & Spec. \\
\midrule
\addlinespace[2pt]
\multicolumn{11}{l}{\scriptsize\textsc{Loss function}} \\
\quad SupCon (Khosla et al. 2020) & \textbf{0.898{\scriptsize$\pm$0.016}} & \textbf{0.945{\scriptsize$\pm$0.012}} & 0.799{\scriptsize$\pm$0.037} & 0.748{\scriptsize$\pm$0.084} & 0.849{\scriptsize$\pm$0.019} & 0.830{\scriptsize$\pm$0.024} & 0.771{\scriptsize$\pm$0.044} & 0.735{\scriptsize$\pm$0.028} & 0.751{\scriptsize$\pm$0.073} & 0.719{\scriptsize$\pm$0.114} \\
\quad MP-NCE (Lee et al. 2022) & 0.861{\scriptsize$\pm$0.012} & 0.923{\scriptsize$\pm$0.005} & 0.775{\scriptsize$\pm$0.043} & 0.657{\scriptsize$\pm$0.124} & 0.892{\scriptsize$\pm$0.037} & 0.822{\scriptsize$\pm$0.043} & 0.767{\scriptsize$\pm$0.061} & 0.745{\scriptsize$\pm$0.032} & 0.679{\scriptsize$\pm$0.071} & 0.811{\scriptsize$\pm$0.054} \\
\quad MP-NCE + auxiliary cosine loss & 0.875{\scriptsize$\pm$0.019} & 0.930{\scriptsize$\pm$0.016} & \textbf{0.799{\scriptsize$\pm$0.011}} & \textbf{0.781{\scriptsize$\pm$0.033}} & 0.817{\scriptsize$\pm$0.049} & 0.835{\scriptsize$\pm$0.023} & 0.772{\scriptsize$\pm$0.036} & 0.758{\scriptsize$\pm$0.018} & 0.727{\scriptsize$\pm$0.040} & 0.790{\scriptsize$\pm$0.074} \\
\addlinespace[2pt]
\multicolumn{11}{l}{\scriptsize\textsc{Epoch sampler (MP-NCE + auxiliary cosine loss)}} \\
\quad Disease-balanced sampler & 0.872{\scriptsize$\pm$0.013} & 0.926{\scriptsize$\pm$0.009} & 0.784{\scriptsize$\pm$0.026} & 0.686{\scriptsize$\pm$0.065} & 0.882{\scriptsize$\pm$0.019} & 0.839{\scriptsize$\pm$0.006} & 0.796{\scriptsize$\pm$0.007} & 0.758{\scriptsize$\pm$0.024} & 0.741{\scriptsize$\pm$0.029} & 0.774{\scriptsize$\pm$0.075} \\
\quad \textbf{Equal-bucket sampler (GlaucomaMarker-R)} & 0.853{\scriptsize$\pm$0.013} & 0.927{\scriptsize$\pm$0.013} & 0.787{\scriptsize$\pm$0.051} & 0.714{\scriptsize$\pm$0.100} & 0.860{\scriptsize$\pm$0.019} & 0.858{\scriptsize$\pm$0.013} & 0.817{\scriptsize$\pm$0.019} & 0.784{\scriptsize$\pm$0.020} & 0.758{\scriptsize$\pm$0.047} & 0.810{\scriptsize$\pm$0.071} \\
\addlinespace[2pt]
\multicolumn{11}{l}{\scriptsize\textsc{Text content (disease-balanced sampler)}} \\
\quad Disease sentence only & 0.878{\scriptsize$\pm$0.011} & 0.939{\scriptsize$\pm$0.001} & 0.787{\scriptsize$\pm$0.026} & 0.681{\scriptsize$\pm$0.050} & 0.892{\scriptsize$\pm$0.019} & 0.851{\scriptsize$\pm$0.019} & 0.804{\scriptsize$\pm$0.018} & 0.774{\scriptsize$\pm$0.022} & 0.707{\scriptsize$\pm$0.071} & 0.840{\scriptsize$\pm$0.032} \\
\quad Biomarker sentences shuffled & 0.879{\scriptsize$\pm$0.012} & 0.937{\scriptsize$\pm$0.003} & 0.763{\scriptsize$\pm$0.054} & 0.633{\scriptsize$\pm$0.125} & 0.892{\scriptsize$\pm$0.019} & 0.839{\scriptsize$\pm$0.014} & 0.802{\scriptsize$\pm$0.018} & 0.751{\scriptsize$\pm$0.012} & 0.746{\scriptsize$\pm$0.133} & 0.756{\scriptsize$\pm$0.130} \\
\addlinespace[2pt]
\multicolumn{11}{l}{\scriptsize\textsc{Image-encoder initialization (final pipeline)}} \\
\quad RETFound ViT-L/16 (final model) & 0.853{\scriptsize$\pm$0.013} & 0.927{\scriptsize$\pm$0.013} & 0.787{\scriptsize$\pm$0.051} & 0.714{\scriptsize$\pm$0.100} & 0.860{\scriptsize$\pm$0.019} & 0.858{\scriptsize$\pm$0.013} & 0.817{\scriptsize$\pm$0.019} & 0.784{\scriptsize$\pm$0.020} & 0.758{\scriptsize$\pm$0.047} & 0.810{\scriptsize$\pm$0.071} \\
\quad ImageNet ViT-L/16 (same architecture) & 0.833{\scriptsize$\pm$0.024} & 0.928{\scriptsize$\pm$0.011} & 0.776{\scriptsize$\pm$0.029} & 0.714{\scriptsize$\pm$0.089} & 0.839{\scriptsize$\pm$0.032} & 0.873{\scriptsize$\pm$0.009} & 0.845{\scriptsize$\pm$0.005} & 0.780{\scriptsize$\pm$0.009} & 0.671{\scriptsize$\pm$0.058} & 0.888{\scriptsize$\pm$0.040} \\
\quad ImageNet ResNet-50 (different architecture) & 0.860{\scriptsize$\pm$0.011} & 0.937{\scriptsize$\pm$0.002} & 0.776{\scriptsize$\pm$0.023} & 0.638{\scriptsize$\pm$0.081} & \textbf{0.914{\scriptsize$\pm$0.037}} & 0.878{\scriptsize$\pm$0.022} & 0.837{\scriptsize$\pm$0.029} & 0.789{\scriptsize$\pm$0.020} & 0.698{\scriptsize$\pm$0.075} & 0.879{\scriptsize$\pm$0.037} \\
\addlinespace[2pt]
\multicolumn{11}{l}{\scriptsize\textsc{FLAIR start, WF\_Glaucoma only}} \\
\quad Plain shuffling, no auxiliary cosine loss &  &  &  &  &  & 0.895{\scriptsize$\pm$0.005} & 0.882{\scriptsize$\pm$0.001} & \textbf{0.831{\scriptsize$\pm$0.009}} & 0.751{\scriptsize$\pm$0.034} & 0.912{\scriptsize$\pm$0.024} \\
\quad Plain shuffling + auxiliary cosine loss &  &  &  &  &  & 0.907{\scriptsize$\pm$0.008} & 0.888{\scriptsize$\pm$0.005} & 0.825{\scriptsize$\pm$0.005} & 0.731{\scriptsize$\pm$0.008} & \textbf{0.919{\scriptsize$\pm$0.004}} \\
\quad Disease-balanced sampler + auxiliary cosine loss &  &  &  &  &  & \textbf{0.912{\scriptsize$\pm$0.009}} & \textbf{0.890{\scriptsize$\pm$0.006}} & 0.824{\scriptsize$\pm$0.008} & \textbf{0.775{\scriptsize$\pm$0.032}} & 0.874{\scriptsize$\pm$0.017} \\
\quad \textbf{Equal-bucket sampler (GlaucomaMarker-F)} &  &  &  &  &  & 0.905{\scriptsize$\pm$0.001} & 0.885{\scriptsize$\pm$0.002} & 0.818{\scriptsize$\pm$0.003} & 0.717{\scriptsize$\pm$0.029} & \textbf{0.919{\scriptsize$\pm$0.034}} \\
\quad Disease sentence only (disease-balanced sampler) &  &  &  &  &  & 0.893{\scriptsize$\pm$0.003} & 0.874{\scriptsize$\pm$0.002} & 0.828{\scriptsize$\pm$0.009} & 0.758{\scriptsize$\pm$0.015} & 0.898{\scriptsize$\pm$0.019} \\
\botrule
\end{tabular}}
\end{table}

\begin{table}[htbp]
\centering
\footnotesize
\setlength{\tabcolsep}{3pt}
\caption{Ablation of GlaucomaMarker's training pipeline on zero-shot disease classification: operating-point metrics at the validation-frozen threshold. Configurations, seeds, and checkpoint selection as in Table~\ref{tab:ablation_loss_ladder}; bold marks the best entry in each column.}
\label{tab:ablation_operating_point}
\resizebox{\ifdim\width>\linewidth\linewidth\else\width\fi}{!}{%
\begin{tabular}{lcccccccccccc}
\toprule
& \multicolumn{6}{c}{Drishti-GS (n=101)} & \multicolumn{6}{c}{WF\_Glaucoma (n=374)} \\
\cmidrule(lr){2-7}\cmidrule(lr){8-13}
Configuration & Acc. & Macro F1 & PPV & NPV & MCC & $\kappa$ & Acc. & Macro F1 & PPV & NPV & MCC & $\kappa$ \\
\midrule
\addlinespace[2pt]
\multicolumn{13}{l}{\scriptsize\textsc{Loss function}} \\
\quad SupCon (Khosla et al. 2020) & 0.779{\scriptsize$\pm$0.055} & 0.764{\scriptsize$\pm$0.051} & 0.918{\scriptsize$\pm$0.008} & 0.606{\scriptsize$\pm$0.084} & 0.559{\scriptsize$\pm$0.081} & 0.538{\scriptsize$\pm$0.094} & 0.731{\scriptsize$\pm$0.048} & 0.722{\scriptsize$\pm$0.042} & 0.622{\scriptsize$\pm$0.074} & 0.832{\scriptsize$\pm$0.025} & 0.462{\scriptsize$\pm$0.059} & 0.451{\scriptsize$\pm$0.072} \\
\quad MP-NCE (Lee et al. 2022) & 0.729{\scriptsize$\pm$0.074} & 0.720{\scriptsize$\pm$0.067} & 0.934{\scriptsize$\pm$0.012} & 0.545{\scriptsize$\pm$0.073} & 0.513{\scriptsize$\pm$0.073} & 0.469{\scriptsize$\pm$0.103} & 0.762{\scriptsize$\pm$0.030} & 0.745{\scriptsize$\pm$0.032} & 0.684{\scriptsize$\pm$0.051} & 0.811{\scriptsize$\pm$0.029} & 0.492{\scriptsize$\pm$0.061} & 0.490{\scriptsize$\pm$0.063} \\
\quad MP-NCE + auxiliary cosine loss & \textbf{0.792{\scriptsize$\pm$0.010}} & \textbf{0.773{\scriptsize$\pm$0.007}} & 0.907{\scriptsize$\pm$0.020} & \textbf{0.624{\scriptsize$\pm$0.021}} & \textbf{0.564{\scriptsize$\pm$0.015}} & \textbf{0.550{\scriptsize$\pm$0.014}} & 0.766{\scriptsize$\pm$0.032} & 0.754{\scriptsize$\pm$0.027} & 0.678{\scriptsize$\pm$0.062} & 0.831{\scriptsize$\pm$0.008} & 0.513{\scriptsize$\pm$0.045} & 0.509{\scriptsize$\pm$0.050} \\
\addlinespace[2pt]
\multicolumn{13}{l}{\scriptsize\textsc{Epoch sampler (MP-NCE + auxiliary cosine loss)}} \\
\quad Disease-balanced sampler & 0.746{\scriptsize$\pm$0.041} & 0.735{\scriptsize$\pm$0.038} & 0.929{\scriptsize$\pm$0.007} & 0.557{\scriptsize$\pm$0.050} & 0.525{\scriptsize$\pm$0.051} & 0.489{\scriptsize$\pm$0.065} & 0.762{\scriptsize$\pm$0.037} & 0.751{\scriptsize$\pm$0.032} & 0.667{\scriptsize$\pm$0.070} & 0.835{\scriptsize$\pm$0.003} & 0.508{\scriptsize$\pm$0.058} & 0.504{\scriptsize$\pm$0.062} \\
\quad \textbf{Equal-bucket sampler (GlaucomaMarker-R)} & 0.759{\scriptsize$\pm$0.070} & 0.746{\scriptsize$\pm$0.066} & 0.919{\scriptsize$\pm$0.013} & 0.580{\scriptsize$\pm$0.088} & 0.536{\scriptsize$\pm$0.101} & 0.508{\scriptsize$\pm$0.117} & 0.791{\scriptsize$\pm$0.031} & 0.779{\scriptsize$\pm$0.028} & 0.708{\scriptsize$\pm$0.063} & 0.851{\scriptsize$\pm$0.016} & 0.563{\scriptsize$\pm$0.047} & 0.560{\scriptsize$\pm$0.052} \\
\addlinespace[2pt]
\multicolumn{13}{l}{\scriptsize\textsc{Text content (disease-balanced sampler)}} \\
\quad Disease sentence only & 0.746{\scriptsize$\pm$0.035} & 0.736{\scriptsize$\pm$0.033} & 0.935{\scriptsize$\pm$0.011} & 0.555{\scriptsize$\pm$0.039} & 0.530{\scriptsize$\pm$0.048} & 0.491{\scriptsize$\pm$0.057} & 0.791{\scriptsize$\pm$0.011} & 0.774{\scriptsize$\pm$0.016} & 0.725{\scriptsize$\pm$0.022} & 0.830{\scriptsize$\pm$0.028} & 0.551{\scriptsize$\pm$0.029} & 0.549{\scriptsize$\pm$0.032} \\
\quad Biomarker sentences shuffled & 0.713{\scriptsize$\pm$0.081} & 0.705{\scriptsize$\pm$0.075} & 0.930{\scriptsize$\pm$0.004} & 0.529{\scriptsize$\pm$0.088} & 0.491{\scriptsize$\pm$0.098} & 0.444{\scriptsize$\pm$0.123} & 0.752{\scriptsize$\pm$0.034} & 0.740{\scriptsize$\pm$0.022} & 0.663{\scriptsize$\pm$0.099} & 0.842{\scriptsize$\pm$0.054} & 0.503{\scriptsize$\pm$0.040} & 0.488{\scriptsize$\pm$0.041} \\
\addlinespace[2pt]
\multicolumn{13}{l}{\scriptsize\textsc{Image-encoder initialization (final pipeline)}} \\
\quad RETFound ViT-L/16 (final model) & 0.759{\scriptsize$\pm$0.070} & 0.746{\scriptsize$\pm$0.066} & 0.919{\scriptsize$\pm$0.013} & 0.580{\scriptsize$\pm$0.088} & 0.536{\scriptsize$\pm$0.101} & 0.508{\scriptsize$\pm$0.117} & 0.791{\scriptsize$\pm$0.031} & 0.779{\scriptsize$\pm$0.028} & 0.708{\scriptsize$\pm$0.063} & 0.851{\scriptsize$\pm$0.016} & 0.563{\scriptsize$\pm$0.047} & 0.560{\scriptsize$\pm$0.052} \\
\quad ImageNet ViT-L/16 (same architecture) & 0.752{\scriptsize$\pm$0.052} & 0.738{\scriptsize$\pm$0.046} & 0.910{\scriptsize$\pm$0.007} & 0.572{\scriptsize$\pm$0.065} & 0.516{\scriptsize$\pm$0.059} & 0.490{\scriptsize$\pm$0.077} & 0.807{\scriptsize$\pm$0.005} & 0.787{\scriptsize$\pm$0.004} & 0.785{\scriptsize$\pm$0.048} & 0.821{\scriptsize$\pm$0.019} & 0.582{\scriptsize$\pm$0.006} & 0.576{\scriptsize$\pm$0.006} \\
\quad ImageNet ResNet-50 (different architecture) & 0.723{\scriptsize$\pm$0.045} & 0.715{\scriptsize$\pm$0.040} & \textbf{0.945{\scriptsize$\pm$0.015}} & 0.532{\scriptsize$\pm$0.049} & 0.513{\scriptsize$\pm$0.041} & 0.461{\scriptsize$\pm$0.063} & 0.812{\scriptsize$\pm$0.007} & 0.794{\scriptsize$\pm$0.012} & 0.777{\scriptsize$\pm$0.035} & 0.833{\scriptsize$\pm$0.031} & 0.593{\scriptsize$\pm$0.021} & 0.589{\scriptsize$\pm$0.024} \\
\addlinespace[2pt]
\multicolumn{13}{l}{\scriptsize\textsc{FLAIR start, WF\_Glaucoma only}} \\
\quad Plain shuffling, no auxiliary cosine loss &  &  &  &  &  &  & \textbf{0.852{\scriptsize$\pm$0.008}} & \textbf{0.838{\scriptsize$\pm$0.009}} & 0.836{\scriptsize$\pm$0.032} & 0.861{\scriptsize$\pm$0.014} & \textbf{0.680{\scriptsize$\pm$0.017}} & \textbf{0.677{\scriptsize$\pm$0.017}} \\
\quad Plain shuffling + auxiliary cosine loss &  &  &  &  &  &  & 0.849{\scriptsize$\pm$0.004} & 0.834{\scriptsize$\pm$0.005} & 0.843{\scriptsize$\pm$0.007} & 0.853{\scriptsize$\pm$0.004} & 0.673{\scriptsize$\pm$0.009} & 0.669{\scriptsize$\pm$0.009} \\
\quad Disease-balanced sampler + auxiliary cosine loss &  &  &  &  &  &  & 0.837{\scriptsize$\pm$0.003} & 0.825{\scriptsize$\pm$0.004} & 0.785{\scriptsize$\pm$0.016} & \textbf{0.868{\scriptsize$\pm$0.014}} & 0.650{\scriptsize$\pm$0.008} & 0.650{\scriptsize$\pm$0.009} \\
\quad \textbf{Equal-bucket sampler (GlaucomaMarker-F)} &  &  &  &  &  &  & 0.844{\scriptsize$\pm$0.011} & 0.827{\scriptsize$\pm$0.008} & \textbf{0.843{\scriptsize$\pm$0.052}} & 0.846{\scriptsize$\pm$0.009} & 0.662{\scriptsize$\pm$0.024} & 0.656{\scriptsize$\pm$0.018} \\
\quad Disease sentence only (disease-balanced sampler) &  &  &  &  &  &  & 0.846{\scriptsize$\pm$0.011} & 0.832{\scriptsize$\pm$0.011} & 0.815{\scriptsize$\pm$0.026} & 0.862{\scriptsize$\pm$0.007} & 0.666{\scriptsize$\pm$0.022} & 0.665{\scriptsize$\pm$0.022} \\
\botrule
\end{tabular}}
\end{table}

\begin{table}[htbp]
\centering
\footnotesize
\setlength{\tabcolsep}{3pt}
\caption{Ablation of GlaucomaMarker's training pipeline on biomarker concept accuracy, linear probing, and cross-modal retrieval. Bio is the mean zero-shot balanced accuracy over the biomarker concepts (11 on Drishti-GS, 5 on WF\_Glaucoma); LP is the full-data linear-probe AUROC and AUPRC; retrieval is image-to-text R@1 (I$\to$T), text-to-image R@1 (T$\to$I), and text-to-image P@5. WF\_Glaucoma retrieval uses the 444 images with OCT ground truth. Configurations, seeds, and checkpoint selection as in Table~\ref{tab:ablation_loss_ladder}; bold marks the best entry in each column.}
\label{tab:ablation_representation}
\resizebox{\ifdim\width>\linewidth\linewidth\else\width\fi}{!}{%
\begin{tabular}{lcccccccccccc}
\toprule
& \multicolumn{6}{c}{Drishti-GS (n=101)} & \multicolumn{6}{c}{WF\_Glaucoma (n=374)} \\
\cmidrule(lr){2-7}\cmidrule(lr){8-13}
Configuration & Bio & LP AUROC & LP AUPRC & I$\to$T R@1 & T$\to$I R@1 & T$\to$I P@5 & Bio & LP AUROC & LP AUPRC & I$\to$T R@1 & T$\to$I R@1 & T$\to$I P@5 \\
\midrule
\addlinespace[2pt]
\multicolumn{13}{l}{\scriptsize\textsc{Loss function}} \\
\quad SupCon (Khosla et al. 2020) & 0.388{\scriptsize$\pm$0.044} & \textbf{0.884{\scriptsize$\pm$0.010}} & \textbf{0.940{\scriptsize$\pm$0.012}} & 0.733{\scriptsize$\pm$0.062} & 0.477{\scriptsize$\pm$0.087} & 0.443{\scriptsize$\pm$0.019} & 0.335{\scriptsize$\pm$0.009} & 0.785{\scriptsize$\pm$0.064} & 0.712{\scriptsize$\pm$0.092} & 0.616{\scriptsize$\pm$0.031} & \textbf{0.470{\scriptsize$\pm$0.069}} & 0.397{\scriptsize$\pm$0.023} \\
\quad MP-NCE (Lee et al. 2022) & 0.409{\scriptsize$\pm$0.022} & 0.851{\scriptsize$\pm$0.020} & 0.911{\scriptsize$\pm$0.013} & 0.733{\scriptsize$\pm$0.075} & \textbf{0.514{\scriptsize$\pm$0.081}} & 0.445{\scriptsize$\pm$0.008} & 0.338{\scriptsize$\pm$0.006} & 0.793{\scriptsize$\pm$0.039} & 0.727{\scriptsize$\pm$0.077} & 0.633{\scriptsize$\pm$0.033} & 0.318{\scriptsize$\pm$0.091} & 0.415{\scriptsize$\pm$0.029} \\
\quad MP-NCE + auxiliary cosine loss & \textbf{0.430{\scriptsize$\pm$0.019}} & 0.867{\scriptsize$\pm$0.029} & 0.930{\scriptsize$\pm$0.021} & \textbf{0.809{\scriptsize$\pm$0.023}} & \textbf{0.514{\scriptsize$\pm$0.027}} & 0.452{\scriptsize$\pm$0.016} & 0.328{\scriptsize$\pm$0.030} & 0.825{\scriptsize$\pm$0.018} & 0.771{\scriptsize$\pm$0.020} & 0.638{\scriptsize$\pm$0.025} & 0.379{\scriptsize$\pm$0.052} & \textbf{0.433{\scriptsize$\pm$0.010}} \\
\addlinespace[2pt]
\multicolumn{13}{l}{\scriptsize\textsc{Epoch sampler (MP-NCE + auxiliary cosine loss)}} \\
\quad Disease-balanced sampler & 0.385{\scriptsize$\pm$0.024} & 0.869{\scriptsize$\pm$0.007} & 0.929{\scriptsize$\pm$0.008} & 0.723{\scriptsize$\pm$0.026} & 0.432{\scriptsize$\pm$0.027} & 0.416{\scriptsize$\pm$0.005} & 0.317{\scriptsize$\pm$0.008} & 0.825{\scriptsize$\pm$0.010} & 0.781{\scriptsize$\pm$0.008} & 0.634{\scriptsize$\pm$0.020} & 0.348{\scriptsize$\pm$0.095} & 0.370{\scriptsize$\pm$0.045} \\
\quad \textbf{Equal-bucket sampler (GlaucomaMarker-R)} & 0.414{\scriptsize$\pm$0.024} & 0.841{\scriptsize$\pm$0.037} & 0.914{\scriptsize$\pm$0.032} & 0.756{\scriptsize$\pm$0.055} & 0.468{\scriptsize$\pm$0.087} & \textbf{0.463{\scriptsize$\pm$0.017}} & 0.354{\scriptsize$\pm$0.020} & 0.850{\scriptsize$\pm$0.011} & 0.802{\scriptsize$\pm$0.015} & 0.651{\scriptsize$\pm$0.010} & 0.379{\scriptsize$\pm$0.052} & 0.382{\scriptsize$\pm$0.045} \\
\addlinespace[2pt]
\multicolumn{13}{l}{\scriptsize\textsc{Text content (disease-balanced sampler)}} \\
\quad Disease sentence only & 0.244{\scriptsize$\pm$0.091} & 0.875{\scriptsize$\pm$0.012} & 0.938{\scriptsize$\pm$0.002} & 0.492{\scriptsize$\pm$0.149} & 0.396{\scriptsize$\pm$0.087} & 0.353{\scriptsize$\pm$0.043} & 0.295{\scriptsize$\pm$0.014} & 0.798{\scriptsize$\pm$0.070} & 0.736{\scriptsize$\pm$0.100} & 0.433{\scriptsize$\pm$0.190} & 0.318{\scriptsize$\pm$0.045} & 0.336{\scriptsize$\pm$0.031} \\
\quad Biomarker sentences shuffled & 0.300{\scriptsize$\pm$0.013} & 0.873{\scriptsize$\pm$0.008} & 0.934{\scriptsize$\pm$0.002} & 0.713{\scriptsize$\pm$0.020} & 0.387{\scriptsize$\pm$0.031} & 0.337{\scriptsize$\pm$0.020} & 0.271{\scriptsize$\pm$0.012} & 0.830{\scriptsize$\pm$0.024} & 0.789{\scriptsize$\pm$0.023} & 0.633{\scriptsize$\pm$0.028} & 0.318{\scriptsize$\pm$0.000} & 0.309{\scriptsize$\pm$0.016} \\
\addlinespace[2pt]
\multicolumn{13}{l}{\scriptsize\textsc{Image-encoder initialization (final pipeline)}} \\
\quad RETFound ViT-L/16 (final model) & 0.414{\scriptsize$\pm$0.024} & 0.841{\scriptsize$\pm$0.037} & 0.914{\scriptsize$\pm$0.032} & 0.756{\scriptsize$\pm$0.055} & 0.468{\scriptsize$\pm$0.087} & \textbf{0.463{\scriptsize$\pm$0.017}} & 0.354{\scriptsize$\pm$0.020} & 0.850{\scriptsize$\pm$0.011} & 0.802{\scriptsize$\pm$0.015} & 0.651{\scriptsize$\pm$0.010} & 0.379{\scriptsize$\pm$0.052} & 0.382{\scriptsize$\pm$0.045} \\
\quad ImageNet ViT-L/16 (same architecture) & 0.308{\scriptsize$\pm$0.044} & 0.844{\scriptsize$\pm$0.009} & 0.920{\scriptsize$\pm$0.011} & 0.749{\scriptsize$\pm$0.051} & 0.441{\scriptsize$\pm$0.062} & 0.422{\scriptsize$\pm$0.052} & 0.287{\scriptsize$\pm$0.016} & 0.867{\scriptsize$\pm$0.013} & 0.833{\scriptsize$\pm$0.024} & 0.673{\scriptsize$\pm$0.006} & 0.333{\scriptsize$\pm$0.069} & 0.324{\scriptsize$\pm$0.014} \\
\quad ImageNet ResNet-50 (different architecture) & 0.307{\scriptsize$\pm$0.016} & 0.872{\scriptsize$\pm$0.003} & 0.934{\scriptsize$\pm$0.001} & 0.739{\scriptsize$\pm$0.032} & 0.342{\scriptsize$\pm$0.031} & 0.447{\scriptsize$\pm$0.034} & 0.328{\scriptsize$\pm$0.012} & 0.860{\scriptsize$\pm$0.033} & 0.805{\scriptsize$\pm$0.055} & 0.674{\scriptsize$\pm$0.018} & \textbf{0.470{\scriptsize$\pm$0.095}} & 0.427{\scriptsize$\pm$0.009} \\
\addlinespace[2pt]
\multicolumn{13}{l}{\scriptsize\textsc{FLAIR start, WF\_Glaucoma only}} \\
\quad Plain shuffling, no auxiliary cosine loss &  &  &  &  &  &  & 0.352{\scriptsize$\pm$0.002} & 0.878{\scriptsize$\pm$0.011} & 0.870{\scriptsize$\pm$0.008} & 0.711{\scriptsize$\pm$0.003} & 0.379{\scriptsize$\pm$0.026} & 0.406{\scriptsize$\pm$0.028} \\
\quad Plain shuffling + auxiliary cosine loss &  &  &  &  &  &  & 0.339{\scriptsize$\pm$0.019} & 0.896{\scriptsize$\pm$0.019} & 0.881{\scriptsize$\pm$0.013} & 0.712{\scriptsize$\pm$0.002} & 0.364{\scriptsize$\pm$0.045} & 0.382{\scriptsize$\pm$0.009} \\
\quad Disease-balanced sampler + auxiliary cosine loss &  &  &  &  &  &  & \textbf{0.356{\scriptsize$\pm$0.010}} & \textbf{0.904{\scriptsize$\pm$0.012}} & \textbf{0.886{\scriptsize$\pm$0.007}} & 0.704{\scriptsize$\pm$0.011} & 0.409{\scriptsize$\pm$0.045} & 0.391{\scriptsize$\pm$0.009} \\
\quad \textbf{Equal-bucket sampler (GlaucomaMarker-F)} &  &  &  &  &  &  & 0.332{\scriptsize$\pm$0.014} & 0.898{\scriptsize$\pm$0.009} & 0.882{\scriptsize$\pm$0.004} & \textbf{0.715{\scriptsize$\pm$0.006}} & 0.424{\scriptsize$\pm$0.026} & 0.391{\scriptsize$\pm$0.033} \\
\quad Disease sentence only (disease-balanced sampler) &  &  &  &  &  &  & 0.344{\scriptsize$\pm$0.009} & 0.889{\scriptsize$\pm$0.002} & 0.877{\scriptsize$\pm$0.002} & 0.705{\scriptsize$\pm$0.006} & 0.394{\scriptsize$\pm$0.069} & 0.333{\scriptsize$\pm$0.014} \\
\botrule
\end{tabular}}
\end{table}

\section*{Discussion}

We showed that text computed from public segmentation masks can teach retinal foundation models to read glaucoma biomarkers. Every sentence can be traced to the mask pixels it came from. We applied the framework to RETFound and FLAIR and evaluated both instances against seven baselines on two external cohorts. On the clinical cohort, GlaucomaMarker-F diagnosed glaucoma better than FLAIR, with an AUROC of 0.905 against 0.848 and an MCC of 0.662 against 0.473, and GlaucomaMarker-R was comparable to FLAIR. Both instances read the biomarkers above the random guessing level, where the baselines with a biomarker readout stayed close to it.

GlaucomaMarker differs from earlier retinal vision-language models in the origin of its training text. FLAIR and EyeCLIP learn from category labels, templates and captions, so their text states what a label or a writer says about an image and contains no quantity measured on it. Reports written by clinicians are costly to collect and differ in what they mention, and captions generated by a language model describe a category in general. GlaucomaMarker instead computes each sentence from the optic disc and cup masks, so each sentence states a measurement. As a result, the model can read the bucket of a measured quantity, such as the vertical cup-to-disc ratio, in addition to the diagnosis, and because the output is a fixed sentence, a reader can score it against the measured value, and against independent OCT on the clinical cohort, without any clinician grading.

The biomarker reading depends on the biomarker sentences in the training text. For GlaucomaMarker-R, removing them left diagnosis unchanged, with an AUROC of 0.878 against 0.872 on Drishti-GS, but lowered the biomarker reading from 0.385 to 0.244, which is below the random guessing level of 0.286. Shuffling the sentences across images gave 0.300, so the reading depends on the sentences being correct for their image. For GlaucomaMarker-F, the sentences raised the readings of the vertical and area cup-to-disc ratios, from 0.317 to 0.384 and from 0.307 to 0.385, and the AUROC from 0.893 to 0.905. The reading also follows the measured structure beyond the diagnosis, since its correlation with the true bucket was positive inside healthy images and inside glaucoma images separately, for example 0.71 and 0.57 for the vertical cup-to-disc ratio on Drishti-GS.

Based on our analysis of public glaucoma data, which Supplementary Figure~\ref{fig:bucket_skew} shows, the images are skewed toward healthy eyes and, for most biomarkers, toward a middle bucket, so a model trained on them can learn to prefer the common buckets. The equal-bucket sampler was designed to remove this preference, which we measured as the difference between the accuracy and the balanced accuracy of the biomarker reading, averaged over the biomarkers, so that zero means all buckets are read equally well. For the RETFound start, the difference moved from 0.134 with the disease-balanced sampler to 0.007 on the internal validation split, and from 0.061 with plain shuffling to -0.042 on Drishti-GS, with non-overlapping seed ranges in both, and for the FLAIR start it moved from 0.057 to -0.031 on the internal split. The auxiliary cosine loss, which acts on the embedding, was followed by a smaller difference on the clinical cohort, from 0.030 to -0.018 for RETFound, but the seed ranges overlapped. For the RETFound start, diagnosis stayed within the seed ranges of plain shuffling on both external cohorts, so the sampler had no cost to diagnosis.

The framework also improved the retinal models beyond the biomarker reading. The released RETFound glaucoma checkpoint did not transfer to the clinical cohort, with an AUROC of 0.442, and GlaucomaMarker-R, which starts from RETFound, reached 0.858 and became comparable to FLAIR at 0.848. GlaucomaMarker-F, which starts from FLAIR, went beyond it, with an AUROC of 0.905 and significant differences in AUROC and MCC. The frozen features gained as well, with the share of nearest neighbors that carry an image's label rising from 0.603 to 0.763 for RETFound, and with a linear probe on two labeled images per class reaching 0.905 for GlaucomaMarker-F against 0.832 for FLAIR. Image-to-text retrieval on WF\_Glaucoma followed the same order, led by GlaucomaMarker-F, then GlaucomaMarker-R, then FLAIR.

The framework applied to two different retinal models, and the starting model determined which capability gained most. For RETFound, the retinal pretraining mattered for the biomarker reading, which was 0.414 against 0.308 for an ImageNet ViT-L on Drishti-GS and 0.354 against 0.287 on WF\_Glaucoma, with non-overlapping seed ranges, while diagnosis was similar. For FLAIR, the retinal pretraining mattered for diagnosis and the embedding, with an MCC of 0.662 against 0.593 for an ImageNet ResNet-50 on WF\_Glaucoma and a probe AUROC of 0.898 against 0.860, while the biomarker reading did not differ. The framework applies to a ViT-L with BioBERT and to a ResNet-50 with Bio\_ClinicalBERT alike, and adds a biomarker reading to either.

The biomarker reading is limited by the data that taught it. The biomarker sentences come from the 3,238 training images with masks, and some buckets have almost no examples, with 33 images of horizontal disc elongation, 9 of moderate or marked rim asymmetry and 4 of severe rim loss. The reading was also weakest in the middle buckets, with a recall of 0.205 for GlaucomaMarker-R and 0.103 for GlaucomaMarker-F on WF\_Glaucoma against 0.505 and 0.580 for the extreme buckets, and the disc shape and rim asymmetry biomarkers were the weakest overall. For GlaucomaMarker-F, the biomarker text raised only the cup-to-disc readings and lowered the readings of the rim area fraction and the disc vertical-to-horizontal ratio, so its gain is specific to the cup-to-disc ratios. Our evidence also rests on two external cohorts, of which the clinical one comes from a single institution, where White patients account for 295 of the 374 labeled eyes, and Drishti-GS could not be used for GlaucomaMarker-F because its pretraining overlaps with that cohort. Future work should add training images for the rare and the middle buckets, and evaluate on cohorts from more institutions and more diverse populations.

GlaucomaMarker shows that the text missing for glaucoma can be computed from public segmentation masks, and that training on it gives retinal foundation models a zero-shot biomarker reading while maintaining high diagnostic performance. The same approach can be applied to other diseases that have segmentation masks and no paired reports.

\section*{Methods}

GlaucomaMarker is the three-stage framework shown in Figure~\ref{fig:architecture_schematic} and summarized in the first subsection of Results. This section follows the framework and then its evaluation. It describes how the text is computed from the masks, the training data, the model and its losses and the epoch sampler, and then the evaluation cohorts, the comparison models, the zero-shot readout, the ablations, the statistical analysis and the implementation details.

\subsection*{Problem setup}
\label{sec:method-problem}

The training set contains triplets $\mathcal{D} = \{(x_i, y_i, m_i)\}_{i=1}^{N}$, where $x_i$ is a fundus photograph, $y_i \in \{0, 1\}$ is a binary glaucoma label that is missing for some images, and $m_i$ is an expert mask of the optic disc and cup that exists for a subset of images. From $m_i$ we compute a vector of eleven biomarkers $b_i$ and turn $(y_i, b_i)$ into a set of sentences $T_i$ from a fixed catalog of 48 sentences. The model has an image encoder $f_v$ and a text encoder $f_t$, each followed by a projection into a shared $d$-dimensional space. Matching image and sentence embeddings are pulled together and the others are pushed apart. At test time an image has no mask and no label, and we read the diagnosis and the biomarkers by comparing its embedding with the embeddings of the fixed sentences.

\subsection*{Biomarker text from segmentation masks}
\label{sec:method-text}

Glaucoma is assessed from the measurable structure of the optic nerve head. Our proposed framework prepares text that states these measurements. As defined in Table~\ref{tab:appendix_bucket_thresholds}, for every image with a disc and cup mask, we computed eleven biomarkers directly from the mask pixels. The diameters are the vertical and horizontal extent of each mask boundary, and the areas are pixel counts. The rim width at each of 360 angles around the disc centroid is the difference between the disc radius and the cup radius, and the rim thickness of the superior and inferior quadrants is the mean rim width over 45 to 135 degrees and over 225 to 315 degrees, divided by the vertical disc diameter. The eccentricity of the disc and of the cup comes from an ellipse fitted to each boundary. The biomarkers are the cup-to-disc ratio measured vertically, horizontally and by area, the rim area fraction, the inferior and superior rim thickness, the vertical asymmetry of the rim, the vertical-to-horizontal ratio of the disc and of the cup, and the eccentricity of the disc and of the cup. We split each biomarker into three to five buckets with predefined cutoffs that we set before training and did not fit to our data. The cutoffs were informed by clinical descriptions of the optic nerve head, which include a vertical cup-to-disc ratio of 0.6 as a screening cutoff~\citep{soh2020}, the upper tail of the normal range of this ratio as the limit of normal~\citep{foster2002}, the rim width relative to the disc diameter as a measure of rim loss~\citep{spaeth2002}, and the vertically oval disc and horizontally oval cup of normal eyes~\citep{gandhi2013}. The vertical cup-to-disc ratio, for example, was split at 0.30, 0.40, 0.60 and 0.75 into very small, small, moderate, large and very large. These buckets are conventions for turning measurements into words, so the biomarker accuracy we report is agreement with these definitions.

Each pair of biomarker and bucket was then mapped to a template sentence from a fixed catalog of 48 sentences, given in Table~\ref{tab:appendix_prompt_catalog}. The catalog holds two disease sentences, one for a glaucomatous and one for a healthy appearance, and three to five bucket sentences for each biomarker. An image with a mask therefore carried one disease sentence, when its label was known, and up to eleven biomarker sentences. No free text was written or generated for individual images.

Because each sentence comes from an explicit computation, we could check the extraction without training a model. Our vertical cup-to-disc ratio agreed with the expert grading published with ORIGA \citep{zhang2010origa} to within 0.01 on all 650 images, with a mean absolute error of 0.0012 and a maximum of 0.0024.

\begin{table}[htbp]
\centering
\footnotesize
\caption{Definition and bucket thresholds of the eleven mask-derived biomarkers. Diameters and areas are measured on the expert disc and cup masks. The area cutoffs are the squares of the cup-to-disc cutoffs, and the rim-thickness cutoffs are derived from the cup-to-disc cutoffs through rim width $\approx 0.5\,(1-\text{vertical CDR})$. $a$ and $b$ are the major and minor axes of the fitted ellipse.}
\label{tab:appendix_bucket_thresholds}
\begin{tabular}{p{2.2cm}p{4.3cm}p{5.2cm}}
\toprule
Biomarker & Definition & Buckets (thresholds) \\
\midrule
Vertical CDR & cup vertical diameter / disc vertical diameter & very small ($<$0.30), small (0.30--0.40), moderate (0.40--0.60), large (0.60--0.75), very large ($\geq$ 0.75) \\
Horizontal CDR & cup horizontal diameter / disc horizontal diameter & very small ($<$0.30), small (0.30--0.40), moderate (0.40--0.60), large (0.60--0.75), very large ($\geq$ 0.75) \\
Area CDR & cup area / disc area & low ($<$0.09), moderate (0.09--0.16), high (0.16--0.36), very high ($\geq$ 0.36) \\
Rim area fraction & (disc area $-$ cup area) / disc area & preserved ($>$0.75), mild loss (0.50--0.75), moderate loss (0.25--0.50), severe loss ($\leq$ 0.25) \\
Inferior rim thickness & mean rim width in the inferior quadrant ($225^\circ$--$315^\circ$) / disc vertical diameter & very thin ($<$0.125), thin (0.125--0.20), moderate (0.20--0.30), thick (0.30--0.35), very thick ($\geq$ 0.35) \\
Superior rim thickness & mean rim width in the superior quadrant ($45^\circ$--$135^\circ$) / disc vertical diameter & very thin ($<$0.125), thin (0.125--0.20), moderate (0.20--0.30), thick (0.30--0.35), very thick ($\geq$ 0.35) \\
Rim vertical asymmetry & $|$inferior $-$ superior$|$ normalized rim thickness & none ($<$0.10), mild (0.10--0.20), moderate (0.20--0.35), marked ($\geq$ 0.35) \\
Disc V/H ratio & disc vertical / horizontal diameter & horizontal elongation ($<$0.90), round (0.90--1.10), vertical elongation ($>$1.10) \\
Cup V/H ratio & cup vertical / horizontal diameter & horizontal elongation ($<$0.90), round (0.90--1.10), vertical elongation ($>$1.10) \\
Disc eccentricity & $\sqrt{1-(b/a)^2}$ of the ellipse fitted to the disc boundary & round ($<$0.30), slightly elliptic (0.30--0.50), elliptic (0.50--0.70), highly elliptic ($\geq$ 0.70) \\
Cup eccentricity & $\sqrt{1-(b/a)^2}$ of the ellipse fitted to the cup boundary & round ($<$0.30), slightly elliptic (0.30--0.50), elliptic (0.50--0.70), highly elliptic ($\geq$ 0.70) \\
\botrule
\end{tabular}
\end{table}

\begin{longtable}{p{2.4cm}p{2.3cm}p{6.8cm}}
\caption{The 48-prompt catalog: the disease sentences and one sentence per biomarker bucket. Every training sentence and every zero-shot prompt is one of these 48 strings.}
\label{tab:appendix_prompt_catalog}\\
\toprule
Attribute & Value & Sentence \\
\midrule
\endfirsthead
\toprule
Attribute & Value & Sentence \\
\midrule
\endhead
Disease & healthy & This retinal image appears normal, with no signs of glaucoma. \\
 & glaucoma & This retinal image shows signs of glaucoma. \\
\midrule
Vertical CDR & very small & The vertical cup-to-disc ratio is very small. \\
 & small & The vertical cup-to-disc ratio is small. \\
 & moderate & The vertical cup-to-disc ratio is moderate. \\
 & large & The vertical cup-to-disc ratio is large. \\
 & very large & The vertical cup-to-disc ratio is very large. \\
\midrule
Horizontal CDR & very small & The horizontal cup-to-disc ratio is very small. \\
 & small & The horizontal cup-to-disc ratio is small. \\
 & moderate & The horizontal cup-to-disc ratio is moderate. \\
 & large & The horizontal cup-to-disc ratio is large. \\
 & very large & The horizontal cup-to-disc ratio is very large. \\
\midrule
Area CDR & low & The overall area cup-to-disc ratio is low. \\
 & moderate & The overall area cup-to-disc ratio is moderate. \\
 & high & The overall area cup-to-disc ratio is high. \\
 & very high & The overall area cup-to-disc ratio is very high. \\
\midrule
Rim area fraction & preserved & The neuroretinal rim area is preserved. \\
 & mild loss & The neuroretinal rim shows mild loss. \\
 & moderate loss & The neuroretinal rim shows moderate loss. \\
 & severe loss & The neuroretinal rim shows severe loss. \\
\midrule
Inferior rim thickness & very thin & The inferior rim is very thin. \\
 & thin & The inferior rim is thin. \\
 & moderate & The inferior rim is moderate in thickness. \\
 & thick & The inferior rim is thick. \\
 & very thick & The inferior rim is very thick. \\
\midrule
Superior rim thickness & very thin & The superior rim is very thin. \\
 & thin & The superior rim is thin. \\
 & moderate & The superior rim is moderate in thickness. \\
 & thick & The superior rim is thick. \\
 & very thick & The superior rim is very thick. \\
\midrule
Rim vertical asymmetry & none & There is no asymmetry between the superior and inferior rim. \\
 & mild & There is mild asymmetry between the superior and inferior rim. \\
 & moderate & There is moderate asymmetry between the superior and inferior rim. \\
 & marked & There is marked asymmetry between the superior and inferior rim. \\
\midrule
Disc V/H ratio & horizontal elongation & The optic disc is horizontally elongated. \\
 & round & The optic disc is round in shape. \\
 & vertical elongation & The optic disc is vertically elongated. \\
\midrule
Cup V/H ratio & horizontal elongation & The optic cup is horizontally elongated. \\
 & round & The optic cup is round in shape. \\
 & vertical elongation & The optic cup is vertically elongated. \\
\midrule
Disc eccentricity & round & The optic disc is round. \\
 & slightly elliptic & The optic disc is slightly elliptical. \\
 & elliptic & The optic disc is elliptical. \\
 & highly elliptic & The optic disc is highly elliptical. \\
\midrule
Cup eccentricity & round & The optic cup is round. \\
 & slightly elliptic & The optic cup is slightly elliptical. \\
 & elliptic & The optic cup is elliptical. \\
 & highly elliptic & The optic cup is highly elliptical. \\
\botrule
\end{longtable}

\subsection*{Training data}
\label{sec:method-data}

We trained on six public datasets, namely AIROGS \citep{devente2024airogs}, Chaksu \citep{kumar2023chaksu}, ORIGA \citep{zhang2010origa}, PAPILA \citep{kovalyk2022papila}, REFUGE \citep{orlando2020refuge} and REFUGE2 \citep{fang2022refuge2}, which gave 9,778 images, of which 3,653 were glaucoma, 5,919 were healthy and 206 were unlabeled, as listed in Table~\ref{tab:dataset_composition_sources}. The disease label of each image is the one published with its dataset. The 206 unlabeled images are glaucoma suspects from Chaksu and PAPILA without a confirmed label, so they carry biomarker sentences and no disease sentence. Five of the datasets have disc and cup masks, which gave 3,238 images with biomarker sentences. Of these, 383 were glaucoma, 2,649 were healthy and 206 were unlabeled. AIROGS has no masks, so its images carry only the disease sentence. Together the images yielded 45,190 image-sentence pairs.

From the 101,442 images of AIROGS we drew at random 3,270 glaucoma and 3,270 healthy images. We divided the five datasets with masks by patient into five folds, so that no patient appears in two folds. The first fold, with 845 images from 784 patients, is the internal validation split, and the other four folds, with 3,238 images, together with the AIROGS images form the training pool. Every GlaucomaMarker result in this paper uses this one partition.

\begin{table}[htbp]
\centering
\small
\caption{Composition of the training pool, internal held-out validation split, and external evaluation cohorts by source dataset.}
\label{tab:dataset_composition_sources}
\resizebox{\ifdim\width>\linewidth\linewidth\else\width\fi}{!}{%
\begin{tabular}{lcccc}
\toprule
Source & n & Glaucoma & Healthy & Unlabeled \\
\midrule
\quad AIROGS & 6540 & 3270 & 3270 & 0 \\
\quad Chaksu & 1065 & 0 & 913 & 152 \\
\quad ORIGA & 528 & 135 & 393 & 0 \\
\quad PAPILA & 380 & 58 & 268 & 54 \\
\quad REFUGE & 629 & 61 & 568 & 0 \\
\quad REFUGE2 & 636 & 129 & 507 & 0 \\
\textbf{\textit{Training pool total}} & \textbf{9778} & \textbf{3653} & \textbf{5919} & \textbf{206} \\
\midrule
\quad Chaksu & 280 & 0 & 244 & 36 \\
\quad ORIGA & 122 & 33 & 89 & 0 \\
\quad PAPILA & 108 & 29 & 65 & 14 \\
\quad REFUGE & 171 & 19 & 152 & 0 \\
\quad REFUGE2 & 164 & 31 & 133 & 0 \\
\textbf{\textit{Internal held-out validation total}} & \textbf{845} & \textbf{112} & \textbf{683} & \textbf{50} \\
\midrule
\quad Drishti-GS & 101 & 70 & 31 & 0 \\
\quad WF\_Glaucoma & 448 & 139 & 235 & 74 \\
\textbf{\textit{External evaluation total}} & \textbf{549} & \textbf{209} & \textbf{266} & \textbf{74} \\
\botrule
\end{tabular}}
\end{table}

\subsection*{Model and objective}
\label{sec:method-pretraining}

We started from retinal models, so that the training text only has to teach the reading of biomarkers. Both instances of the framework are dual encoders whose outputs are projected into a 512-dimensional space and L2-normalized. GlaucomaMarker-R starts from the RETFound ViT-L/16 \citep{zhou2023retfound}, which was pretrained on unlabeled retinal images with masked autoencoding, and pairs it with BioBERT \citep{lee2020biobert} and two-layer MLP projection heads. GlaucomaMarker-F starts from the released FLAIR \citep{silva2025flair}, a ResNet-50 image encoder and a clinical BERT text encoder with the linear projection heads of FLAIR. It keeps the image input and text pooling of FLAIR, except that the text embedding averages over real tokens only. Table~\ref{tab:instances} lists the two instances. Everything below applies to both. This gives image embeddings $u_i = \mathrm{proj}_v(f_v(x_i))$ and sentence embeddings $v_{i,k} = \mathrm{proj}_t(f_t(T_{i,k}))$ for each sentence $T_{i,k}$ of image $i$.

\begin{table}[htbp]
\centering
\caption{The two instances of GlaucomaMarker. Selected epochs are for seeds 42, 43 and 44.}
\label{tab:instances}
\begin{tabular}{lll}
\toprule
 & GlaucomaMarker-R & GlaucomaMarker-F \\
\midrule
Starting model & RETFound & FLAIR \\
Image encoder & ViT-L/16 & ResNet-50 \\
Text encoder & BioBERT & Bio\_ClinicalBERT \\
Projection heads & Two-layer MLP & FLAIR linear heads \\
Selected epochs & 20, 17, 14 & 19, 19, 17 \\
\bottomrule
\end{tabular}
\end{table}

Several sentences in a batch can share the same biomarker and bucket, so a loss with a single positive per image would treat true positives as negatives. We therefore used a multi-positive loss. Let $B$ be the number of images in a batch and $\tau = 0.07$ the softmax temperature, and write $s_{ij} = u_i^\top v_j / \tau$ for the scaled cosine similarity of image $i$ and sentence $j$. Each sentence belongs to a group, which is a biomarker and bucket or a disease label. For image $i$, the positives $\mathcal{P}(i)$ are the sentences in the batch that share a group with any sentence of that image, and the negatives $\mathcal{N}(i)$ are all other sentences. We used the multi-positive InfoNCE form of UniCLIP, which we call MP-NCE \citep{lee2022uniclip}. In the image-to-text direction it is
\begin{equation}
\mathcal{L}_{\text{i2t}} = \frac{1}{B} \sum_{i=1}^{B} \frac{1}{|\mathcal{P}(i)|} \sum_{p \in \mathcal{P}(i)} \left[ -s_{ip} + \log\Big( e^{s_{ip}} + \sum_{n \in \mathcal{N}(i)} e^{s_{in}} \Big) \right],
\end{equation}
and $\mathcal{L}_{\text{t2i}}$ applies the same rule from each sentence to the images, with $\mathcal{L}_{\text{MP-NCE}} = \tfrac{1}{2}(\mathcal{L}_{\text{i2t}} + \mathcal{L}_{\text{t2i}})$. Each positive has its own denominator, which sums over negatives only. This stops easy positives from suppressing the gradient of hard ones, as can happen with the shared denominator of the SupCon form \citep{khosla2020supcon}.

Public glaucoma datasets are dominated by healthy images, so we added an auxiliary cosine loss that pulls each image toward the sentences it carries, whatever the composition of the batch. It is inspired by the fixed class targets of targeted supervised contrastive learning \citep{li2022tsc}, with the sentence embeddings as the targets. For an image $i$ with sentence set $\mathcal{S}_i$,
\begin{equation}
\mathcal{L}_{\text{cos}} = \frac{1}{B} \sum_{i=1}^{B} \frac{1}{|\mathcal{S}_i|} \sum_{k \in \mathcal{S}_i} \left( 1 - u_i^\top\, \mathrm{sg}(v_{i,k}) \right),
\end{equation}
where $\mathrm{sg}$ stops the gradient, so the targets come from the current text encoder at every step and no gradient reaches the text encoder through this term. This differs from the original method, which uses precomputed targets spread uniformly on the hypersphere. The training objective was
\begin{equation}
\mathcal{L} = \mathcal{L}_{\text{MP-NCE}} + \lambda\, \mathcal{L}_{\text{cos}},
\end{equation}
with a fixed weight $\lambda = 0.5$. We compared this loss with four other long-tail rebalancing methods, BCL \citep{zhu2022bcl}, PaCo \citep{cui2021paco}, RCL \citep{dealvis2025rcl} and ReCL \citep{chang2025recl}, in a single-seed comparison given in Supplementary Table~\ref{tab:ablation_rebalancing_technique}.

\subsection*{Equal-bucket epoch sampler}
\label{sec:method-sampler}

\begin{figure}[htbp]
\centering
\includegraphics[width=\textwidth]{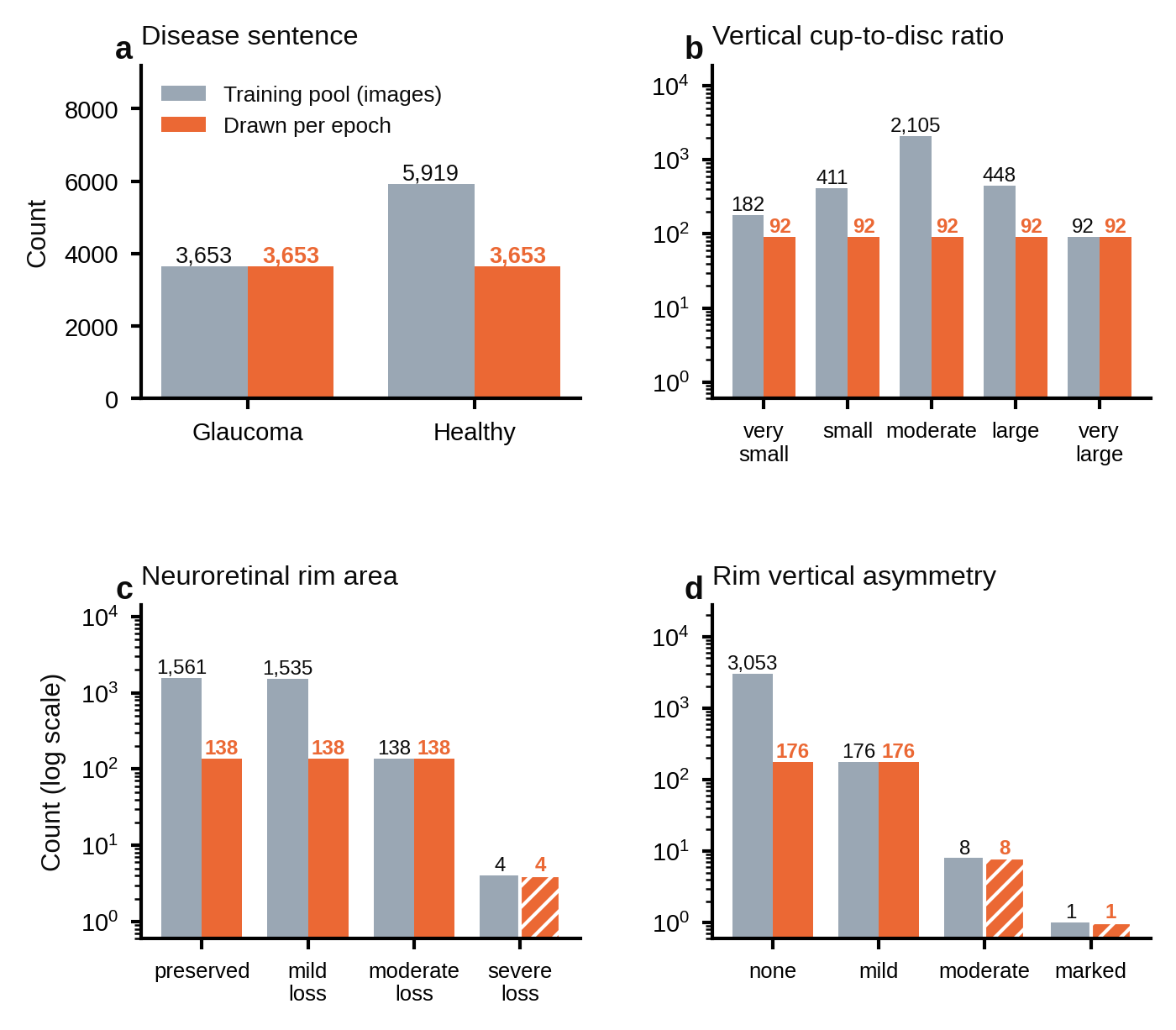}
\caption{\textbf{Effect of the equal-bucket epoch sampler on the training pool.} Gray bars are the natural number of images and orange bars are the sentences drawn in one epoch, as the mean of five epochs. \textbf{a} The disease sentence. \textbf{b} to \textbf{d} Three biomarkers, the vertical cup-to-disc ratio, the neuroretinal rim area and the rim vertical asymmetry, on a logarithmic scale. Hatched bars mark buckets with fewer than 20 images, which keep their natural count. Biomarker sentences exist for the 3,238 images with masks.}
\label{fig:sampler_demo}
\end{figure}

The training images are skewed toward healthy eyes and, for most biomarkers, toward a middle bucket, as Supplementary Figure~\ref{fig:bucket_skew} shows. About two thirds of the images, 2,105 of 3,238, fell in the middle cup-to-disc bucket, so a model trained on natural counts rarely sees the other findings. We therefore built each epoch with an equal-bucket sampler that balanced both the disease classes and the biomarker buckets. For the disease sentence, every epoch included all glaucoma images and an equal number of healthy images, freshly drawn at random, so the two sentence counts were exactly equal. For each biomarker, every well-supported bucket, meaning one backed by at least 20 distinct images, contributed the same number of sentences per epoch, equal to the size of the smallest well-supported bucket of that biomarker, which was 92 sentences per bucket for the vertical cup-to-disc ratio. The images for each bucket were drawn at random and re-drawn every epoch, and no image was repeated within an epoch. Buckets backed by fewer than 20 images kept their natural count. Each drawn image carried only the sentences it was selected for, so one epoch held about 8,200 image draws. We balanced exactly within every epoch because we selected the final checkpoint at a single epoch. Figure~\ref{fig:sampler_demo} shows the resulting counts for three biomarkers.

The auxiliary cosine loss and the epoch sampler both address class imbalance, but at different points in training. The sampler controls what the model sees, and the loss controls how the embedding space is shaped once it has seen it. Results, Ablation studies tests these choices with three seeds per configuration.

\subsection*{Evaluation cohorts}
\label{sec:method-cohorts}

We evaluated on two cohorts that played no part in training, threshold selection or checkpoint selection. Drishti-GS \citep{sivaswamy2014drishti} is a public benchmark of 101 images, 70 glaucoma and 31 healthy, with expert disc and cup segmentation. Because GlaucomaMarker-F starts from FLAIR, whose pretraining data overlap with this cohort, we did not evaluate it on Drishti-GS. WF\_Glaucoma is a clinical cohort from Wake Forest University School of Medicine, collected independently of every dataset in the training pool. It has 448 color fundus photographs of 225 patients, taken with a Topcon Maestro at three sites, and we scored each eye as one image. Of these, 374 carried a binary glaucoma label, 139 glaucoma and 235 healthy, and were used for classification, and the 74 without a label were not used for classification. The ground truth of the disease is the clinician diagnosis recorded for each cohort. The ground truth of a biomarker is a measurement that we turned into a bucket with the same cutoffs as the training text. On Drishti-GS the measurements come from the expert masks, processed by the same pipeline as the training text. On WF\_Glaucoma they come from the OCT scan measurements made by clinicians, which cover five of the eleven biomarkers and are available for 444 of the images.

The study used retrospective, de-identified images and was conducted in accordance with the Declaration of Helsinki. The use of WF\_Glaucoma was approved by the Wake Forest University School of Medicine institutional review board, and informed consent was waived because the data were de-identified. The public datasets were released with their own ethical approvals, and we used only their published images and labels.

\subsection*{Comparison models}
\label{sec:method-baselines}

We compared both instances with seven baselines, all used as publicly released checkpoints. Five are vision-language models. FLAIR \citep{silva2025flair} and EyeCLIP \citep{shi2025eyeclip} are retinal models pretrained on category labels or on template and caption text, UniMed-CLIP \citep{khattak2024unimedclip} is a general medical model, FairCLIP \citep{luo2024fairclip} is a fairness-aware fundus model, and KeepFIT \citep{wu2024mmretinal} is a retinal image-text model with a CNN backbone pretrained on MM-Retinal. Each was evaluated zero-shot with its own text encoder, and read the biomarkers with the same bucket sentences as our models. The other two baselines are supervised glaucoma classifiers, the glaucoma checkpoints released with RETFound \citep{zhou2023retfound} and with RETFound-DE \citep{sun2025retfoundde}, a data-efficient model built on RETFound. We used them as released, and their disease score was the probability of any glaucoma class. These two have no text encoder, so they have no biomarker readout. FLAIR and UniMed-CLIP list Drishti-GS in their published pretraining data, KeepFIT pretrains on a compilation of public datasets that comes from FLAIR, and GlaucomaMarker-F starts from FLAIR. We therefore left Drishti-GS blank for these four models and report them on WF\_Glaucoma only. Every baseline is a single run, and each was thresholded on its own validation output where available, as described under Statistical analysis.

\subsection*{Zero-shot readout and evaluation tasks}
\label{sec:method-inference}

At test time we encoded each of the 48 sentences once with the frozen text encoder and cached the embeddings. Every output of the model is therefore one of these 48 sentences. Every zero-shot result in this paper used these cached embeddings and the frozen image embedding $u = \mathrm{proj}_v(f_v(x))$. Let $v_{g}$ and $v_{h}$ be the embeddings of the glaucomatous and the healthy disease sentence. The probability of glaucoma was the softmax of the two cosine similarities,
\begin{equation}
p(\text{glaucoma} \mid x) = \frac{e^{u^\top v_{g}}}{e^{u^\top v_{g}} + e^{u^\top v_{h}}},
\end{equation}
and an image was called glaucoma when $p \geq t^{*}$. We chose the threshold $t^{*}$ on the internal validation split of the model, which held 845 images from the same five public datasets and was held out from training. We maximized the Matthews correlation coefficient there and froze the threshold before evaluating either external cohort. For a biomarker $b$ with bucket sentences $\{v_{b,c}\}$, the predicted bucket was
\begin{equation}
\hat{c}_b(x) = \arg\max_{c}\; u^\top v_{b,c},
\end{equation}
taken over the sentences of that biomarker. For retrieval, an image was scored against all 48 sentences, and each sentence was scored against all images of a cohort. Image-to-text Recall@1 counted an image as correct when its highest-scoring sentence matched one of its true labels. For text-to-image retrieval, each sentence with at least five true images in a cohort ranked all images of that cohort. We report the fraction of top-ranked images that matched, Recall@1, and the mean fraction of the five top-ranked images that matched, P@5.

We also evaluated the frozen image features on their own. For the nearest-neighbor measure, we found the 10 nearest neighbors of every WF\_Glaucoma image by cosine similarity and report the mean share of them that carry the label of the image. For the linear probe, we fitted a logistic regression, with its regularization chosen by cross-validation, on the frozen pre-projection image features of the internal training split and evaluated it on each cohort. The label-efficiency curves refit the probe on 10 to 100 percent of the training split, and the few-shot curves refit it on 1, 2, 4, 8 or 16 labeled images per class, with five resamples for each setting. For the demographic analysis, we computed the AUROC within each subgroup of gender, race and age on WF\_Glaucoma, and the equity-scaled AUROC of Luo et al.~\cite{luo2024harvardgf},
\begin{equation}
\mathrm{ES\text{-}AUROC} = \frac{\mathrm{AUROC}}{1 + \sum_{g} \lvert \mathrm{AUROC} - \mathrm{AUROC}_{g} \rvert},
\end{equation}
where the sum runs over the subgroups of one attribute.

\subsection*{Ablation design}
\label{sec:method-ablation}

Each ablation changed one factor of the final pipeline and kept the rest fixed. Every configuration was trained for 20 epochs on the same data with seeds 42, 43 and 44, selected at the epoch with the highest sum of validation disease AUROC and mean biomarker balanced accuracy, and evaluated once on each cohort with the threshold frozen on the internal validation split. The factors were the loss, the sampler, the training text, the starting weights of the image encoder and, for the FLAIR start, the sampler and the text. The loss was SupCon \citep{khosla2020supcon}, MP-NCE, or MP-NCE with the auxiliary cosine loss, each trained with plain shuffling. The sampler was plain shuffling, the disease-balanced sampler, which balances only the disease sentence, or the equal-bucket sampler, each with MP-NCE and the auxiliary cosine loss. The training text was the full text, the disease sentence alone, or the full text with the biomarker sentences shuffled across images, which keeps their number and form and removes their link to the image. The starting weights were RETFound, ImageNet ViT-L/16 with the same architecture, or ImageNet ResNet-50 with a different architecture, each under the final pipeline. For the FLAIR start, the configurations were plain shuffling with and without the auxiliary cosine loss, the disease-balanced sampler with the loss, the equal-bucket sampler, and the disease sentence alone, and they were evaluated on WF\_Glaucoma only, because the pretraining of FLAIR overlaps with Drishti-GS. A single-seed comparison of the auxiliary cosine loss with four other rebalancing methods is given in Supplementary Table~\ref{tab:ablation_rebalancing_technique}.

We call a difference between two configurations clear when the seed ranges of the two do not overlap, which describes the three seeds and is not a test. To measure the preference of the biomarker reading for common buckets, we used the difference between accuracy and balanced accuracy, averaged over the biomarkers,
\begin{equation}
\Delta = \frac{1}{|\mathcal{B}|} \sum_{b \in \mathcal{B}} \left( \mathrm{Acc}_b - \mathrm{BalAcc}_b \right),
\end{equation}
where $\mathcal{B}$ is the set of biomarkers with ground truth in a cohort, $\mathrm{Acc}_b$ is the share of images whose bucket was predicted correctly and $\mathrm{BalAcc}_b$ is the mean recall over the buckets present. The difference $\Delta$ is positive when the reading favors common buckets, zero when all buckets are read equally well, and negative when rare buckets are read slightly better.

\subsection*{Statistical analysis}
\label{sec:method-stats}

We report AUROC and AUPRC as threshold-free ranking metrics, and accuracy, balanced accuracy, sensitivity, specificity, F1, MCC and Cohen's kappa at a single operating threshold. Following KAD \citep{zhang2023kad}, the threshold was chosen on a validation set that is separate from the reported test set. For each seed we chose it once on the internal validation split and froze it before evaluating either cohort, and each baseline was thresholded the same way on its own validation output where available. For a biomarker with $K$ buckets present in a cohort, we report balanced accuracy, the mean recall over those buckets, because plain accuracy rewards always predicting the most common bucket. A model that chooses a bucket without using the image is correct for one bucket in every $K$ on average, so we call $1/K$ the random guessing level, which averaged over the biomarkers is 0.257 on WF\_Glaucoma and 0.286 on Drishti-GS.

Every GlaucomaMarker result is the mean over three seeds, while baselines are single runs. We computed 95 percent confidence intervals as the 2.5th and 97.5th percentiles of 10,000 bootstrap resamples with a fixed random seed. Resamples were drawn by patient on WF\_Glaucoma, where the 374 labeled eyes come from 198 patients, and by image on Drishti-GS, which has no patient information. Within every resample the metrics of GlaucomaMarker were averaged over its three seeds. A difference between two models is reported with the interval of the difference. Where a test is stated, its two-sided $P$ value is $2\min\big(\tfrac{r_{\le 0}+1}{10{,}001}, \tfrac{r_{\ge 0}+1}{10{,}001}\big)$, with $r_{\le 0}$ and $r_{\ge 0}$ the numbers of resamples in which the difference was at most zero and at least zero. Two families of tests were defined before the analysis, each adjusted with the Holm method at a significance level of 0.05. The first has four tests, GlaucomaMarker-F and GlaucomaMarker-R against FLAIR on WF\_Glaucoma, for AUROC and MCC. The second has six tests on the biomarker reading, three against the random guessing level and three paired against the best baseline. The three seeds are described by their range. The comparisons that were reported with an interval and without a test are listed in Supplementary Table~\ref{tab:statistical_comparisons}.

\subsection*{Implementation details}
\label{sec:method-implementation}

During training, images were randomly cropped and resized to $512 \times 512$ with a crop scale of 0.8 to 1.0, flipped horizontally with probability 0.5, and jittered in brightness, contrast and saturation with strength 0.1. At evaluation, images were resized to $512 \times 512$. All images were normalized with the ImageNet mean and standard deviation, which the FLAIR-based encoder undoes internally to give the original input of FLAIR. Text was tokenized and padded to 64 tokens. Both encoders and their projection heads were trained jointly for 20 epochs with AdamW \citep{loshchilov2019adamw}, a learning rate of $10^{-5}$ for all parameters, weight decay 0.05, a linear warmup over the first 10 percent of steps followed by cosine decay, batch size 16, bfloat16 mixed precision and a softmax temperature of 0.07. We trained three independent seeds, 42, 43 and 44, for each instance and report the mean $\pm$ standard deviation over seeds for every GlaucomaMarker result. For each seed, model selection used the epoch with the highest sum of validation disease AUROC and mean biomarker balanced accuracy on the internal validation split, with the epochs listed in Table~\ref{tab:instances}. Each selected checkpoint was then evaluated once, frozen, on each external cohort. We trained each model on a single NVIDIA RTX 6000 Ada GPU with 48 GB of memory, using Python 3.12.7 and PyTorch 2.7.0.

\section*{Data availability}

We will release the catalog of 48 sentences and the full training dataset with its 45,190 image-text pairs. WF\_Glaucoma is a private clinical cohort and will not be released.

\section*{Code availability}

We will release the trained GlaucomaMarker-R and GlaucomaMarker-F models and the code that computes the biomarker sentences from segmentation masks.

\bibliography{references}

\section*{Acknowledgements}


\section*{Author contributions}

A.S.N. conceived the framework, developed the code, trained the models and ran the experiments. S.S. contributed to the analysis. N.C. and A.C.T. collected and curated the WF\_Glaucoma clinical data. J.X., T.L. and J.I.L. provided clinical expertise. M.N.A. supervised the study. A.S.N. wrote the manuscript, and all authors reviewed and approved it.

\section*{Competing interests}

The authors declare no competing interests.

\section*{Additional information}

\textbf{Supplementary information} accompanies this paper as a separate PDF.

\textbf{Correspondence} and requests for materials should be addressed to the corresponding author.

\clearpage
\setcounter{table}{0}
\setcounter{figure}{0}
\setcounter{page}{1}
\renewcommand{\thepage}{S\arabic{page}}
\def\theHpage{S\arabic{page}}
\def\theHtable{S\arabic{table}}
\def\theHfigure{S\arabic{figure}}
\renewcommand{\tablename}{Supplementary Table}
\renewcommand{\figurename}{Supplementary Figure}

\section*{Supplementary Information for GlaucomaMarker - Biomarker-Grounded Glaucoma Vision-Language Model}

\begin{figure}[t]
\centering
\includegraphics[width=\textwidth]{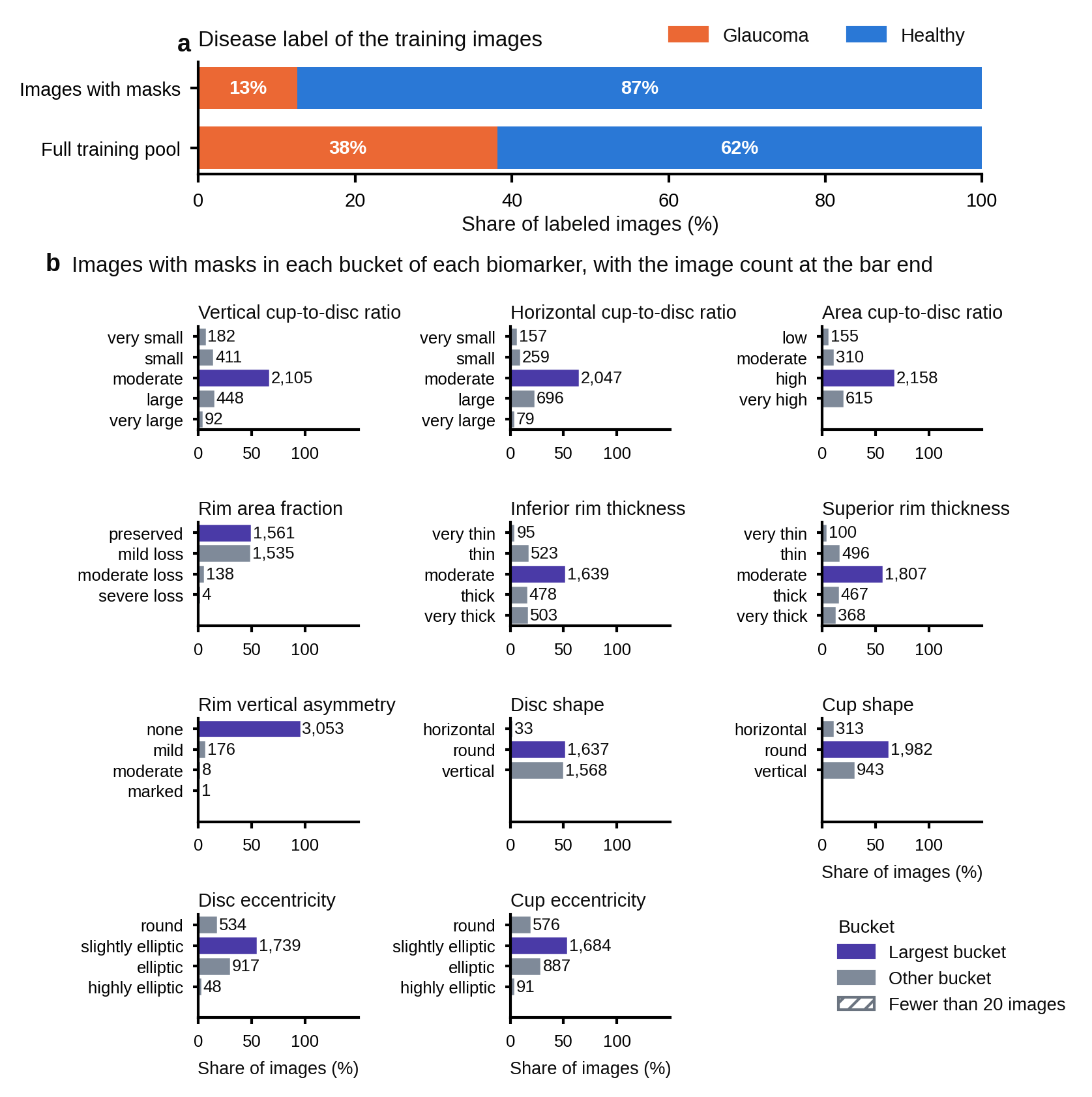}
\caption{\textbf{Training images by disease label and by biomarker bucket.} \textbf{a} Share of glaucoma and healthy images among the training images with masks and in the full training pool. \textbf{b} Number of images in each bucket of each biomarker, shown as the share of the images of that biomarker, with the image count at the end of each bar. The largest bucket is purple and hatched buckets have fewer than 20 images. The counts cover the 3,238 images with masks.}
\label{fig:bucket_skew}
\end{figure}

Supplementary Figure~\ref{fig:bucket_skew} shows the number of training images in each bucket of each biomarker and the share of glaucoma and healthy images. Supplementary Table~\ref{tab:linked_biomarker_definition} defines the disease-linked biomarkers used in the Results. Supplementary Table~\ref{tab:confusion_matrices} gives the confusion-matrix counts and the decision threshold of every model on both cohorts, including FLAIR, UniMed-CLIP, and KeepFIT on WF\_Glaucoma. Supplementary Table~\ref{tab:appendix_ablation_concepts} gives the zero-shot balanced accuracy of every ablation configuration for each biomarker concept, on Drishti-GS, with 11 biomarkers, and on the WF\_Glaucoma images with OCT ground truth, with 5 biomarkers. Supplementary Table~\ref{tab:statistical_comparisons} gives the comparisons that were reported with a 95\% interval and no test, for diagnosis, the embedding, and fairness. Supplementary Tables~\ref{tab:ablation_seed_ranges} and~\ref{tab:ablation_gap} give the seed values behind the ablation results, and Supplementary Table~\ref{tab:ablation_rebalancing_technique} gives the one-seed comparison of five rebalancing techniques.

\begin{table}[htbp]
\centering
\small
\caption{Association between each biomarker's true bucket and the true diagnosis, as Cram\'er's $V$, in the training split and in the three evaluation sets. A biomarker is called disease-linked when $V$ is at least 0.30 in the training split. Each $n$ counts the images with a glaucoma label. Blank cells have no ground truth.}
\label{tab:linked_biomarker_definition}
\resizebox{\ifdim\width>\linewidth\linewidth\else\width\fi}{!}{%
\begin{tabular}{lccccc}
\toprule
Biomarker & \#classes & Training split (n=3,032) & Internal val.\ (n=795) & Drishti-GS (n=101) & WF\_Glaucoma (n=370) \\
\midrule
\addlinespace[2pt]
\multicolumn{6}{l}{\scriptsize\textsc{Disease-linked}} \\
\quad Vertical CDR & 5 & 0.56 & 0.57 & 0.74 & 0.75 \\
\quad Rim thickness (sup.) & 5 & 0.50 & 0.49 & 0.37 &  \\
\quad Area CDR & 4 & 0.47 & 0.47 & 0.71 & 0.52 \\
\quad Rim thickness (inf.) & 5 & 0.46 & 0.47 & 0.38 &  \\
\quad Rim area fraction & 4 & 0.46 & 0.46 & 0.64 & 0.70 \\
\quad Horizontal CDR & 5 & 0.42 & 0.39 & 0.74 &  \\
\addlinespace[2pt]
\multicolumn{6}{l}{\scriptsize\textsc{Other}} \\
\quad Cup V/H ratio & 3 & 0.15 & 0.18 & 0.30 &  \\
\quad Cup eccentricity & 4 & 0.11 & 0.13 & 0.35 &  \\
\quad Disc eccentricity & 4 & 0.06 & 0.07 & 0.17 & 0.10 \\
\quad Rim vert. asymmetry & 4 & 0.05 & 0.04 & 0.27 &  \\
\quad Disc V/H ratio & 3 & 0.04 & 0.01 & 0.16 & 0.17 \\
\botrule
\end{tabular}}
\end{table}

\begin{table}[htbp]
\centering
\small
\caption{Confusion matrices at the validation-set MCC-optimal decision threshold, frozen before evaluation on either external cohort. FLAIR, UniMed-CLIP, and KeepFIT are blank or omitted on Drishti-GS because their pretraining data overlap with Drishti-GS, and the FLAIR-initialized model is blank there for the same reason. $^{\dagger}$GlaucomaMarker rows are the mean over three seeds (threshold and counts), so counts are not integers.}
\label{tab:confusion_matrices}
\resizebox{\ifdim\width>\linewidth\linewidth\else\width\fi}{!}{%
\begin{tabular}{lcccccccccc}
\toprule
& \multicolumn{5}{c}{Drishti-GS (n=101)} & \multicolumn{5}{c}{WF\_Glaucoma (n=374)} \\
\cmidrule(lr){2-6}\cmidrule(lr){7-11}
Model & Threshold & TP & FN & TN & FP & Threshold & TP & FN & TN & FP \\
\midrule
EyeCLIP & 0.5195 & 23 & 47 & 16 & 15 & 0.5195 & 122 & 17 & 46 & 189 \\
FairCLIP & 0.2365 & 6 & 64 & 30 & 1 & 0.2365 & 30 & 109 & 192 & 43 \\
RETFound & 0.7289 & 3 & 67 & 31 & 0 & 0.7289 & 13 & 126 & 219 & 16 \\
RETFound-DE & 0.8559 & 7 & 63 & 31 & 0 & 0.8559 & 0 & 139 & 233 & 2 \\
FLAIR &  &  &  &  &  & 0.5408 & 117 & 22 & 152 & 83 \\
UniMed-CLIP &  &  &  &  &  & 0.9997 & 6 & 133 & 234 & 1 \\
KeepFIT (MM-Retinal) &  &  &  &  &  & 0.5025 & 35 & 104 & 219 & 16 \\
\textbf{GlaucomaMarker-R}$^{\dagger}$ & 0.5204 & 50.0 & 20.0 & 26.7 & 4.3 & 0.5204 & 105.3 & 33.7 & 190.3 & 44.7 \\
\textbf{GlaucomaMarker-F}$^{\dagger}$ &  &  &  &  &  & 0.4918 & 99.7 & 39.3 & 216.0 & 19.0 \\
\botrule
\end{tabular}}
\end{table}

\begin{table}[htbp]
\centering
\scriptsize
\setlength{\tabcolsep}{3pt}
\caption{Ablation of the GlaucomaMarker framework, zero-shot balanced accuracy for each biomarker concept, on Drishti-GS (11 biomarkers) and on the WF\_Glaucoma images with OCT ground truth (5 biomarkers, n=444). Values are means over three seeds (standard deviations are omitted for width). Configurations, seeds, and checkpoint selection as in Table~\ref{tab:ablation_loss_ladder}; bold marks the best entry in each column.}
\label{tab:appendix_ablation_concepts}
\resizebox{\ifdim\width>\linewidth\linewidth\else\width\fi}{!}{%
\begin{tabular}{lcccccccccccccccc}
\toprule
& \multicolumn{11}{c}{Drishti-GS (n=101)} & \multicolumn{5}{c}{WF\_Glaucoma (n=444)} \\
\cmidrule(lr){2-12}\cmidrule(lr){13-17}
Configuration & V-CDR & H-CDR & Area CDR & Rim area & Rim inf. & Rim sup. & Rim asym. & Disc V/H & Cup V/H & Disc ecc. & Cup ecc. & V-CDR & Area CDR & Rim area & Disc V/H & Disc ecc. \\
\midrule
\addlinespace[2pt]
\multicolumn{17}{l}{\scriptsize\textsc{Loss function}} \\
\quad SupCon (Khosla et al. 2020) & 0.459 & 0.508 & 0.591 & 0.457 & 0.278 & 0.296 & 0.211 & 0.352 & 0.452 & 0.321 & 0.340 & 0.314 & 0.356 & 0.403 & 0.310 & 0.293 \\
\quad MP-NCE (Lee et al. 2022) & 0.441 & \textbf{0.598} & 0.686 & 0.483 & 0.287 & 0.353 & 0.239 & 0.393 & 0.434 & 0.317 & 0.268 & 0.317 & 0.337 & 0.392 & 0.333 & \textbf{0.311} \\
\quad MP-NCE + auxiliary cosine loss & \textbf{0.544} & 0.563 & \textbf{0.739} & 0.540 & 0.270 & 0.344 & 0.236 & 0.456 & 0.416 & 0.274 & 0.355 & 0.311 & 0.343 & 0.368 & \textbf{0.346} & 0.274 \\
\addlinespace[2pt]
\multicolumn{17}{l}{\scriptsize\textsc{Epoch sampler (MP-NCE + auxiliary cosine loss)}} \\
\quad Disease-balanced sampler & 0.447 & 0.528 & 0.587 & 0.396 & 0.269 & 0.312 & 0.248 & 0.494 & 0.449 & 0.138 & \textbf{0.366} & 0.306 & 0.345 & 0.354 & 0.321 & 0.258 \\
\quad \textbf{Equal-bucket sampler (GlaucomaMarker-R)} & 0.405 & 0.538 & 0.627 & 0.546 & \textbf{0.316} & \textbf{0.376} & 0.228 & 0.302 & \textbf{0.602} & 0.270 & 0.347 & 0.354 & 0.360 & 0.427 & 0.338 & 0.293 \\
\addlinespace[2pt]
\multicolumn{17}{l}{\scriptsize\textsc{Text content (disease-balanced sampler)}} \\
\quad Disease sentence only & 0.144 & 0.168 & 0.130 & \textbf{0.559} & 0.164 & 0.125 & 0.194 & 0.395 & 0.284 & 0.207 & 0.314 & 0.245 & 0.244 & 0.456 & 0.336 & 0.196 \\
\quad Biomarker sentences shuffled & 0.250 & 0.250 & 0.333 & 0.383 & 0.200 & 0.200 & 0.250 & \textbf{0.521} & 0.333 & \textbf{0.333} & 0.250 & 0.200 & 0.250 & 0.333 & 0.320 & 0.250 \\
\addlinespace[2pt]
\multicolumn{17}{l}{\scriptsize\textsc{Image-encoder initialization (final model)}} \\
\quad RETFound ViT-L/16 (final model) & 0.405 & 0.538 & 0.627 & 0.546 & \textbf{0.316} & \textbf{0.376} & 0.228 & 0.302 & \textbf{0.602} & 0.270 & 0.347 & 0.354 & 0.360 & 0.427 & 0.338 & 0.293 \\
\quad ImageNet ViT-L/16 (same architecture) & 0.200 & 0.290 & 0.305 & 0.485 & 0.281 & 0.324 & \textbf{0.254} & 0.229 & 0.569 & 0.202 & 0.252 & 0.272 & 0.316 & 0.303 & 0.296 & 0.246 \\
\quad ImageNet ResNet-50 (different architecture) & 0.284 & 0.262 & 0.272 & 0.433 & 0.268 & 0.362 & 0.196 & 0.320 & 0.451 & 0.213 & 0.321 & 0.348 & 0.351 & 0.380 & 0.318 & 0.242 \\
\quad FLAIR ResNet-50 (fundus-pretrained, different architecture) &  &  &  &  &  &  &  &  &  &  &  & \textbf{0.384} & \textbf{0.385} & \textbf{0.466} & 0.196 & 0.232 \\
\botrule
\end{tabular}}
\end{table}

\begin{table}[htbp]
\centering
\footnotesize
\caption{Comparisons reported with a 95\% bootstrap interval of the difference and no test. The intervals come from 10,000 resamples, drawn by patient on WF\_Glaucoma and by image on Drishti-GS, with the three seeds of GlaucomaMarker averaged inside each resample. ES-AUC is the equity-scaled AUROC.}
\label{tab:statistical_comparisons}
\resizebox{\ifdim\width>\linewidth\linewidth\else\width\fi}{!}{%
\begin{tabular}{lllll}
\toprule
Cohort & Comparison & Metric & Difference & 95\% CI \\
\midrule
\addlinespace
\multicolumn{5}{l}{\textit{Diagnosis, not part of the tested families}} \\
WF\_Glaucoma & GlaucomaMarker-F minus GlaucomaMarker-R & AUROC & 0.047 & 0.008 to 0.088 \\
WF\_Glaucoma & GlaucomaMarker-F minus GlaucomaMarker-R & MCC & 0.099 & 0.012 to 0.186 \\
Drishti-GS & GlaucomaMarker-R minus RETFound-DE & AUROC & 0.080 & $-$0.005 to 0.171 \\
Drishti-GS & GlaucomaMarker-R minus RETFound-DE & MCC & 0.354 & 0.203 to 0.503 \\
\addlinespace
\multicolumn{5}{l}{\textit{Embedding}} \\
WF\_Glaucoma & GlaucomaMarker-R minus RETFound, nearest-neighbor label agreement & Agreement & 0.160 & 0.127 to 0.194 \\
WF\_Glaucoma & GlaucomaMarker-F minus FLAIR, nearest-neighbor label agreement & Agreement & 0.019 & 0.002 to 0.037 \\
WF\_Glaucoma & GlaucomaMarker-F minus FLAIR, linear probe & AUROC & 0.079 & 0.033 to 0.128 \\
WF\_Glaucoma & GlaucomaMarker-R minus FLAIR, linear probe & AUROC & 0.031 & $-$0.027 to 0.089 \\
Drishti-GS & GlaucomaMarker-R minus RETFound-DE, linear probe & AUROC & $-$0.075 & $-$0.129 to $-$0.025 \\
Drishti-GS & GlaucomaMarker-R minus FairCLIP, linear probe & AUROC & $-$0.014 & $-$0.100 to 0.074 \\
\addlinespace
\multicolumn{5}{l}{\textit{Fairness, equity-scaled AUROC difference from FLAIR on WF\_Glaucoma}} \\
WF\_Glaucoma & GlaucomaMarker-F minus FLAIR, by gender & ES-AUC & 0.047 & $-$0.004 to 0.119 \\
WF\_Glaucoma & GlaucomaMarker-F minus FLAIR, by race & ES-AUC & 0.063 & $-$0.013 to 0.151 \\
WF\_Glaucoma & GlaucomaMarker-F minus FLAIR, by age band & ES-AUC & 0.095 & $-$0.061 to 0.211 \\
WF\_Glaucoma & GlaucomaMarker-R minus FLAIR, by gender & ES-AUC & 0.019 & $-$0.066 to 0.100 \\
WF\_Glaucoma & GlaucomaMarker-R minus FLAIR, by race & ES-AUC & 0.013 & $-$0.085 to 0.103 \\
WF\_Glaucoma & GlaucomaMarker-R minus FLAIR, by age band & ES-AUC & 0.028 & $-$0.152 to 0.178 \\
\botrule
\end{tabular}}
\end{table}

\begin{table}[htbp]
\centering
\footnotesize
\caption{Seed values of the ablation configurations, as the mean over the three seeds with the lowest and highest seed in parentheses. AUROC and MCC are zero-shot disease classification at the validation-frozen threshold, and Bio is the mean zero-shot balanced accuracy over the biomarkers with ground truth. FLAIR-start configurations are shown on WF\_Glaucoma only. Two seed ranges that do not overlap are the descriptive check for a difference between configurations.}
\label{tab:ablation_seed_ranges}
\resizebox{\ifdim\width>\linewidth\linewidth\else\width\fi}{!}{%
\begin{tabular}{lcccccc}
\toprule
& \multicolumn{3}{c}{Drishti-GS} & \multicolumn{3}{c}{WF\_Glaucoma} \\
\cmidrule(lr){2-4}\cmidrule(lr){5-7}
Configuration & AUROC & MCC & Bio & AUROC & MCC & Bio \\
\midrule
\addlinespace[2pt]
\multicolumn{7}{l}{\scriptsize\textsc{Loss function}} \\
\quad SupCon & 0.898{\scriptsize\ (0.880; 0.910)} & 0.559{\scriptsize\ (0.512; 0.652)} & 0.388{\scriptsize\ (0.345; 0.432)} & 0.830{\scriptsize\ (0.811; 0.857)} & 0.462{\scriptsize\ (0.415; 0.529)} & 0.335{\scriptsize\ (0.328; 0.345)} \\
\quad MP-NCE & 0.861{\scriptsize\ (0.849; 0.873)} & 0.513{\scriptsize\ (0.428; 0.555)} & 0.409{\scriptsize\ (0.384; 0.426)} & 0.822{\scriptsize\ (0.775; 0.861)} & 0.492{\scriptsize\ (0.450; 0.562)} & 0.338{\scriptsize\ (0.333; 0.345)} \\
\quad MP-NCE + auxiliary cosine loss & 0.875{\scriptsize\ (0.853; 0.887)} & 0.564{\scriptsize\ (0.547; 0.575)} & 0.430{\scriptsize\ (0.409; 0.445)} & 0.835{\scriptsize\ (0.808; 0.851)} & 0.513{\scriptsize\ (0.461; 0.543)} & 0.328{\scriptsize\ (0.296; 0.356)} \\
\addlinespace[2pt]
\multicolumn{7}{l}{\scriptsize\textsc{RETFound start, epoch sampler (with the auxiliary cosine loss)}} \\
\quad Disease-balanced sampler & 0.872{\scriptsize\ (0.856; 0.881)} & 0.525{\scriptsize\ (0.492; 0.584)} & 0.385{\scriptsize\ (0.357; 0.403)} & 0.839{\scriptsize\ (0.832; 0.844)} & 0.508{\scriptsize\ (0.461; 0.573)} & 0.317{\scriptsize\ (0.308; 0.324)} \\
\quad Equal-bucket sampler (GlaucomaMarker-R) & 0.853{\scriptsize\ (0.840; 0.865)} & 0.536{\scriptsize\ (0.449; 0.646)} & 0.414{\scriptsize\ (0.392; 0.439)} & 0.858{\scriptsize\ (0.848; 0.873)} & 0.563{\scriptsize\ (0.516; 0.611)} & 0.354{\scriptsize\ (0.335; 0.375)} \\
\addlinespace[2pt]
\multicolumn{7}{l}{\scriptsize\textsc{RETFound start, text content (disease-balanced sampler)}} \\
\quad Disease sentence only & 0.878{\scriptsize\ (0.865; 0.888)} & 0.530{\scriptsize\ (0.492; 0.584)} & 0.244{\scriptsize\ (0.180; 0.348)} & 0.851{\scriptsize\ (0.829; 0.865)} & 0.551{\scriptsize\ (0.519; 0.578)} & 0.295{\scriptsize\ (0.279; 0.304)} \\
\quad Biomarker sentences shuffled & 0.879{\scriptsize\ (0.865; 0.888)} & 0.491{\scriptsize\ (0.407; 0.599)} & 0.300{\scriptsize\ (0.286; 0.311)} & 0.839{\scriptsize\ (0.825; 0.854)} & 0.503{\scriptsize\ (0.462; 0.542)} & 0.271{\scriptsize\ (0.257; 0.278)} \\
\addlinespace[2pt]
\multicolumn{7}{l}{\scriptsize\textsc{Image-encoder initialization}} \\
\quad ImageNet ViT-L/16 & 0.833{\scriptsize\ (0.807; 0.854)} & 0.516{\scriptsize\ (0.449; 0.558)} & 0.308{\scriptsize\ (0.271; 0.357)} & 0.873{\scriptsize\ (0.863; 0.881)} & 0.582{\scriptsize\ (0.575; 0.586)} & 0.287{\scriptsize\ (0.276; 0.305)} \\
\quad ImageNet ResNet-50 & 0.860{\scriptsize\ (0.847; 0.867)} & 0.513{\scriptsize\ (0.474; 0.555)} & 0.307{\scriptsize\ (0.294; 0.325)} & 0.878{\scriptsize\ (0.863; 0.904)} & 0.593{\scriptsize\ (0.573; 0.616)} & 0.328{\scriptsize\ (0.317; 0.340)} \\
\addlinespace[2pt]
\multicolumn{7}{l}{\scriptsize\textsc{FLAIR start (WF\_Glaucoma only)}} \\
\quad Plain shuffling, no auxiliary cosine loss &  &  &  & 0.895{\scriptsize\ (0.892; 0.901)} & 0.680{\scriptsize\ (0.666; 0.698)} & 0.352{\scriptsize\ (0.349; 0.354)} \\
\quad Plain shuffling + auxiliary cosine loss &  &  &  & 0.907{\scriptsize\ (0.898; 0.913)} & 0.673{\scriptsize\ (0.663; 0.680)} & 0.339{\scriptsize\ (0.318; 0.355)} \\
\quad Disease-balanced sampler + auxiliary cosine loss &  &  &  & 0.912{\scriptsize\ (0.902; 0.919)} & 0.650{\scriptsize\ (0.641; 0.656)} & 0.356{\scriptsize\ (0.350; 0.367)} \\
\quad Equal-bucket sampler (GlaucomaMarker-F) &  &  &  & 0.905{\scriptsize\ (0.904; 0.905)} & 0.662{\scriptsize\ (0.642; 0.688)} & 0.332{\scriptsize\ (0.317; 0.344)} \\
\quad Disease sentence only (disease-balanced sampler) &  &  &  & 0.893{\scriptsize\ (0.890; 0.896)} & 0.666{\scriptsize\ (0.643; 0.687)} & 0.344{\scriptsize\ (0.336; 0.354)} \\
\botrule
\end{tabular}}
\end{table}

\begin{table}[htbp]
\centering
\footnotesize
\caption{Preference for common buckets in the biomarker reading, measured as accuracy minus balanced accuracy averaged over the biomarkers. Zero means all buckets are read equally well and positive means the common buckets are favored. Each cell is the mean over the three seeds, with the lowest and highest seed in parentheses. The internal set is the held-out validation split, and FLAIR-start configurations are shown on the internal set and WF\_Glaucoma only.}
\label{tab:ablation_gap}
\resizebox{\ifdim\width>\linewidth\linewidth\else\width\fi}{!}{%
\begin{tabular}{lccc}
\toprule
& \multicolumn{1}{c}{Internal validation} & \multicolumn{1}{c}{Drishti-GS} & \multicolumn{1}{c}{WF\_Glaucoma} \\
\cmidrule(lr){2-2}\cmidrule(lr){3-3}\cmidrule(lr){4-4}
Configuration & Accuracy minus balanced accuracy & Accuracy minus balanced accuracy & Accuracy minus balanced accuracy \\
\midrule
\addlinespace[2pt]
\multicolumn{4}{l}{\scriptsize\textsc{Loss function}} \\
\quad SupCon & +0.110{\scriptsize\ (+0.076; +0.132)} & +0.042{\scriptsize\ ($-$0.034; +0.089)} & +0.029{\scriptsize\ ($-$0.014; +0.070)} \\
\quad MP-NCE & +0.108{\scriptsize\ (+0.096; +0.115)} & +0.080{\scriptsize\ (+0.070; +0.095)} & +0.030{\scriptsize\ (+0.016; +0.044)} \\
\quad MP-NCE + auxiliary cosine loss & +0.096{\scriptsize\ (+0.079; +0.105)} & +0.061{\scriptsize\ (+0.021; +0.087)} & $-$0.018{\scriptsize\ ($-$0.057; +0.024)} \\
\addlinespace[2pt]
\multicolumn{4}{l}{\scriptsize\textsc{RETFound start, epoch sampler (with the auxiliary cosine loss)}} \\
\quad Disease-balanced sampler & +0.134{\scriptsize\ (+0.104; +0.149)} & +0.040{\scriptsize\ ($-$0.011; +0.081)} & $-$0.028{\scriptsize\ ($-$0.049; $-$0.014)} \\
\quad Equal-bucket sampler (GlaucomaMarker-R) & +0.007{\scriptsize\ ($-$0.023; +0.031)} & $-$0.042{\scriptsize\ ($-$0.081; +0.015)} & $-$0.003{\scriptsize\ ($-$0.040; +0.041)} \\
\addlinespace[2pt]
\multicolumn{4}{l}{\scriptsize\textsc{RETFound start, text content (disease-balanced sampler)}} \\
\quad Disease sentence only & +0.058{\scriptsize\ ($-$0.008; +0.163)} & $-$0.010{\scriptsize\ ($-$0.021; +0.010)} & $-$0.020{\scriptsize\ ($-$0.054; $-$0.001)} \\
\quad Biomarker sentences shuffled & +0.298{\scriptsize\ (+0.239; +0.336)} & +0.077{\scriptsize\ (+0.064; +0.087)} & +0.060{\scriptsize\ (+0.027; +0.083)} \\
\addlinespace[2pt]
\multicolumn{4}{l}{\scriptsize\textsc{Image-encoder initialization}} \\
\quad ImageNet ViT-L/16 & +0.014{\scriptsize\ (+0.010; +0.023)} & $-$0.002{\scriptsize\ ($-$0.018; +0.009)} & $-$0.085{\scriptsize\ ($-$0.096; $-$0.064)} \\
\quad ImageNet ResNet-50 & $-$0.034{\scriptsize\ ($-$0.044; $-$0.027)} & +0.050{\scriptsize\ (+0.038; +0.059)} & $-$0.018{\scriptsize\ ($-$0.028; $-$0.007)} \\
\addlinespace[2pt]
\multicolumn{4}{l}{\scriptsize\textsc{FLAIR start (WF\_Glaucoma only)}} \\
\quad Plain shuffling, no auxiliary cosine loss & +0.104{\scriptsize\ (+0.098; +0.114)} &  & +0.043{\scriptsize\ (+0.041; +0.045)} \\
\quad Plain shuffling + auxiliary cosine loss & +0.057{\scriptsize\ (+0.042; +0.084)} &  & $-$0.012{\scriptsize\ ($-$0.072; +0.019)} \\
\quad Disease-balanced sampler + auxiliary cosine loss & +0.095{\scriptsize\ (+0.040; +0.140)} &  & +0.045{\scriptsize\ (+0.005; +0.065)} \\
\quad Equal-bucket sampler (GlaucomaMarker-F) & $-$0.031{\scriptsize\ ($-$0.037; $-$0.018)} &  & $-$0.031{\scriptsize\ ($-$0.045; $-$0.018)} \\
\quad Disease sentence only (disease-balanced sampler) & $-$0.028{\scriptsize\ ($-$0.047; $-$0.004)} &  & $-$0.039{\scriptsize\ ($-$0.078; +0.013)} \\
\botrule
\end{tabular}}
\end{table}

\begin{table}[htbp]
\centering
\small
\caption{One-seed comparison of five long-tail rebalancing techniques, each adapted to our multi-positive image-text setting and layered on the MP-NCE base loss, on the RETFound-initialized model trained on the five datasets with masks. The adaptations are simplified from the published methods, and the single seed means the comparison does not rank them. The technique adapted from TSC had the best threshold-based metrics on both cohorts and became the auxiliary cosine loss. Bold marks the best entry in each column.}
\label{tab:ablation_rebalancing_technique}
\resizebox{\ifdim\width>\linewidth\linewidth\else\width\fi}{!}{%
\begin{tabular}{lcccccccccc}
\toprule
& \multicolumn{5}{c}{Drishti-GS (n=101)} & \multicolumn{5}{c}{WF\_Glaucoma (n=374)} \\
\cmidrule(lr){2-6}\cmidrule(lr){7-11}
Rebalancing technique & AUROC & AUPRC & Acc. & F1 & MCC & AUROC & AUPRC & Acc. & F1 & MCC \\
\midrule
MP-NCE + BCL, adapted (Zhu et al. 2022) & \textbf{0.891} & \textbf{0.946} & 0.594 & 0.594 & 0.396 & 0.655 & 0.552 & 0.639 & 0.366 & 0.160 \\
MP-NCE + PaCo, adapted (Cui et al. 2021) & 0.804 & 0.896 & 0.564 & 0.577 & 0.292 & \textbf{0.800} & 0.695 & 0.722 & 0.523 & 0.375 \\
MP-NCE + RCL, adapted (De Alvis et al. 2025) & 0.883 & 0.940 & 0.515 & 0.473 & 0.310 & 0.644 & 0.517 & 0.650 & 0.328 & 0.173 \\
MP-NCE + ReCL, adapted (Chang et al. 2025) & 0.877 & 0.938 & 0.545 & 0.521 & 0.343 & 0.690 & 0.554 & 0.658 & 0.396 & 0.208 \\
MP-NCE + TSC, adapted (Li et al. 2022) & 0.856 & 0.932 & \textbf{0.653} & \textbf{0.679} & \textbf{0.440} & 0.799 & \textbf{0.707} & \textbf{0.725} & \textbf{0.669} & \textbf{0.445} \\
\botrule
\end{tabular}}
\end{table}

\end{document}